\documentclass[11pt]{article}
\usepackage{times}
\usepackage{graphicx}
\usepackage{color}
\usepackage{multirow}
\usepackage[authoryear]{natbib}
\usepackage{rotating}
\usepackage{bbm}
\usepackage{latexsym}
\newcommand{\N}{\mathbb{N}}

\newcommand{\argmax}{\operatornamewithlimits{argmax}}

\newcommand{\bx}{{\bm{x}}}

\newcommand{\1}{\mbox{1}\hspace{-0.25em}\mbox{l}}

\usepackage{algorithm}
\usepackage{algcompatible}

\usepackage{mathrsfs}
 \usepackage{bm}
 \usepackage{amsmath,amssymb} 
 \usepackage{amsthm}
\usepackage{natbib}

\newtheorem{theorem}{Theorem}%[section]
\newtheorem{lemma}{Lemma}%[section]
\newtheorem{definition}{Definition}%[section]
\newcommand{\captionfonts}{\normalsize}

\makeatletter  
\long\def\@makecaption#1#2{%
  \vskip\abovecaptionskip
  \sbox\@tempboxa{{\captionfonts #1: #2}}%
  \ifdim \wd\@tempboxa >\hsize
    {\captionfonts #1: #2\par}
  \else
    \hbox to\hsize{\hfil\box\@tempboxa\hfil}%
  \fi
  \vskip\belowcaptionskip}
\makeatother   
\begin{document}
\hspace{13.9cm}1

\ \vspace{20mm}\\

{\LARGE Active learning for level set estimation under input uncertainty and its extensions}

\ \vspace{10pt} \\
{\bf \large Yu Inatsu$^{\displaystyle 1, \displaystyle \dagger}$,
\ \ Masayuki Karasuyama$^{\displaystyle 1}$, 
\ \ Keiichi Inoue$^{\displaystyle 2}$, 
 \  Ichiro Takeuchi$^{\displaystyle 3, \displaystyle 4}$
}\vspace{10pt} \\
{$^{\displaystyle 1}$Nagoya Institute of Technology}\\
{$^{\displaystyle 2}$The Institute for Solid State Physics, The University of Tokyo}\\
{$^{\displaystyle 3}$Nagoya University}\\
{$^{\displaystyle 4}$RIKEN AIP}\\
{$^{\displaystyle \dagger}$E-mail: inatsu.yu@nitech.ac.jp}
%
% \vspace{10pt} \\ 
%\ \\[-2mm]
%{\bf Keywords:} Active learning, Classification, Gaussian process, Input uncertainty, Cost sensitive

\thispagestyle{empty}
\markboth{}{NC instructions}
\ \vspace{-0mm}\\
%
%Abstract
\begin{center} {\bf Abstract} \end{center}
Testing under what conditions the product satisfies the desired properties is a fundamental problem in manufacturing industry.
If the condition and the property are respectively regarded as the input and the output of a black-box function, this task can be interpreted as the problem called \emph{Level Set Estimation (LSE)} --- the problem of identifying input regions such that the function value is above (or below) a threshold.
Although various methods for LSE problems have been developed so far, there are still many issues to be solved for their practical usage.
As one of such issues, we consider the case where the input conditions cannot be controlled precisely, i.e., LSE problems under \emph{input uncertainty}.
We introduce a basic framework for handling input uncertainty in LSE problem, and then propose efficient methods with proper theoretical guarantees.
The proposed methods and theories can be generally applied to a variety of challenges related to LSE under input uncertainty such as cost-dependent input uncertainties and unknown input uncertainties.
We apply the proposed methods to artificial and real data to demonstrate the applicability and effectiveness.
%%%%%%%%%%%

%%%%%%%%%%%%%%%%%%%%%%%%%%%%%%%%%%%%%%%%%%%%%%%%%%%%%%%%%%%%%%%%%%%%%%%
%%%%%%%%%%%%%%%%%%%%%%%%%%%%%%%%%%%%%%%%%%%%%%%%%%%%%%%%%%%%%%%%%%%%%%%
%%%%%%%%%%%%%%%%%%%%%%%%%%%%%%%%%%%%%%%%%%%%%%%%%%%%%%%%%%%%%%%%%%%%%%%
%%%%%%%%%%%%%%%%%%%%%     			Section 1            %%%%%%%%%%%%%%%%%%%%%%%%
%%%%%%%%%%%%%%%%%%%%%%%%%%%%%%%%%%%%%%%%%%%%%%%%%%%%%%%%%%%%%%%%%%%%%%%
%%%%%%%%%%%%%%%%%%%%%%%%%%%%%%%%%%%%%%%%%%%%%%%%%%%%%%%%%%%%%%%%%%%%%%%
%%%%%%%%%%%%%%%%%%%%%%%%%%%%%%%%%%%%%%%%%%%%%%%%%%%%%%%%%%%%%%%%%%%%%%%
\section{Introduction}
 In this paper, we consider a type of active learning (AL) problem called \emph{level set estimation (LSE)}~\cite{bryan2006active}.
The goal of LSE is to efficiently identify the \emph{level set} $\{\bm x \in D \mid f(\bm x) > h \}$ of an unknown high-cost real-valued function $f:D \to \mathbb{R}$, i.e., the input region in which the function output $f(\bm x)$ is greater than a  threshold $h$.
LSE plays an important role in quality control processes in manufacturing, because engineers want to ensure that all parts of a product satisfy the required properties with as few inspections as possible. 
For example, the task of extracting a region satisfying a required physical property from a solid material can be formulated as an LSE problem. 
In order to investigate a physical property, each position of a solid material is subjected to X-ray irradiation. 
Since X-ray irradiation is costly, it is desirable to find the level set (a region in the solid material in which the required physical properties are satisfied) with as few rounds of X-ray irradiation as possible. 
We also encounter an LSE problem in bio-engineering, e.g., in the task of constructing new functional proteins such as drugs or foods, by artificially modifying amino acid sequences of proteins.
Here, bio-engineers need to identify the level set (the region in the protein feature space in which the protein satisfies the required functional properties) by repeatedly modifying amino acid sequences of proteins. 
Various extensions of LSE problem have also been recently studied \citep{inatsu2020b,inatsu2020a}.

Although various methods for LSE problems have been proposed in the literature \citep{Gotovos:2013:ALL:2540128.2540322,zanette2018robust}, there are still several issues to be solved for their practical usage.
One such issue is to deal with the case where the input conditions cannot be controlled precisely. 
For example, in the case of the solid material described above, it is often the case that the position of the X-ray irradiation cannot be precisely controlled. 
%.
Also, in the bio-engineering example, random errors may occur with a certain probability when making substitutions to the target amino acid.
In order to deal with such a practical issue, it is necessary to consider \emph{LSE problem under input uncertainty} and to construct a method specific to the problem setup.
In this paper, we introduce a basic framework for handling input uncertainty in LSE problem, and then propose efficient method with proper theoretical guarantee.

The proposed method and theory can cover many practical situations resulting from input uncertainty.
One such situation studied in this paper is \emph{cost-sensitive input uncertainty} --- a situation where there is a trade-off between the input uncertainty and the cost.
In such a situation, it is desirable to be able to guarantee the quality of the entire product with as little total cost as possible by effectively combining low cost function evaluation that have high input uncertainty, with high cost function evaluation that have low input uncertainty.
Furthermore, we study a situation where sufficient knowledge about the input uncertainty is not available.
While some existing methods can be used when the mechanism of input uncertainty (e.g., the probability distribution of input uncertainty) is fully known, there is no known method that can guarantee convergence when the knowledge of input uncertainty is insufficient.

The basic strategy of conventional AL methods is to select the inputs in which the uncertainty reduction of the corresponding outputs is beneficial to the target task (see, e.g., \cite{settles2009active}).
Unfortunately, under input uncertainty, this basic AL strategy cannot be used as it is because the input point cannot be freely specified.
In fact, the convergence of existing LSE methods such as \cite{Gotovos:2013:ALL:2540128.2540322,zanette2018robust} cannot be guaranteed under input uncertainty. 
In this paper, we propose an AL method for LSE with input uncertainty that combines the following two components.  The first component is properly takes into account the \emph{integrated uncertainty} according to the input uncertainty distribution, i.e.,  precisely evaluates how the uncertainty of an unknown function decreases using an integral calculation with respect to the input uncertainty. The second component is to randomly select evaluation points. The proposed method is based on combining these two components with   probabilities $1-p_t$  and  $p_t$, respectively.
We first consider the case in which the input uncertainty distribution is known, and then extend the result to the case in which the input uncertainty distribution is unknown. 
We investigate the theoretical properties of the proposed LSE method and show that it can identify the true level set with high probability under certain conditions. 
Furthermore, through numerical experiments using artificial and real datasets, we demonstrate the effectiveness of the proposed method. 

\paragraph{Related works}
Bayesian optimization (BO) based on Bayesian inference has been used for various target tasks including LSE (see \cite{shahriari2016taking} for comprehensive survey of BO).
Several LSE methods based on Gaussian process (GP) model have been studied. 
For example, \cite{bryan2006active} proposed the STRADDLE strategy based on credible intervals. 
In addition, \cite{Gotovos:2013:ALL:2540128.2540322} proposed an LSE method using a confidence region which is the intersection  of credible intervals and derived theoretical bounds. 
Furthermore, recently, \cite{zanette2018robust} proposed an LSE method called MILE based on the expected classification improvement, and \cite {pmlr-v89-shekhar19a} proposed a new LSE with tighter theoretical bounds and lower computational costs. 
Similarly, \cite{NIPS2016_6080} has proposed a method for combining the maximization problem and LSE, and \cite {sui2015safe,DBLP:conf/icml/SuiZBY18,turchetta2016safe,DBLP:conf/aaai/WachiSYO18} have used  LSE for efficient safety area identification.
There are several existing studies dealing with input uncertainty in GP model. 
Recently, \cite{beland2017bayesian} has considered BO for minimizing an integral function which is computed by integrating an unknown function with respect to input distributions, and 
\cite{pmlr-v89-oliveira19a} has proposed an upper confidence bound algorithm under uncertainty inputs. 
Moreover, in the framework of time series analysis, \cite {NIPS2002_2313} has proposed an acquisition function based on the integral with respect to input distributions. 
Furthermore, in the context of Bayesian quadrature (see, e.g., \cite {o1991bayes}), \cite{pmlr-v80-xi18a,DBLP:conf/uai/GessnerGM19} proposed a method for efficiently computing the target integral value with respect to input distributions. 
These existing studies on input uncertainty have some similarities with our study in that they are all based on integral calculations of input uncertainty distributions, but these existing techniques cannot be directly used for LSE under input uncertainty. 
As an existing study dealing with LSE under input uncertainty, \cite{iwazaki2019bayesian} proposed an experimental design method to efficiently identify a set called \emph{Reliable Level Set} defined under input uncertainty.
Their classification target is a probability function $p$ calculated based on  a black-box function $f$ and  input uncertainty.
In contrast,  our classification target is $f$ itself,  and there is a difference in this point.
Although there are many existing studies on cost-sensitive BOs (e.g., \cite{NIPS2013_5086,pmlr-v89-song19b,NIPS2017_7016,scott2011correlated}), 
they all considered cost-dependent output precision, i.e., a situation where the higher the cost, more accurate output values of a black-box function can be obtained. 
However,
none of them deal with cost-dependent input uncertainty.

\begin{figure*}[tb]
\begin{center}
\scalebox{1}{
 \begin{tabular}{c}
 \includegraphics[width=1\textwidth]{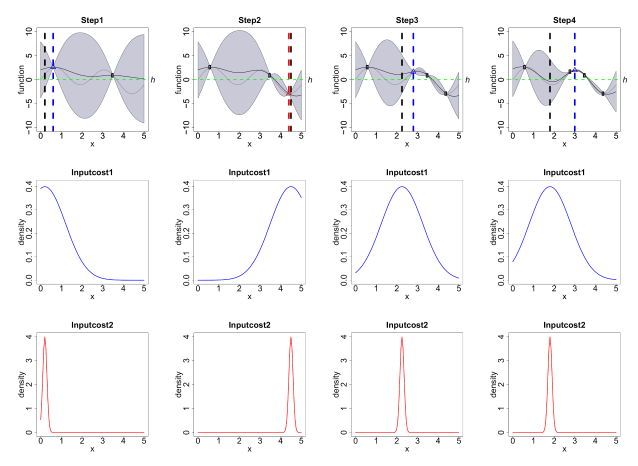} 
 \end{tabular}}
\end{center}
 \caption{
 An illustration of LSE problem with cost-dependent input uncertainty: an example of Gaussian Process  model-based LSE with two different function evaluation options where option 1 has a low cost but high input uncertainty (middle row plots), while option 2 has a high cost but low input uncertainty (bottom row plots).
}
 \label{fig:explanation}
\end{figure*}

\paragraph{Conceptual diagram of the problem setting}
A conceptual diagram of cost-dependent input uncertainty  
is given in Figure \ref{fig:explanation}.
 In the top row plots in Figure \ref{fig:explanation}, the black dashed lines indicate the desired input points, whereas the blue and red dashed lines indicate the actual input points due to input uncertainty. 
 In this example, option 1 (low cost with large input uncertainty) were selected in steps 1, 3, and 4, whereas option 2 (high cost with small input uncertainty) were selected in step 2. 
 The choices of option 1 in steps 1 and 3 (as well as the choice of option 2 in step 2) were effective in the sense that the uncertainty of the GP model was effectively reduced. 
 On the other hand, the choice of option 1 in step 4 was not effective because the function was evaluated at highly different input point and the uncertainty of the GP model could not be effectively reduced. 
 This example illustrates that, in LSE problems with cost-dependent input uncertainty, the proper choice of function evaluation options is important. 

%%%%%%%%%%%%%%%%%%%%%%%%%%%%%%%%%%%%%%%%%%%%%%%%%%%%%%%%%%%%%%%%%%%%%%%%%%%%%%%%%%%%%%%%
%%%%%%%%%%%%%%%%%%%%%%%%%%%%%%%%%%%%%%%%%%%%%%%%%%%%%%%%%%%%%%%%%%%%%%%%%%%%%%%%%%%%%%%%
%%%%%%%%%%%%%%%%%%%%%%%%%%%%%%%%          Figure 1           %%%%%%%%%%%%%%%%%%%%%%%%%%%%%%%%%%%%
%%%%%%%%%%%%%%%%%%%%%%%%%%%%%%%%%%%%%%%%%%%%%%%%%%%%%%%%%%%%%%%%%%%%%%%%%%%%%%%%%%%%%%%%
%%%%%%%%%%%%%%%%%%%%%%%%%%%%%%%%%%%%%%%%%%%%%%%%%%%%%%%%%%%%%%%%%%%%%%%%%%%%%%%%%%%%%%%%
%%%%%%%%%%%%%%%%%%%%%%%%%%%%%%%%%%%%%%%%%%%%%%%%%%%%%%%%%%%%%%%%%%%%%%%%%%%%%%%%%%%%%%%%

\paragraph{Contributions}
Our main contributions in this paper are as follows:
\begin{itemize}
 \item We propose a new AL algorithm for LSE problems under 
input uncertainty by extending the recent LSE method in \cite{zanette2018robust}. 

\item 
We extend our proposed method to various practical situations such as cost-dependent input uncertainty and unknown input distributions.

 \item 
We theoretically analyze the proposed method in depth and show that it has the following four properties. First, we show the convergence of the proposed algorithm, i.e., the desired level set can be identified with probability one under certain regularity conditions. Second, we  show that the number of necessary function evaluations for level set identification was finite with probability one. 
Third, we derive the bound of the number of trials required for the algorithm to complete with high probability.
Finally, we show that these results hold even if  input distributions are unknown.

\item Through numerical experiments using synthetic data and real data, we confirm that our proposed method has the same or better performance than other methods.
\end{itemize}

%%%%%%%%%%%%%%%%%%%%%%%%%%%%%%%%%%%%%%%%%%%%%%%%%%%%%%%%%%%%%%%%%%%%%%%
%%%%%%%%%%%%%%%%%%%%%%%%%%%%%%%%%%%%%%%%%%%%%%%%%%%%%%%%%%%%%%%%%%%%%%%
%%%%%%%%%%%%%%%%%%%%%%%%%%%%%%%%%%%%%%%%%%%%%%%%%%%%%%%%%%%%%%%%%%%%%%%
%%%%%%%%%%%%%%%%%%%%%     			Section 2            %%%%%%%%%%%%%%%%%%%%%%%%
%%%%%%%%%%%%%%%%%%%%%%%%%%%%%%%%%%%%%%%%%%%%%%%%%%%%%%%%%%%%%%%%%%%%%%%
%%%%%%%%%%%%%%%%%%%%%%%%%%%%%%%%%%%%%%%%%%%%%%%%%%%%%%%%%%%%%%%%%%%%%%%
%%%%%%%%%%%%%%%%%%%%%%%%%%%%%%%%%%%%%%%%%%%%%%%%%%%%%%%%%%%%%%%%%%%%%%%
\section{Preliminaries}
Let $f: D \to \mathbb{R}$ be a black-box function on $ D \subset \mathbb{R}^d$ with expensive to evaluate. 
In this paper, we consider an LSE problem for $ f $ on a finite subset $ \Omega $ of $ D $. 
The   upper and lower level sets for $f$ on $\Omega$ at threshold $h$ are defined as follows: 
\begin{definition}
Let $h$ be a threshold. Then, an upper level set $H$ and a lower level set $L$ are defined as 
\begin{align}
H= \{ {\bm{x}} \in \Omega \ | \ f(\bx ) >h \} , \  L= \{ {\bm{x}} \in \Omega \ | \ f(\bx ) \leq h \} .  \nonumber 
\end{align}
\end{definition} 
In this paper, we consider 
 settings where inputs have uncertainty. 
For more generality, hereafter, we consider 
cost-dependent input uncertainties for an input point ${\bm{x}}$. 
Assume that we have $k$ different options (such as equipments and apparatus) for obtaining ${\bm{x}}$, and these options have different costs 
$0<c_1 <c_2 \cdots <c_k$. 
When an option $i \in \{1,\ldots,k\} \equiv [k]$ is used for obtaining  ${\bm{x}} \in \Omega$, 
the actual obtained point is not  ${\bm{x}}$ but at ${\bm{s}} ({\bm{x}},c_i ) \in D$ where ${\bm{s}} ({\bm{x}},c_i) $ is considered as a random sample 
from a random variable ${\bm{S}} ({\bm{x}},c_i)$.
Thus, the actual function evaluation is also done not 
exactly at ${\bm{x}}$ but at ${\bm{s}} ({\bm{x}},c_i ) \in D$.
Moreover, for each ${\bm{s}} ({\bm{x}},c_i )$, assume that  the value of $f({\bm{s}} ({\bm{x}},c_i ))$ can be observed as $f({\bm{s}} ({\bm{x}},c_i )) + \varrho$, where  
$\varrho$ is an independent Gaussian noise distributed as $\mathcal{N} (0,\sigma^2)$. 
Since there is a trade-off between the costs and the input uncertainties, we need to select appropriate  evaluation 
options from the $k$ different choices at each step. 
Note that if $k = 1$, only a single cost can be used. 
Thus, this is a more general formulation  including the classical case. 
In this paper, we first assume that the probability density function 
 of ${\bm{S}} ({\bm{x}}, c_i) $, 
denoted by 
$g({\bm{s}}|  {\bm\theta}^{(c_i)}_{{\bm{x}}} ) $ with parameters $ {\bm\theta}^{(c_i)}_{{\bm{x}}} $, is known\footnote{Note that we assume that  ${\bm{S}}(\bx,c_i)$  is a continuous random variable, but the discussion in this paper can be applied even if  ${\bm{S}}(\bx,c_i)$ is  discrete.
In that case, the integration operation in Section \ref{sec:proposed} must be replaced with a summation operation.}, but 
later extends to the case where the parameters are unknown and must be estimated in Section  \ref{some-extension}.

\subsection{Gaussian process}
In this paper, GP is used for modeling the black-box function $ f $. 
Let $ \mathcal {G} \mathcal {P} (0, k (\bm s, \bm s^\prime)) $ be a GP prior  for the function $ f $, 
where $k(\bm s, \bm s^\prime): D \times D \to \mathbb{R}$ is a positive-definite kernel. 
Therefore, for any finite set of points $ \bm s_1, \ldots, \bm s_t \in D $, 
a joint distribution of $ (f (\bm s_1), \ldots, f (\bm s_t)) ^\top $ is given by $\mathcal{N}_t ({\bm\mu} _t, {\bm {K}} _ t)$, where 
$ \mathcal{N}_t ({\bm\mu}_t, {\bm{K}}_t)$ is 
a $t$-dimensional normal distribution with mean vector  $\bm\mu_t$ and covariance matrix  ${\bm{K}}_t$, 
${\bm\mu}_t=(0,\ldots,0)^\top \equiv {\bm{0}}_t$, and the $(i,j)$ element of ${\bm{K}}_t$ is $k(\bm s_i,\bm s_j)$.  
From the properties of GP, a posterior distribution of $f$ after adding the data set  $\{ (  {\bm{s}}_j ( {\bm{x}}_j,c_{i_j}  ) , y_j \}^t_{j=1} $ 
 is also  GP. 
Then, a posterior mean $ \mu_t (\bm x) $, variance  $ \sigma^2_t (\bm x) $ and covariance $ k_t (\bm x, \bm x ') $ of $f$ at $\bx$
 are given by 
\begin{align*}
\mu_t (\bm x) = {\bm{k}}_t (\bm x) ^\top{\bm{C}}_t^{-1} {\bm{y}}_t,
\sigma^2_t (\bm x)= k_t (\bm x,\bm x),\
k_t(\bm x,\bm x')= k(\bm x,\bm x')- {\bm{k}}_t (\bm x) ^\top {\bm{C}}_t^{-1} {\bm{k}}_t (\bm x^\prime),
\end{align*}
where ${\bm{k}}_t (\bm x)= (k(\bm s_1 ({\bm{x}}_1,c_{i_1} ),\bm x),\ldots,k(\bm s_t ({\bm{x}}_t,c_{i_t} ),\bm x))^\top$, 
${\bm{C}}_t = ({\bm{K}}_t +\sigma^2 {\bm{I}}_t)$, 
$\bm y_t =(y_1,\ldots,y_t)^\top$, and 
${\bm{I}}_t$ is a $t$-dimensional identity matrix.

%%%%%%%%%%%%%%%%%%%%%%%%%%%%%%%%%%%%%%%%%%%%%%%%%%%%%%%%%%%%%%%%%%%%%%%
%%%%%%%%%%%%%%%%%%%%%%%%%%%%%%%%%%%%%%%%%%%%%%%%%%%%%%%%%%%%%%%%%%%%%%%
%%%%%%%%%%%%%%%%%%%%%%%%%%%%%%%%%%%%%%%%%%%%%%%%%%%%%%%%%%%%%%%%%%%%%%%
%%%%%%%%%%%%%%%%%%%%%     			Section 3            %%%%%%%%%%%%%%%%%%%%%%%%
%%%%%%%%%%%%%%%%%%%%%%%%%%%%%%%%%%%%%%%%%%%%%%%%%%%%%%%%%%%%%%%%%%%%%%%
%%%%%%%%%%%%%%%%%%%%%%%%%%%%%%%%%%%%%%%%%%%%%%%%%%%%%%%%%%%%%%%%%%%%%%%
%%%%%%%%%%%%%%%%%%%%%%%%%%%%%%%%%%%%%%%%%%%%%%%%%%%%%%%%%%%%%%%%%%%%%%%
\section{Proposed method}\label{sec:proposed}
In this section, we propose an efficient AL method for LSE under cost-dependent input uncertainty.
First, we explain a LSE method based on credible intervals.

\subsection{Credible interval and LSE}
For each ${\bm x} \in \Omega$, 
let $Q_t ({\bm x}) =[ l_t ({\bm x}), u_t ({\bm x})]$ be a credible interval of $f({\bm x})$ at the $t$th trial, where 
$l_t ({\bm x}) = \mu_t ({\bm x}) - \beta^{1/2} \sigma_t ({\bm x}) $, 
 $u_t ({\bm x}) = \mu_t ({\bm x}) + \beta^{1/2} \sigma_t ({\bm x}) $, and $\beta^{1/2} \geq 0$. 
In addition, let $\epsilon$ be a positive accuracy parameter. 
Then,  we define estimated sets $H_t$ and $L_t$ respectively of $H$ and $L$  as  
\begin{equation}
H_t = \{ {\bm x} \in \Omega \ | \ l_t ({\bm x}) > h- \epsilon \} , \ 
L_t = \{ {\bm x} \in \Omega \ | \ u_t ({\bm x}) < h+ \epsilon \} .  \label{eq:HtLt}
\end{equation}
Moreover, we  define  an unclassified set $U_t = \Omega \setminus (H_t \cup L_t)$. 
Each step of LSE can be interpreted as the problem of classifying ${\bm{x}} \in U_{t-1} $ into $H_t$ or $L_t$. 
From \eqref{eq:HtLt}, $H_t$ and $L_t$ depend on $Q_t ({\bm x})$, and $Q_t ({\bm x})$ is calculated based on input points (and its function values).
Hence, in order to obtain \eqref{eq:HtLt} efficiently, it is important to appropriately determine a next input point to be evaluated.
Furthermore, in this paper we consider $k$ different costs for obtaining an input ${\bm{x}}$.
In the next subsection, we propose an acquisition function  to determine the next input point and the evaluation cost of the input point
 under cost-dependent input uncertainty.

\subsection{Acquisition function}
 We extend the MILE acquisition function proposed by  \cite{zanette2018robust}. 
MILE is based on the idea that the next evaluation point is the point that maximizes the expected classification improvement when a new point is added. 
Since inputs have cost-dependent uncertainty in our setting,  
 we consider the \emph{integral} with respect to the input distribution of the expected classification improvement, and define the integral divided by the cost as our acquisition function value. 
Moreover, by using the 
randomized strategy, we can show that our proposed algorithm converges with probability 1. 

\paragraph{Integral with respect to input of expected classification improvement per unit cost}\label{subsubsec1}
Let ${\bm{s}} ^\ast \in D$ be a new point, and let $y^\ast = f( {\bm{s}} ^\ast )+ \varrho$ be the observed value for  ${\bm{s}} ^\ast$. 
In addition, let $H_{t} ({\bm{s}}^\ast,y^\ast)$ and 
$L_{t} ({\bm{s}}^\ast,y^\ast)$ be estimated sets respectively of $H$ and $L$ when 
$ ({\bm{s}}^\ast,y^\ast)$ is added, and let 
$HL_{t} ({\bm{s}}^\ast,y^\ast) = H_{t} ({\bm{s}}^\ast,y^\ast) \cup L_{t} ({\bm{s}}^\ast,y^\ast)$, 
$HL_t = H_t \cup L_t$. 
 Then, when the observation cost of the input point $ {\bm x} \in \Omega $  is $c_i$, the integral  of the expected classification improvement  per unit cost is given by
\begin{align}
a_t ({\bm x},c_i ) 
=   c^{-1}_i   \left \{ \int   {\rm E}_{y^\ast} [|  HL_{t} ({\bm{s}}^\ast,y^\ast) | 
]   g( {\bm{s}} ^\ast| \theta^{(c_i)}_{\bm{x}} ) d {\bm{s}^\ast}  
 - |HL_t | \right \} . \label{eq:AFteigi} 
\end{align}
Furthermore, the expectation in \eqref{eq:AFteigi} can be written as follows:
\begin{lemma}\label{lem:AF}
The expectation  in  \eqref{eq:AFteigi} can be written as 
\begin{align*}
  {\rm E}_{y^\ast} [|  HL_{t} ({\bm{s}}^\ast,y^\ast) | 
]    
&=   \sum_{ {\bm{a}} \in \Omega }  \Phi  \left ( 
\frac{\sqrt{\sigma^2_{t} ({\bm{s}}^\ast)+\sigma^2}}{|k_t ({\bm{a}},{\bm{s}}^\ast)|} \times c^+_{t} ({\bm{a}}|{\bm{s}}^\ast )
\right )   
 \\
&\quad + \sum_{ {\bm{a}} \in \Omega }  \Phi  \left ( 
\frac{\sqrt{\sigma^2_{t} ({\bm{s}}^\ast)+\sigma^2}}{|k_t ({\bm{a}},{\bm{s}}^\ast)|} \times c^-_{t} ({\bm{a}}|{\bm{s}}^\ast ) 
\right ) .
\end{align*}
Here, $c^+_{t} ({\bm{a}}|{\bm{s}}^\ast ) $ and $c^-_{t} ({\bm{a}}|{\bm{s}}^\ast )$ are respectively given by 
\begin{align*}
c^+_{t} ({\bm{a}}|{\bm{s}}^\ast )  &=  \mu_{t} ({\bm{a}}) - \beta^{1/2} \sigma_{t} ({\bm{a}}|{\bm{s}}^\ast ) -h+\epsilon, \\
c^-_{t} ({\bm{a}}|{\bm{s}}^\ast )  &=  -\mu_{t} ({\bm{a}}) - \beta^{1/2} \sigma_{t} ({\bm{a}}|{\bm{s}}^\ast ) +h+\epsilon,
\end{align*}
 and 
$\sigma^2_t ({\bm{a}} |{\bm{s}}^\ast)$ is the posterior variance of $f$ at the point ${\bm{a}}$ after adding 
${\bm{s}}^\ast$ to the dataset 
 $\{ (  {\bm{s}}_j ( {\bm{x}}_j,c_{i_j}  ) , y_j \}^t_{j=1} $. 
Moreover, when $k_t ({\bm{a}},{\bm{s}}^\ast) =0$, we define that 
\begin{align*}
 \Phi  \left ( 
\frac{\sqrt{\sigma^2_{t} ({\bm{s}}^\ast)+\sigma^2}}{|k_t ({\bm{a}},{\bm{s}}^\ast)|} \times c^+_{t} ({\bm{a}}|{\bm{s}}^\ast )
\right )   
&= \left \{
\begin{array}{ll}
1 & \text{if} \quad c^+_{t} ({\bm{a}}|{\bm{s}}^\ast )  >0 \\
0 & \text{otherwise}
\end{array}
\right. , \\
 \Phi  \left ( 
\frac{\sqrt{\sigma^2_{t} ({\bm{s}}^\ast)+\sigma^2}}{|k_t ({\bm{a}},{\bm{s}}^\ast)|} \times c^-_{t} ({\bm{a}}|{\bm{s}}^\ast )
\right )   
&= \left \{
\begin{array}{ll}
1 & \text{if} \quad c^-_{t} ({\bm{a}}|{\bm{s}}^\ast )  >0 \\
0 & \text{otherwise}
\end{array}
\right. .
\end{align*}
\end{lemma}
The proof is given in  Appendix \ref{app1}. 
Moreover, the details of approximation for the integral in \eqref{eq:AFteigi} are given in Subsection \ref{app-approx}.

\paragraph{The randomized strategy}\label{subsubsec2}
In the proposed algorithm,  
we   select the pair $({\bm{x}}, c_i )$ stochastically. 
Let  $\mathcal{C} = \{  ({\bm{x}}, c_i ) \ | \ {\bm{x}} \in \Omega, \ i \in [k] \}$, and let 
 $C_t$ be a discrete random variable whose range is $\mathcal{C}$. Furthermore, let 
$\kappa_i= {\rm P} (C_t =  ( {\bm{x}}, c_i ) ) $ 
 be a probability mass function of  $C_t$, 
where 
$0<\kappa_i <1$ and $|\Omega| \sum_{i=1}^k \kappa_i =1$.
Note that the subscript of  $\kappa_i$ is independent of ${\bm x}$. 
In other words,  all pairs with the same cost $c_i$ are selected with equal probability $\kappa_i/|\Omega|$.

\subsection{Proposed algorithm}
Using the results so far, we propose an algorithm for LSE with cost-dependent input uncertainty as follows. 
For each trial,  $({\bm x} , c_i)$ is chosen by 
 maximizing $a_t ({\bm{x}},c_i) $ with probability $1-p_t $,  and otherwise   $({\bm x} , c_i)$ is chosen based on the randomized strategy. 
The pseudo code of the proposed algorithm is given in Algorithm \ref {ALG1}, 
where  $\mathcal{B} (p_t)$ is Bernoulli distribution which takes 1 with probability $p_t$. 
Note that in the 10th line of Algorithm \ref{ALG1},  the argmax operator for $a_t ({\bm x},c_i)$ is evaluated over all candidate points.
Recall that $a_t ({\bm x},c_i)$ is   the expected classification improvement per unit cost. 
In addition, due to input uncertainty, $a_t ({\bm x},c_i)$ may become large in the case of evaluating classified points  than that of evaluating unclassified points.
For this reason, the argmax operator is evaluated over all candidate points instead of $U_t$.
Incidentally, even if the argmax operator is evaluated over $U_t$, theoretical guarantees given in Section \ref{sec:THEOREM} hold because its proofs are based on the randomized strategy.
However, in the practical sense, taking the argmax on all candidate points works better. 
Also note that the combination of maximizing $a_t ({\bm x},c_i)$ and using the randomized strategy  is similar to the $\epsilon$-greedy algorithm for reinforcement learning (see, e.g., \cite{sutton2018reinforcement}).
By using the $\epsilon$-greedy algorithm, it can be expected that a better point than a current optimal solution is selected.
As a result, it is possible to avoid the problem that a non-optimal point is repeatedly selected.
Similarly, in our proposed algorithm, using the randomized strategy enables us to avoid   the problem that  the same point is repeatedly  selected  and the classification does not complete.
It should be noted that this problem is particularly significant 
when input distributions are misspecified.
In this sense, using the randomized strategy works better when input distributions are misspecified (usefulness of the randomized strategy is confirmed in Subsection \ref{experiment1}).

\subsection{Approximation of $a_t ({\bm x},c_i )$} \label{app-approx}
Since  integral operation about ${\bm{S}} ({\bm{x}},c_i)$ in $a_t ({\bm x}, c_i) $ is computationally expensive, 
 we consider two approximations of $a_t ({\bm x}, c_i) $. 

Let  ${\bm{s}}^{(1)} ({\bm x},c_i), \ldots, {\bm{s}}^{(M)} ({\bm x},c_i)$ be independent random variables from 
${\bm{S}} ({\bm x} , c_i) $. Then, 
 $a_t ({\bm x}, c_i) $ can be approximated as 
\begin{equation}
\begin{split}
&a_t ({\bm x},c_i ) \\
&\approx c^{-1}_i M^{-1} \\
&\left [ \sum_{j=1}^M \left \{ \sum_{ {\bm{a}} \in \Omega }  \Phi  \left ( 
\frac{\sqrt{\sigma^2_{t} ({\bm{s}}^{(j)} ({\bm x},c_i))+\sigma^2}}{|k_t ({\bm{a}},{\bm{s}}^{(j)} ({\bm x},c_i))|} 
 \times ( \mu_{t} ({\bm{a}}) - \beta^{1/2} \sigma_{t} ({\bm{a}}|{\bm{s}}^{(j)} ({\bm x},c_i) ) -h+\epsilon )
\right )  \right . \right .  \\  
& \left . 
  \left . + \sum_{ {\bm{a}} \in \Omega }  \Phi  \left ( 
\frac{\sqrt{\sigma^2_{t} ({\bm{s}}^{(j)} ({\bm x},c_i))+\sigma^2}}{|k_t ({\bm{a}},{\bm{s}}^{(j)} ({\bm x},c_i)|} 
 \times ( -\mu_{t} ({\bm{a}}) - \beta^{1/2} \sigma_{t} ({\bm{a}}|{\bm{s}}^{(j)} ({\bm x},c_i)) +h+\epsilon )
\right ) \right . \right . \\
& \left . \left . 
  - |HL_t  | \right \} \right ]  .
\end{split} \label{eq:approx1}
\end{equation}
However, in  \eqref{eq:approx1}, 
 it is necessary to compute the posterior variance for each 
   ${\bm{s}}^{(j)} ({\bm x},c_i)$. 
As the result,           the computational cost of \eqref{eq:approx1} is 
 $\mathcal{O} (t^2|\Omega| M)$. 
Therefore,  the total computational cost required for one trial is 
$\mathcal{O} ( t^2 k |\Omega|^2 M )$ because it is necessary to compute for all 
 ${\bm x} \in \Omega$ and $c_i $, $i \in [k]$. 

As another choice, we can consider the following  approximate distribution of $ {\bm {S}} ({\bm x}, c_i) $. 
Let $[{\bm{a}}]_\Omega$ be an element of  $\Omega$ which is the closest point to ${\bm{a}}$. 
Then, we define $[{\bm{S}} ({\bm x} , c_i)]_\Omega \equiv \tilde{{\bm{S}}} ({\bm x} , c_i)$. 
Note that $ \tilde{{\bm{S}}} ({\bm x} , c_i)$ is the discrete random variable whose observed value  is in $\Omega$. 
Then,  $a_t ({\bm x}, c_i) $ can be approximated by using $ \tilde{{\bm{S}}} ({\bm x} , c_i)$ as 
\begin{equation}
\begin{split}
&a_t ({\bm x},c_i ) \\
&\approx c^{-1}_i  \sum_{{\bm{b}} \in \Omega} \left \{ \sum_{ {\bm{a}} \in \Omega }  \Phi  \left ( 
\frac{\sqrt{\sigma^2_{t} ({\bm{b}})+\sigma^2}}{|k_t ({\bm{a}},{\bm{b}})|}   \times ( \mu_{t} ({\bm{a}}) - \beta^{1/2} \sigma_{t} ({\bm{a}}|{\bm{b}} ) -h+\epsilon )
\right )  \right .  \\  
&  \left . 
+ \sum_{ {\bm{a}} \in \Omega }  \Phi  \left ( 
\frac{\sqrt{\sigma^2_{t} ({\bm{b}})+\sigma^2}}{|k_t ({\bm{a}},{\bm{b}})|}
  \times ( -\mu_{t} ({\bm{a}}) - \beta^{1/2} \sigma_{t} ({\bm{a}}|{\bm{b}}) +h+\epsilon )
\right ) 
- |HL_t| \right \} p_{ \tilde{{\bm{S}}} ({\bm x} , c_i)} ( {\bm{b}} ) , 
\end{split} \label{eq:approx2}
\end{equation}
where 
$p_{ \tilde{{\bm{S}}} ({\bm x} , c_i)} ( {\bm{a}} ) \equiv {\rm P} ( \tilde{{\bm{S}}} ({\bm x} , c_i) = {\bm{a}} )$ is the probability mass function of 
$ \tilde{{\bm{S}}} ({\bm x} , c_i)$. 
Unlike \eqref{eq:approx1}, in \eqref{eq:approx2}, 
the calculation results in the braces $\{ \}$ are same for all 
 ${\bm x} \in \Omega$ and $c_i$. 
 Thus, 
 for the calculation in the braces $\{ \}$, it is sufficient to calculate once for each 
 ${\bm{a}} \in \Omega$ and ${\bm{b}} \in \Omega$, and its calculation cost is given by  $\mathcal{O} ( t^2 |\Omega|^2)$. 
Moreover, the computational cost required to calculate $ a_t ({\bm x}, c_i) $ is $ \mathcal {O} (| \Omega |) $. 
Therefore, the total cost of calculating  $a_t ({\bm x}, c_i) $ is given by $ \mathcal{O} (  (t^2 +k) |\Omega|^2  )$. 
This approximation is useful when $ \tilde {{\bm {S}}} ({\bm x}, c_i) $ is a good approximation of $ {{\bm {S}}} ({\bm x}, c_i) $.

\begin{algorithm}[!t]                  
\caption{LSE under cost dependent input uncertainty}         
\label{ALG1}                          
\begin{algorithmic}[1]                  
\REQUIRE 
 Initial training data,  
GP prior $\mathcal{G}\mathcal{P} (0,k({\bm x},{\bm x}'))$, 
probabilities $\{ p_t \}_{t \in \N} $,  $\{ \kappa_j \}_{j=1}^{k}$, 
accuracy parameters $\beta \geq 0$, $\epsilon >0$
\ENSURE Estimated sets $\widehat{H}$ and $ \widehat{L}$
\STATE ${H}_0 \gets \emptyset$, ${L}_0\gets \emptyset$, $U_0 \gets \Omega$
\STATE $t \gets 1$
\WHILE{$U_{t-1} \neq \emptyset$}
	\STATE ${H}_t \gets {H}_{t-1}$, ${L}_t\gets {L}_{t-1}$, $U_t \gets U_{t-1}$
	\FORALL{$ {\bm x} \in \Omega $}
	\STATE Compute credible interval  $Q_t ({\bm x})$   from GP
	\ENDFOR
	\STATE 
Compute $H_t$, $L_t$ and $U_t$ from \eqref{eq:HtLt} and generate $r_t $ from $\mathcal{B} (p_t)$
		\IF{$r_t=0$} 
		\STATE $({\bm x}_{t},c_{i_{t}} ) = \argmax _{({\bm x}, c_i)  }  a_t ({\bm x},c_i)$
		 \ELSIF{$r_t=1 $} \label{line11}
		\STATE Generate $({\bm x}_{t},c_{i_{t}} )$ from $C_t$ \label{line12}
		 \ENDIF
	\STATE Generate   ${\bm{s}}_{t} ({\bm x}_{t},c_{i_{t}} ) $ from ${\bm{S}} ({\bm x}_{t},c_{i_{t}} ) $ 
\STATE $y_t \gets f({\bm{s}}_{t} ({\bm x}_{t},c_{i_{t}} )  ) +\varrho_t$
\STATE $t \gets t+1$
\ENDWHILE
\STATE $\widehat{H} \gets {H}_{t-1}$, $\widehat{L} \gets {L}_{t-1}$
\end{algorithmic}
\end{algorithm}

%%%%%%%%%%%%%%%%%%%%%%%%%%%%%%%%%%%%%%%%%%%%%%%%%%%%%%%%%%%%%%%%%%%%%%%
%%%%%%%%%%%%%%%%%%%%%%%%%%%%%%%%%%%%%%%%%%%%%%%%%%%%%%%%%%%%%%%%%%%%%%%
%%%%%%%%%%%%%%%%%%%%%%%%%%%%%%%%%%%%%%%%%%%%%%%%%%%%%%%%%%%%%%%%%%%%%%%
%%%%%%%%%%%%%%%%%%%%%     			Section 4            %%%%%%%%%%%%%%%%%%%%%%%%
%%%%%%%%%%%%%%%%%%%%%%%%%%%%%%%%%%%%%%%%%%%%%%%%%%%%%%%%%%%%%%%%%%%%%%%
%%%%%%%%%%%%%%%%%%%%%%%%%%%%%%%%%%%%%%%%%%%%%%%%%%%%%%%%%%%%%%%%%%%%%%%
%%%%%%%%%%%%%%%%%%%%%%%%%%%%%%%%%%%%%%%%%%%%%%%%%%%%%%%%%%%%%%%%%%%%%%%
\section{Extensions}\label{some-extension}
In this section, we give two extensions of the proposed method. 
The first is an extension to the situation where  error variances  also change depending on  costs, 
and the second  covers the case where   input distributions are unknown.

\subsection{Cost-dependent error variances}\label{subsec:extend_var}
Let $c^{(\text{out})}_1,\ldots , c^{(\text{out})}_{k^\ast} $ be costs with 
$0<c^{(\text{out})}_1 < \cdots < c^{(\text{out})}_{k^\ast} $. 
For each $c^{(\text{out})}_o $, $o \in [k^\ast]$ and  ${\bm{s}} \in D$, a value of $f$ can be observed as 
 $y^{(o)} = f({\bm{s}}) + \varrho ^{(o)} $, 
where $\varrho ^{(o)}$ is an independent Gaussian noise distributed as 
$ \mathcal{N} (0, \sigma^{( o) 2} )$.  
Then, the posterior mean, variance and covariance of     $f$ after adding    the dataset   $\{ (  {\bm{s}}_j ( {\bm{x}}_j,c_{i_j}  ) , y^{(o_j)}_j \}^t_{j=1} $ are given by 
\begin{align*}
\mu_t (\bm x) = {\bm{k}}_t (\bm x) ^\top{\bar{\bm{C}}}_t^{-1} {\bm{y}}_t,
\sigma^2_t (\bm x)= k_t (\bm x,\bm x),\ 
k_t(\bm x,\bm x')= k(\bm x,\bm x')- {\bm{k}}_t (\bm x) ^\top {\bar{\bm{C}}}_t^{-1} {\bm{k}}_t (\bm x^\prime),
\end{align*}
 where 
${\bar{\bm{C}}}_t = {\bm{K}}_t +\text{diag}  (\sigma^{(o_1)2},  \ldots ,  \sigma^{(o_t)2}       )$. 
In this case, if   observation costs for   the input point $ {\bm x} \in \Omega $ and the function value are respectively 
$ c_i $ and $ c ^ {(\text {out})} _ o $, 
then the integral $ a_t ({\bm x}, c_i, c ^ {(\text {out})} _ o) $ of the expected classification improvement per unit cost can be defined  in the same way as  \eqref{eq:AFteigi}. 
Therefore, similarly to Lemma \ref{lem:AF}, $ a_t ({\bm x}, c_i, c ^ {(\text {out})} _ o) $ can be written as follows:
\begin{lemma}\label{lem:AFver2}
The acquisition function $ a_t ({\bm x}, c_i, c ^ {(\text {out})} _ o) $ can be written as
\begin{align}
&a_t ({\bm x},c_i, c^{(\text{out})}_o  ) \nonumber \\
&= (c_i+c^{(\text{out})}_o)^{-1}  \int \left [ \sum_{ {\bm{a}} \in \Omega }  \left \{ \Phi  \left ( 
\frac{\sqrt{\sigma^2_{t} ({\bm{s}}^\ast)+\sigma^{(o)2}}}{|k_t ({\bm{a}},{\bm{s}}^\ast)|} c^+_{t} ({\bm{a}}|{\bm{s}}^\ast ) 
\right )  \right .   
 \right  . \nonumber \\
&\quad \quad \quad \quad \quad \quad \quad \quad+    \left . \left . \Phi  \left ( 
\frac{\sqrt{\sigma^2_{t} ({\bm{s}}^\ast)+\sigma^{(o)2}}}{|k_t ({\bm{a}},{\bm{s}}^\ast)|} c^-_{t} ({\bm{a}}|{\bm{s}}^\ast )
\right ) \right \}
- |HL_t | \right ]     g( {\bm{s}} ^\ast| \theta^{(c_i)}_{\bm{x}} ) d {\bm{s}^\ast}. \label{eq:AFteigiver2}
\end{align}
Here, if $k_t ({\bm{a}},{\bm{s}}^\ast) =0$, $\Phi ( \cdot )$ is defined as in Lemma  \ref{lem:AF}. 
\end{lemma}
This lemma can be proven by following the same line of the proof of  Lemma \ref{lem:AF}. 
In addition, similarly in Subsection \ref{subsubsec2}, we consider to select the pair $({\bm{x}}, c_i ,c^{(\rm out)}_j)$ stochastically. 
Let $\mathcal{\tilde{C}} = \{  ({\bm{x}}, c_i, c^{(\rm out)} _j) \ | \ {\bm{x}} \in \Omega, \ i \in [k] , \ j \in [k^\ast] \}$, and let  
$\tilde{C}_t$ be a discrete random variable whose range is $\mathcal{\tilde{C}}$. 
Moreover, let $\kappa_{i,j} = {\rm P} (\tilde{C}_t =  ( {\bm{x}}, c_i,c^{(\rm out)} _j ) )$ be a probability mass function of $\tilde{C}_t$, where 
$0 < \kappa_{i,j} <1$ and $|\Omega | \sum _{i=1}^k  \sum_{j=1}^{k^\ast} \kappa_{i,j}=1$.
 Then, under the cost-dependent noise variances setting, the pseudo code of the proposed algorithm is given in Algorithm \ref{ALG2}.

\begin{algorithm}[!t]                  
\caption{LSE under cost dependent input uncertainty and noise variance}         
\label{ALG2}                          
\begin{algorithmic}[1]                  
\REQUIRE 
 Initial training data,  
GP prior $\mathcal{G}\mathcal{P} (0,k({\bm x},{\bm x}'))$, 
probabilities $\{ p_t \}_{t \in \N} $,  $\{ \kappa_{i,j} \}$, 
accuracy parameters $\beta \geq 0$, $\epsilon >0$
\ENSURE Estimated sets $\widehat{H}$ and $ \widehat{L}$
\STATE $\widehat{H}_0 \gets \emptyset$, $\widehat{L}_0\gets \emptyset$, $U_0 \gets \Omega$
\STATE $t \gets 1$
\WHILE{$U_{t-1} \neq \emptyset$}
	\STATE $\widehat{H}_t \gets \widehat{H}_{t-1}$, $\widehat{L}_t\gets \widehat{L}_{t-1}$, $U_t \gets U_{t-1}$
	\FORALL{$ {\bm x} \in \Omega $}
	\STATE Compute credible interval  $Q_t ({\bm x})$   from GP
	\ENDFOR
	\STATE 
Compute $H_t$, $L_t$ and $U_t$ from \eqref{eq:HtLt} and generate $r_t $ from $\mathcal{B} (p_t)$
		\IF{$r_t=0$} 
		\STATE $({\bm x}_{t},c_{i_{t}}, c^{(\rm out)}_ {j_t}  ) = \argmax _{({\bm x}, c_i,c^{(\rm out)}_j)  }  a_t ({\bm x},c_i,c^{(\rm out)}_j)$
		 \ELSIF{$r_t=1 $} \label{line11}
		\STATE Generate $({\bm x}_{t},c_{i_{t}} ,c^{(\rm out)}_ {j_t} )$ from $\tilde{C}_t$ \label{line12}
		 \ENDIF
	\STATE Generate   ${\bm{s}}_{t} ({\bm x}_{t},c_{i_{t}} ) $ from ${\bm{S}} ({\bm x}_{t},c_{i_{t}} ) $ 
\STATE $y_t \gets f({\bm{s}}_{t} ({\bm x}_{t},c_{i_{t}} )  ) +\varepsilon^{(j_t)}_t$
\STATE $t \gets t+1$
\ENDWHILE
\STATE $\widehat{H} \gets \widehat{H}_{t-1}$, $\widehat{L} \gets \widehat{L}_{t-1}$
\end{algorithmic}
\end{algorithm}

\subsection{Unknown input distributions}\label{subsec:extend}
Here, we discuss the case where the density function $g({\bm{s}} | {\bm\theta}^{(c_i)}_{\bm x} )$ is unknown. 
In this case, it is necessary to estimate it. 
One natural approach is to estimate an unknown parameter  ${\bm\theta}^{(c_i)}_{\bm x}$ under the assumption that the density function has the known form $g({\bm{s}} | {\bm\theta}^{(c_i)}_{\bm x} )$.
However, if we assume a different $ {\bm \theta} ^ {(c_i)} _ {\bm x} $ for each point $ {\bm x} \in \Omega $ (and $ c_i $), it is difficult to estimate the parameters. 
For this reason, we assume that ${\bm\theta}^{(c_i)}_{\bm x}$ can be expressed as 
 ${\bm\theta}^{(c_i)}_{\bm x} =  (  \tilde{\bm\theta}^{(c_i)} _{\bm x} ,  {\bm\xi}^{(c_i) } ) $, 
 where $ \tilde{\bm\theta}^{(c_i)} _{\bm x}$ is known, and $ {\bm\xi}^{(c_i) } $ is unknown. 
Then, by assuming a prior distribution $\pi ( {\bm\xi}^{(c_i) }) $ for 
$ {\bm\xi}^{(c_i) }$,   we can compute the posterior distribution $\pi_t ( {\bm\xi}^{(c_i) }) $ after adding 
the data $\{ ( {\bm{s}}_j ({\bm x}_j,c_{i,j}),y_j ) \}^t_{j=1} $. 
Therefore, by using this,  
 $g({\bm{s}} | {\bm\theta}^{(c_i)}_{\bm x} )$
 can be estimated as 
\begin{align}
g_t ({\bm{s}} | {\bm\theta}^{(c_i)}_{\bm x} ) =  \int  g({\bm{s}} | {\bm\theta}^{(c_i)}_{\bm x} )  \pi_t ( {\bm\xi}^{(c_i) })  d  {\bm\xi}^{(c_i) } . \nonumber% 
\end{align}

%%%%%%%%%%%%%%%%%%%%%%%%%%%%%%%%%%%%%%%%%%%%%%%%%%%%%%%%%%%%%%%%%%%%%%%
%%%%%%%%%%%%%%%%%%%%%%%%%%%%%%%%%%%%%%%%%%%%%%%%%%%%%%%%%%%%%%%%%%%%%%%
%%%%%%%%%%%%%%%%%%%%%%%%%%%%%%%%%%%%%%%%%%%%%%%%%%%%%%%%%%%%%%%%%%%%%%%
%%%%%%%%%%%%%%%%%%%%%     			Section 5            %%%%%%%%%%%%%%%%%%%%%%%%
%%%%%%%%%%%%%%%%%%%%%%%%%%%%%%%%%%%%%%%%%%%%%%%%%%%%%%%%%%%%%%%%%%%%%%%
%%%%%%%%%%%%%%%%%%%%%%%%%%%%%%%%%%%%%%%%%%%%%%%%%%%%%%%%%%%%%%%%%%%%%%%
%%%%%%%%%%%%%%%%%%%%%%%%%%%%%%%%%%%%%%%%%%%%%%%%%%%%%%%%%%%%%%%%%%%%%%%
\section{Theoretical results}\label{sec:THEOREM}
In this section, we give seven theorems about accuracy and convergence of the proposed algorithm. 
First, for each ${\bm x} \in \Omega$, we define a misspecification loss at the end of the algorithm as 
\begin{align*}
e_h ({\bm x}) = 
\left \{
\begin{array}{ll}
\max \{ 0,  f({\bm x}) -h \}  & \text{if} \  {\bm x} \in \widehat{L}  \\
\max \{ 0, h- f({\bm x}) \} & \text{if} \ {\bm x} \in \widehat{H} 
\end{array}
\right . .
\end{align*}
Then, the following theorem holds:
\begin{theorem}\label{thm:seido}
For any $h \in \mathbb{R}$,  $\delta \in (0,1)$ and  $\epsilon >0$, if 
$\beta = 2 \log ( |\Omega| \delta^{-1} )$, then 
with probability at least $1-\delta$, the misspecification loss is less than $\epsilon$ 
when   Algorithm \ref{ALG1}  completes classification (i.e., $U_t =\emptyset$).
That is, the following inequality holds:
$$
{\rm P} \left ( \max _{{\bm x} \in \Omega} e_h ({\bm x}) \leq \epsilon \right ) \geq 1-\delta.
$$
\end{theorem}
The proof is given in Appendix \ref{app2}. 
Note that $\epsilon $ and $\beta$ are user-specified input hyper-parameters. 
Next, we consider the convergence of Algorithm \ref{ALG1}.  
Recall that inputs have uncertainty in this paper  unlike the usual BO setting.
Therefore, the desired input point may be greatly different from the actually input point. 
Furthermore, this can happen every trial. 
This implies that  a probabilistic evaluation is needed when we analyze the convergence of the algorithm. 
Hence, in order to make a probabilistic evaluation, we assume the following three conditions:
\begin{description}
\item [(A1)] Probabilities  $\{ p_t \} _ {t \in \N} $ satisfy   $\sum_{t=1}^\infty p_t  = \infty$.
\item [(A2)] For any $ {\bm x} \in \Omega$ and $ \eta >0$, there exists $  {\bm x}' \in \Omega $ and $c_i$ such that  ${\rm P} ( {\bm{S}} ({\bm x}' , c_i ) \in \mathscr {N} ( {\bm x} ; \eta ) ) >0$, where $ \mathscr {N} ( {\bm x} ; \eta ) \equiv \{ {\bm{a}} \in D \ | \ \| {\bm{a}} - {\bm x} \|  <\eta \}$.
\item [(A3)]  For any ${\bm x} \in \Omega$, the kernel function $k $ is continuous at $({\bm x},{\bm x} )$. 
\end{description}
The condition ${\sf (A1)}$ holds when  each $p_t $ is larger than a positive constant $c$. 
Moreover,   ${\sf (A1)}$ holds even if   $ p_t = o( t^{-1} )$. 
The condition ${\sf (A2)}$ 
\footnote{The negation $\lnot ${\sf (A2)}  of {\sf (A2)} is as follows: 
There exists $ {\bm x} \in \Omega$ and $ \eta >0$ such that ${\rm P} ( {\bm{S}} ({\bm x}' , c_i ) \in \mathscr {N} ( {\bm x} ; \eta ) ) =0$ 
for any  $  {\bm x}' \in \Omega $ and $c_i$. 
This implies that regardless of the input cost, no matter what input point ${\bm x}'$ (including ${\bm x}$) is evaluated,  points in   $\eta$-neighborhood  of ${\bm x}$ cannot be observed. 
However, since ${\bm x}$ itself is also a classification target, this claim is nonsense. 
For this reason, we assume the condition {\sf (A2)}  which is the negation of $\lnot ${\sf (A2)}.
} 
requires the existence of  an input $ {\bm x}' \in \Omega $ and a cost $ c_i $ that can take a value around  $ {\bm x} \in \Omega $.
The condition ${\sf (A3)}$ only requires that $k$ is continuous on $\{ ({\bm x},{\bm x}) \ | \ {\bm x} \in \Omega\}$, not $D \times D$.
Thus, ${\sf (A1)}$--${\sf (A3)}$ are mild conditions.
Then, the following theorem holds: 
\begin{theorem}\label{thm:owaru}
Assume that  {\sf (A1)} -- {\sf (A3) }  hold. 
Then, 
 for any $h \in \mathbb{R}$, $\epsilon >0$ and 
$\beta >0 $, with probability 
1, the following holds for any ${\bm x} \in \Omega$:
$$
\sigma^2_t ({\bm x} )    \to 0 \quad  (\text{as} \   t \to \infty ).
$$
Furthermore, with probability 1, 
the number of evaluations of points required to complete Algorithm \ref{ALG1} is finite.
\end{theorem}
The proof is given in  Appendix \ref{app3}. 
Note that Theorem \ref{thm:owaru} describes that if we do not have  the termination condition in the algorithm and keep it running forever,  all posterior variances converge  to zero with probability 1.  
On the other hand, from the classification rule, if the posterior variance of ${\bm x}$ is  less than $\beta^{-1} \epsilon ^2$, it is classified into either one. 
Thus, Theorem \ref{thm:owaru} also states that Algorithm \ref{ALG1}  is actually completed after a finite number of trials with probability 1.
Also note that Theorem \ref{thm:owaru} guarantees that Algorithm \ref{ALG1} completes after a finite number of trials, but does not state its bound.
However, because of the input uncertainty, it is difficult to derive the  bound of  the number of trials required for Algorithm \ref{ALG1}  to complete with probability 1.
For this reason, we derive  theorems that give bounds for Algorithm \ref{ALG1} to complete with high probability.
First, we define several notations. 
For each $ t \geq 1$ and ${\bm x} \in \Omega$, define
\begin{align}
\varsigma^2_t ({\bm x}) = k( {\bm x},{\bm x})  
-(  k({\bm x},{\bm x})    {\bm{1}}_t)^\top   (  k({\bm x},{\bm x})  {\bm{1}}_t {\bm{1}}^\top_t + \sigma^2 {\bm{I}}_t )^{-1} 
  (  k({\bm x},{\bm x})    {\bm{1}}_t) . \label{eq:varsigma}
\end{align}
Here, ${\bm{1}}_t$ is a $t$-dimensional vector where every element is equal to one. 
Hence, 
$\varsigma^2_t ({\bm x})$ is the posterior variance of $f({\bm x})$ 
when ${\bm x}$ is chosen $t$ times. 
Next, for the  pair of  input points $({\bm x} _1 ,\ldots,{\bm x}_t ) \equiv {\bm x}^{(t)} $, we define $\varsigma^2 _{{\bm x}^{(t)}} ({\bm x})$  as 
\begin{align}
\varsigma^2 _{{\bm x}^{(t)}} ({\bm x}) = k({\bm x} , {\bm x}) - {\bm{k}}_{{\bm x}^{(t)}} ({\bm x}) ^\top ( {\bm{K}}_{{\bm x}^{(t)}}+\sigma^2 {\bm{I}}_t )^{-1} {\bm{k}}_{{\bm x}^{(t)}} ({\bm x}), \label{eq:xttsuikavar}
\end{align}
where ${\bm{k}}_{{\bm x}^{(t)}} ({\bm x})$  is a $t$-dimensional vector whose $i$th element is   $k({\bm x}_i, {\bm x} )$, and $ {\bm{K}}_{{\bm x}^{(t)}}$ is 
a $t \times t$ matrix whose 
$(i,j)$th element is 
$k({\bm x}_i,{\bm x}_j)$. Then, the following theorem holds:
\begin{theorem}\label{thm:pro_owaru}
Assume that  {\sf (A1)} -- {\sf (A3) } hold. 
Let $\delta \in (0,1)$ and 
let $T^\ast$ be a smallest positive integer satisfying $T^\ast > 2 \beta \sigma^2 \epsilon ^{-2}$. 
Furthermore, for each ${\bm x} \in \Omega $, define the positive probabilities $\tilde{p}_{\bm x} >0$ and  $\tilde{p}^\ast >0$ as 
\begin{align}
\tilde{p}_{\bm x} =    \sum_{ {\bm{a}} \in \Omega }   \sum_{i=1}^k  \kappa_{ i }   {\rm P} (  {\bm{S}} ({\bm{a}},c_i )  \in \mathscr{N} ({\bm x}; \nu) ) \label{eq:kakuritupx}
\end{align}
and $\tilde{p}^\ast = \min _{{\bm x} \in \Omega} \tilde{p}_{\bm x} $, where 
$\nu $ is a positive number satisfying  
\begin{align}
^\forall {\bm x} \in \Omega, \ ^\forall {\bm{a}}^{(T^\ast)} =({\bm{a}}_1,\ldots, {\bm{a}}_{T^\ast} ) \in \bigotimes_{i=1}^{T^\ast}  \mathscr{N} ({\bm x} ;\nu) ,\quad 
| \varsigma^2_{T^\ast} ({\bm x}) - \varsigma^2_{{\bm{a}}^{(T^\ast)} } ({\bm x}) | < \frac{\epsilon^2}{2 \beta}. \label{eq:tikai}
\end{align}
Here, the notation $\bigotimes$ means the Cartesian product.
Then, there exists non-negative integers $N_0 , \ldots ,  N_{|\Omega|T^\ast}$ such that 
$0=N_0 < N_1 < \cdots <N_{|\Omega| T^\ast}$ and 
\begin{align}
\prod _{t= N_{j-1} +1 } ^{N_j} ( 1- p_t \tilde{p}^\ast )  <  \frac{\delta}{|\Omega| T^\ast   }, \quad (^\forall  j \in [|\Omega| T^\ast]).  \label{eq:probineq}
\end{align}
Moreover, with probability at least $1-\delta$, 
Algorithm \ref{ALG1} completes after at most 
$N_{ |\Omega| T^\ast } $ trials.
\end{theorem}
Here, the existence of $\nu$ satisfying \eqref{eq:tikai} in 
Theorem \ref{thm:pro_owaru} can be guaranteed by 
 {\sf (A3)}. 
Moreover, the positivity of $\tilde{p}_{\bm x}$ can be derived by 
$0< \kappa_{i} <1$ and  {\sf (A2)}. 
Furthermore, the existence of non-negative integers $N_0 , \ldots ,  N_{|\Omega|T^\ast}$ can be obtained by 
  $\tilde{p}^\ast >0$ and 
 {\sf (A1)}. 
 On the other hand, the interpretation of $N_{|\Omega|T^\ast}$ in Theorem \ref{thm:pro_owaru} is not good. 
 In order to solve this problem, we consider the following  {\sf (A1')}:
\begin{description}
\item [(A1')] There exists $p^\ast \in (0,1)$ such that   the inequality $p^\ast  \leq p_t $ holds for any $ t \in \N$.
\end{description}
Note that  {\sf (A1')} is stronger than  {\sf (A1)}. 
Then, the following theorem holds:
\begin{theorem}\label{thm:pro_owaru2}
Assume that 
{\sf (A1'), (A2), (A3)} hold. 
Let $\delta$, $T^\ast$ and $\tilde{p}^\ast $ be numbers defined in Theorem \ref{thm:pro_owaru}, and let 
 $p= p^\ast \tilde{p}^\ast$. In addition, let $r$ be the smallest positive integer satisfying
$$
r > \frac{ \log ( |\Omega| T^\ast \delta ^{-1} )}{- \log (1-p) }.
$$
Then, with probability at least $1-\delta$, Algorithm \ref{ALG1} completes after at most $r|\Omega|T^\ast$ trials.
\end{theorem}
In addition, theorems similar to 
Theorem \ref{thm:pro_owaru} and  \ref{thm:pro_owaru2}   can also be derived under the setting in Subsection \ref{subsec:extend_var}.
Let  $\tilde{\sigma}^2 = \max \{ \sigma^{(1)2},\ldots, \sigma^{(k^\ast)2} \}$, and let 
\begin{align}
\tilde{\varsigma}^2_t ({\bm x}) = k( {\bm x},{\bm x}) -   (  k({\bm x},{\bm x})    {\bm{1}}_t)^\top   (  k({\bm x},{\bm x})  {\bm{1}}_t {\bm{1}}^\top_t + \tilde{\sigma}^2 {\bm{I}}_t )^{-1} 
  (  k({\bm x},{\bm x})    {\bm{1}}_t). \label{eq:varsigmatilde}
\end{align}
Furthermore, for the pair of input points $({\bm x} _1 ,\ldots,{\bm x}_t ) \equiv {\bm x}^{(t)} $, we define $\tilde{\varsigma}^2 _{ {\bm x}^{(t)}} ({\bm x}) $ as 
\begin{align}
\tilde{\varsigma}^2 _{{\bm x}^{(t)}} ({\bm x}) = k({\bm x} , {\bm x}) - {\bm{k}}_{{\bm x}^{(t)}} ({\bm x}) ^\top ( {\bm{K}}_{{\bm x}^{(t)}}+\tilde{\sigma}^2 {\bm{I}}_t )^{-1} {\bm{k}}_{{\bm x}^{(t)}} ({\bm x}). \label{eq:xttsuikavartilde}
\end{align}
Then, the following theorems hold:
\begin{theorem}\label{thm:pro_owaru3}
 Assume the setting in Subsection \ref{subsec:extend_var}. 
Also assume that 
  {\sf (A1)} -- {\sf (A3) } hold. 
Let $\tilde{\delta} \in (0,1)$ and 
let $\tilde{T}^\ast$ be a smallest positive integer satisfying $\tilde{T}^\ast > 2 \beta \tilde{\sigma}^2 \epsilon ^{-2}$. 
In addition, for each ${\bm x} \in \Omega $, define the positive probabilities $\hat{p}_{\bm x} >0$ and $\hat{p}^\ast >0 $ as 
\begin{align}
\hat{p}_{\bm x} =    \sum_{ {\bm{a}} \in \Omega }   \sum_{i=1}^k  \tilde{\kappa}_{ i }   {\rm P} (  {\bm{S}} ({\bm{a}},c_i )  \in \mathscr{N} ({\bm x}; \tilde{\nu}) ) \label{eq:kakuritupx3}
\end{align}
and  $\hat{p}^\ast = \min _{{\bm x} \in \Omega} \hat{p}_{\bm x} $, where  
 $\tilde{\kappa} _{i} = \sum _{j=1}^{k^\ast} \kappa _{i,j}$ and 
  $\tilde{\nu} $ is a positive integer satisfying  
\begin{align}
^\forall {\bm x} \in \Omega, \ ^\forall {\bm{a}}^{(\tilde{T}^\ast)} =({\bm{a}}_1,\ldots, {\bm{a}}_{\tilde{T}^\ast} ) \in \bigotimes_{i=1}^{\tilde{T}^\ast}  \mathscr{N} ({\bm x} ;\tilde{\nu}) ,\quad 
| \tilde{\varsigma}^2_{\tilde{T}^\ast} ({\bm x}) - \tilde{\varsigma}^2_{{\bm{a}}^{(\tilde{T}^\ast)} } ({\bm x}) | < \frac{\epsilon^2}{2 \beta}. \label{eq:tikai3}
\end{align}
Then, there exists positive integers $\tilde{N}_0 , \ldots ,  \tilde{N}_{|\Omega|\tilde{T}^\ast}$ such that 
  $0=\tilde{N}_0 < \tilde{N}_1 < \cdots <\tilde{N}_{|\Omega| \tilde{T}^\ast}$ and 
\begin{align}
\prod _{t= \tilde{N}_{j-1} +1 } ^{\tilde{N}_j} ( 1- p_t \hat{p}^\ast )  <  \frac{\tilde{\delta}}{|\Omega| \tilde{T}^\ast   },\quad ( ^\forall j \in [|\Omega| \tilde{T}^\ast]). \label{eq:probineq3}
\end{align}
Moreover, with probability at least 
  $1-\tilde{\delta}$,    Algorithm \ref{ALG2} completes after at most  $N_{ |\Omega| \tilde{T}^\ast } $ trials. 
\end{theorem}
\begin{theorem}\label{thm:pro_owaru4}
Assume the setting in Subsection \ref{subsec:extend_var}. 
Also assume that 
{\sf (A1'), (A2), (A3)} hold. 
Let $\tilde{\delta}$, $\tilde{T}^\ast$ and $\hat{p}^\ast $ be numbers defined in Theorem \ref{thm:pro_owaru3}, and let 
 $\hat{p}= p^\ast \hat{p}^\ast$. Furthermore, let $\tilde{r}$ be the smallest positive integer satisfying 
$$
\tilde{r} > \frac{\log ( |\Omega| \tilde{T}^\ast \tilde{\delta} ^{-1} )}{ -\log (1-\hat{p})}.
$$
Then, with probability at least 
  $1-\tilde{\delta}$, 
Algorithm \ref{ALG2} completes after at most 
 $\tilde{r}|\Omega|\tilde{T}^\ast$ trials.
\end{theorem}
Finally, the following theorem guarantees that Theorem \ref{thm:owaru}--\ref{thm:pro_owaru4} hold even if input distributions are unknown.
\begin{theorem}\label{thm:owaru5}
Assume that the input distribution ${\bm{S}} ({\bm x},c_i)$ is  unknown. 
Then, 
Theorem \ref{thm:owaru}--\ref{thm:pro_owaru4} hold when ${\bm{S}} ({\bm x},c_i)$ is  estimated such as in Subsection \ref{subsec:extend}. 
In addition,  Theorem \ref{thm:owaru}--\ref{thm:pro_owaru4} hold even if estimated input distributions do not converge to true input distributions, or if misspecified distributions are used without estimation.
\end{theorem}
The proofs of Theorem \ref{thm:pro_owaru}--\ref{thm:owaru5} are given in Appendix \ref{app:lemmaproof}.

%%%%%%%%%%%%%%%%%%%%%%%%%%%%%%%%%%%%%%%%%%%%%%%%%%%%%%%%%%%%%%%%%%%%%%%
%%%%%%%%%%%%%%%%%%%%%%%%%%%%%%%%%%%%%%%%%%%%%%%%%%%%%%%%%%%%%%%%%%%%%%%
%%%%%%%%%%%%%%%%%%%%%%%%%%%%%%%%%%%%%%%%%%%%%%%%%%%%%%%%%%%%%%%%%%%%%%%
%%%%%%%%%%%%%%%%%%%%%     			Section 6            %%%%%%%%%%%%%%%%%%%%%%%%
%%%%%%%%%%%%%%%%%%%%%%%%%%%%%%%%%%%%%%%%%%%%%%%%%%%%%%%%%%%%%%%%%%%%%%%
%%%%%%%%%%%%%%%%%%%%%%%%%%%%%%%%%%%%%%%%%%%%%%%%%%%%%%%%%%%%%%%%%%%%%%%
%%%%%%%%%%%%%%%%%%%%%%%%%%%%%%%%%%%%%%%%%%%%%%%%%%%%%%%%%%%%%%%%%%%%%%%
\section{Numerical experiments}
In this section, we confirm the usefulness of the proposed method through numerical experiments using synthetic  and real data.
The results of numerical experiments not included in this main text are given in Appendix \ref{app-additional}.

\subsection{Synthetic experiments}\label{experiment1}
In this subsection, we compare the proposed method with some existing methods using synthetic functions.

\begin{figure*}[!t]
\begin{center}
\scalebox{1}{
 \begin{tabular}{c}
 \includegraphics[width=1\textwidth]{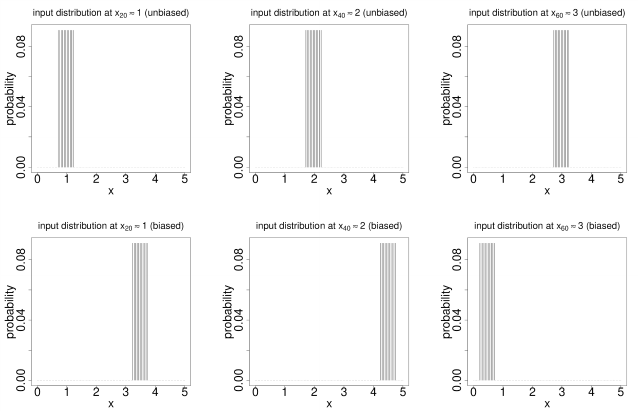} 
 \end{tabular}}
\end{center}

 \caption{
Input distributions for three points $x_{20}, x_{40}, x_{60}$ (left, center and right columns). 
The upper and lower rows represent the unbiased  and   biased cases, respectively. 
} 
 \label{fig:abababa}
\end{figure*}

\paragraph{Usefulness of  the randomized strategy}
First, we confirmed the usefulness of the randomized strategy. 
We considered the function $f(x) =   \cos (\pi x) + \sin (2 \pi x)$ as a true function, and 
defined the grid points obtained by uniformly cutting the region $ [0,5] $ into $ 100 $   as $ \Omega $. 
In addition, we used the Gaussian kernel with 
 $\sigma^2_f = 2 $ and  $L = 0.1$. 
Moreover, we set   $\sigma^2 = 10^{-8} $, 
  $h=0.4$,  $\epsilon = 10^{-12}$ and 
 $\beta^{1/2} =3$. 
 In this experiment, we considered only one cost $c_1 =1$. 
Furthermore, for any $t \geq 1$ and $x \in \Omega$, we used $p_t =0.3$ and $\kappa_{x,1} =0.01$.
In this setting, for each $x \in \Omega =\{ x_i =5(i-1)/99 \mid i=1,\ldots,100 \}$, we considered the following two input distributions:
\begin{description}
\item [(Unbiased)] If $ i \in \{1,\ldots, 5\} \cup \{96,\ldots,100\}$, then $S(x_i, c_1) = x_i$ with probability 1. Similarly, if $i \in \{6,\ldots, 95\}$, then $S(x_i,c_1) = x_j $ with probability $1/11$, where $j \in \{ i-5,\ldots, i+5\}$. 
\item [(Biased)] If $i \in \{46,\ldots, 55\}$, then  $S(x_i, c_1) = x_{\eta(i+50)}$ with probability 1, where $\eta (a) = a$ if $a \leq 100$ and otherwise $\eta (a) = a-100$. Similarly, 
 If $ i \in \{1,\ldots, 45\} \cup \{56,\ldots,100\}$, then $S(x_i, c_1) = x_j $ with probability $1/11$ where $ j \in \{ \eta(i+50) -5 ,\ldots, 
\eta(i+50) +5 \}$.
\end{description}
Note that ${\rm E} [S (x_i ,c_1)] =x_i $ for  the unbiased case, and ${\rm E} [S(x_i,c_1)] = x_{ \eta(i+50) } \neq x_i$ for the biased case. 
Figure \ref{fig:abababa} shows  input distributions at three points $x_{20}, x_{40}, x_{60}$ for two cases.

%%%%%%%%%%%%%%%%%%%%%%%%%%%%%%%%%%%%%%%%%%%%%%%%%%%%%%%%%%%%%%%%%%%%%%%%%%%%%%%%%%%%%%%%
%%%%%%%%%%%%%%%%%%%%%%%%%%%%%%%%%%%%%%%%%%%%%%%%%%%%%%%%%%%%%%%%%%%%%%%%%%%%%%%%%%%%%%%%
%%%%%%%%%%%%%%%%%%%%%%%%%%%%%%%%          Figure 2           %%%%%%%%%%%%%%%%%%%%%%%%%%%%%%%%%%%%
%%%%%%%%%%%%%%%%%%%%%%%%%%%%%%%%%%%%%%%%%%%%%%%%%%%%%%%%%%%%%%%%%%%%%%%%%%%%%%%%%%%%%%%%
%%%%%%%%%%%%%%%%%%%%%%%%%%%%%%%%%%%%%%%%%%%%%%%%%%%%%%%%%%%%%%%%%%%%%%%%%%%%%%%%%%%%%%%%
%%%%%%%%%%%%%%%%%%%%%%%%%%%%%%%%%%%%%%%%%%%%%%%%%%%%%%%%%%%%%%%%%%%%%%%%%%%%%%%%%%%%%%%%

\begin{figure*}[!t]
\begin{center}
\scalebox{1}{
 \begin{tabular}{c}
 \includegraphics[width=1\textwidth]{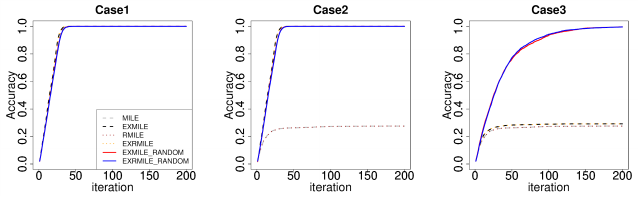} 
 \end{tabular}}
\end{center}

 \caption{
Average accuracy based on 100 Monte Carlo simulations in the 
one dimensional function   $f(x) =   \cos (\pi x) + \sin (2 \pi x)$. 
The  left, center and right figures represent the Case 1, 2 and 3, respectively.
}
 \label{fig:robust}
\end{figure*}

Then, we compared the following six acquisition functions: 
\begin{description}
\item [(MILE)] Compute MILE acquisition function without integrating against input distribution. 
\item [(RMILE)] Compute robust MILE acquisition function (RMILE) without integrating against input distribution, where we used $\gamma =0.01$ as the robustness parameter. 
\item [(EXMILE)] Compute MILE acquisition function with integrating against input distribution. 
\item [(EXRMILE)] Compute RMILE acquisition function  with integrating against input distribution, where we used $\gamma =0.01$ as the robustness parameter. 
\item [(EXMILE\_RANDOM)] Compute EXMILE  with randomized probability $p_t =0.3$. 
\item [(EXRMILE\_RANDOM)] Compute EXRMILE with randomized probability $p_t =0.3$.
\end{description}
Under this setting, we considered the following three cases:
\begin{description}
\item [(Case 1)] Unbiased input distribution is used as the true input distribution, 
and it is also used for calculating EXMILE (EXRMILE) and EXMILE\_RANDOM (EXRMILE\_RANDOM).
\item [(Case 2)] Biased input distribution is used as the true input distribution, 
and it is also used for calculating EXMILE (EXRMILE) and EXMILE\_RANDOM  (EXRMIL\-E\_RANDOM).
\item [(Case 3)] Biased input distribution is used as the true input distribution, but 
the unbiased input distribution  is  used for calculating EXMILE (EXRMILE) and  EXMILE\_RA\-NDOM (EXRMILE\_RANDOM).
\end{description}
Note that in case 3, the input distribution is misspecified.
Then, one initial point was taken at random, and points were acquired until the total cost (iteration) reached 200. The classification performance was evaluated using   the following accuracy:
$$
\text{Accuracy} =  \frac{ | H \cap  (H_t \setminus L_t ) | + | L \cap  (L_t \setminus H_t ) |       }{| \Omega |}.
$$
The average obtained by 100 Monte Carlo simulations is given in Figure \ref{fig:robust}.

%%%%%%%%%%%%%%%%%%%%%%%%%%%%%%%%%%%%%%%%%%%%%%%%%%%%%%%%%%%%%%%%%%%%%%%%%%%%%%%%%%%%%%%%
%%%%%%%%%%%%%%%%%%%%%%%%%%%%%%%%%%%%%%%%%%%%%%%%%%%%%%%%%%%%%%%%%%%%%%%%%%%%%%%%%%%%%%%%
%%%%%%%%%%%%%%%%%%%%%%%%%%%%%%%%          Figure 3           %%%%%%%%%%%%%%%%%%%%%%%%%%%%%%%%%%%%
%%%%%%%%%%%%%%%%%%%%%%%%%%%%%%%%%%%%%%%%%%%%%%%%%%%%%%%%%%%%%%%%%%%%%%%%%%%%%%%%%%%%%%%%
%%%%%%%%%%%%%%%%%%%%%%%%%%%%%%%%%%%%%%%%%%%%%%%%%%%%%%%%%%%%%%%%%%%%%%%%%%%%%%%%%%%%%%%%
%%%%%%%%%%%%%%%%%%%%%%%%%%%%%%%%%%%%%%%%%%%%%%%%%%%%%%%%%%%%%%%%%%%%%%%%%%%%%%%%%%%%%%%%

\begin{figure*}[!t]
\begin{center}
\scalebox{0.85}{
 \begin{tabular}{c}
 \includegraphics[width=1\textwidth]{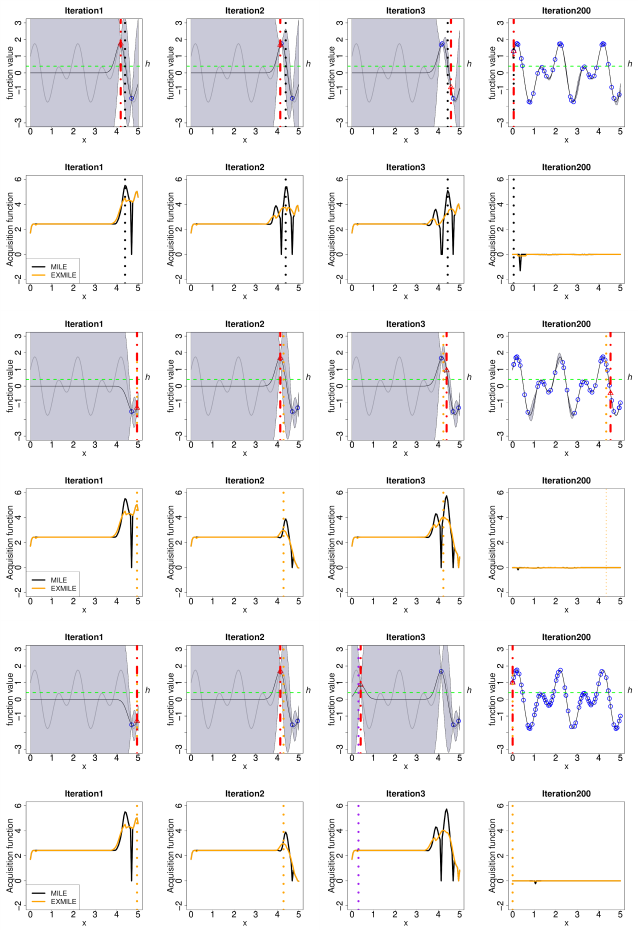} 
 \end{tabular}}
\end{center}
 \caption{
 Behaviors of each method in Case 1. 
The first (second), third (fourth) and fifth (sixth) rows represent MILE, EXMILE and EXMILE\_RANDOM, respectively. 
The black, orange and purple dashed lines represent the selected input points by MILE, EXMILE and randomly, and the red dashed lines mean  the actually evaluated points. 
}
 \label{fig:kyodoA}
\end{figure*}

\begin{figure*}[!t]
\begin{center}
\scalebox{0.85}{
 \begin{tabular}{c}
 \includegraphics[width=1\textwidth]{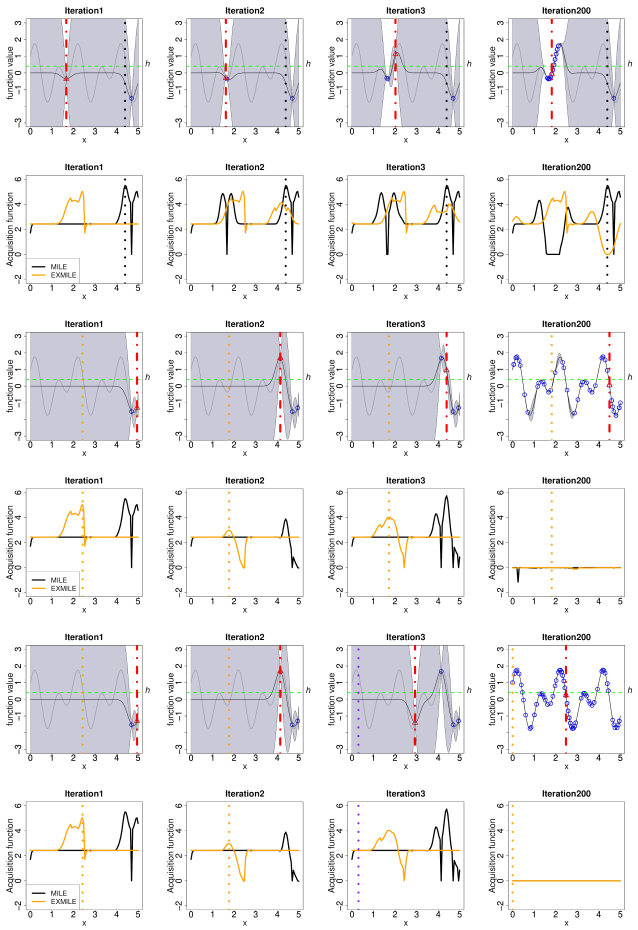} 
 \end{tabular}}
\end{center}
\caption{
 Behaviors of each method in Case 2. 
The first (second), third (fourth) and fifth (sixth) rows represent MILE, EXMILE and EXMILE\_RANDOM, respectively. 
The black, orange and purple dashed lines represent the selected input points by MILE, EXMILE and randomly, and the red dashed lines mean  the actually evaluated points. 
}
 \label{fig:kyodoB}
\end{figure*}

\begin{figure*}[!t]
\begin{center}
\scalebox{0.85}{
 \begin{tabular}{c}
 \includegraphics[width=1\textwidth]{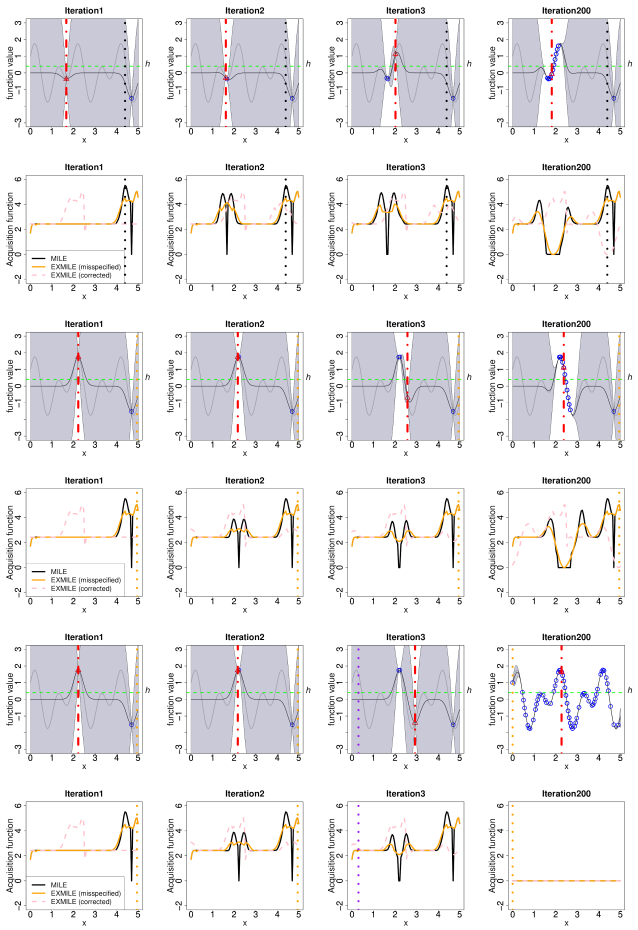}
 \end{tabular}}
\end{center}
 \caption{
 Behaviors of each method in Case 3. 
The first (second), third (fourth) and fifth (sixth) rows represent MILE, EXMILE and EXMILE\_RANDOM, respectively. 
The black, orange and purple dashed lines represent the selected input points by MILE, EXMILE and randomly, and the red dashed lines mean  the actually evaluated points. 
}
 \label{fig:kyodoC}
\end{figure*}

From Figure \ref{fig:robust}, it can be confirmed that in Case 1, all methods  perform well.
In particular, MILE and RMILE, which do not consider the input distribution,  also worked well.
This is because the input distribution is unbiased, and the difference in whether to consider input uncertainty is not too large.
On the other hand, in case 2, MILE and RMILE did not perform well because the input distribution is biased and it may keep on selecting  the same point. 
In contrast, other four methods considering input uncertainty worked well.
Finally, in case 3, where the input distribution is misspecified, the four methods without randomized strategy did not work well.
The reason is that  the input distribution is misspecified, and as a result EXMILE and EXRMILE may also keep to select the same point.
On the other hand, EXMILE\_RANDOM and EXRMILE\_RANDOM  worked well because the problem of taking the same point can be avoided by using the randomized strategy.
The behaviors of MILE, EXMILE and EXMILE\_RANDOM in each case are given in Figure \ref{fig:kyodoA}-\ref{fig:kyodoC}.
The behaviors of RMILE, EXRMILE and EXRMILE\_RANDOM are omitted because they are almost the same as those of MILE, EXMILE and EXMILE\_RANDOM, respectively.
Here, \cite{zanette2018robust} showed by numerical experiments that the difference between MILE and RMILE  is small when the error variance $\sigma^2$ is small. 
Moreover, in this experiment, we used $\sigma^2 =10^{-8}$. 
This is the reason why the performances of MILE, EXMILE and EXMILE\_RANDOM are almost the same as those of  RMILE, EXRMILE and EXRMILE\_RANDOM, respectively. 
On the other hand, under no input uncertainty cases, \cite{zanette2018robust} also showed by experimentally that the performance of RMILE is better than that of MILE when the error variance is large and its value is misspecified. 
However, we compared (EX)MILE and (EX)RMILE under various settings (including this setting) and found that there was almost no difference between them when inputs have uncertainty.
For this reason, in the subsequent experiments we omitted the results of (EX)RMILE.

%%%%%%%%%%%%%%%%%%%%%%%%%%%%%%%%%%%%%%%%%%%%%%%%%%%%%%%%%%%%%%%%%%%%%%%%%%%%%%%%%%%%%%%%
%%%%%%%%%%%%%%%%%%%%%%%%%%%%%%%%%%%%%%%%%%%%%%%%%%%%%%%%%%%%%%%%%%%%%%%%%%%%%%%%%%%%%%%%
%%%%%%%%%%%%%%%%%%%%%%%%%%%%%%%%          Figure 4           %%%%%%%%%%%%%%%%%%%%%%%%%%%%%%%%%%%%
%%%%%%%%%%%%%%%%%%%%%%%%%%%%%%%%%%%%%%%%%%%%%%%%%%%%%%%%%%%%%%%%%%%%%%%%%%%%%%%%%%%%%%%%
%%%%%%%%%%%%%%%%%%%%%%%%%%%%%%%%%%%%%%%%%%%%%%%%%%%%%%%%%%%%%%%%%%%%%%%%%%%%%%%%%%%%%%%%
%%%%%%%%%%%%%%%%%%%%%%%%%%%%%%%%%%%%%%%%%%%%%%%%%%%%%%%%%%%%%%%%%%%%%%%%%%%%%%%%%%%%%%%%

%%%%%%%%%%%%%%%%%%%%%%%%%%%%%%%%%%%%%%%%%%%%%%%%%%%%%%%%%%%%%%%%%%%%%%%%%%%%%%%%%%%%%%%%
%%%%%%%%%%%%%%%%%%%%%%%%%%%%%%%%%%%%%%%%%%%%%%%%%%%%%%%%%%%%%%%%%%%%%%%%%%%%%%%%%%%%%%%%
%%%%%%%%%%%%%%%%%%%%%%%%%%%%%%%%          Figure 5           %%%%%%%%%%%%%%%%%%%%%%%%%%%%%%%%%%%%
%%%%%%%%%%%%%%%%%%%%%%%%%%%%%%%%%%%%%%%%%%%%%%%%%%%%%%%%%%%%%%%%%%%%%%%%%%%%%%%%%%%%%%%%
%%%%%%%%%%%%%%%%%%%%%%%%%%%%%%%%%%%%%%%%%%%%%%%%%%%%%%%%%%%%%%%%%%%%%%%%%%%%%%%%%%%%%%%%
%%%%%%%%%%%%%%%%%%%%%%%%%%%%%%%%%%%%%%%%%%%%%%%%%%%%%%%%%%%%%%%%%%%%%%%%%%%%%%%%%%%%%%%%

%%%%%%%%%%%%%%%%%%%%%%%%%%%%%%%%%%%%%%%%%%%%%%%%%%%%%%%%%%%%%%%%%%%%%%%%%%%%%%%%%%%%%%%%
%%%%%%%%%%%%%%%%%%%%%%%%%%%%%%%%%%%%%%%%%%%%%%%%%%%%%%%%%%%%%%%%%%%%%%%%%%%%%%%%%%%%%%%%
%%%%%%%%%%%%%%%%%%%%%%%%%%%%%%%%          Figure 6           %%%%%%%%%%%%%%%%%%%%%%%%%%%%%%%%%%%%
%%%%%%%%%%%%%%%%%%%%%%%%%%%%%%%%%%%%%%%%%%%%%%%%%%%%%%%%%%%%%%%%%%%%%%%%%%%%%%%%%%%%%%%%
%%%%%%%%%%%%%%%%%%%%%%%%%%%%%%%%%%%%%%%%%%%%%%%%%%%%%%%%%%%%%%%%%%%%%%%%%%%%%%%%%%%%%%%%
%%%%%%%%%%%%%%%%%%%%%%%%%%%%%%%%%%%%%%%%%%%%%%%%%%%%%%%%%%%%%%%%%%%%%%%%%%%%%%%%%%%%%%%%

\paragraph{Sinusoidal function}
We considered the function $ f (x_1, x_2) = \sin (10x_1) + \cos (4x_2)-\cos (3x_1x_2) $ which was used in \cite{bryan2006active} as  a true function, 
and defined the grid point obtained by uniformly cutting the region $ [0,1] \times [0,2] $ into $ 30 \times 60 $   as $ \Omega $. 
In addition, we used the Gaussian kernel with 
 $\sigma^2_f = e^2 $ and  $L = 2 e^{-3}$. 
Moreover, we set   $\sigma^2 = e^{-2} $, 
  $h=1$,  $\epsilon = 10^{-12}$ and 
 $\beta^{1/2} =1.96$. 

In this experiment, we considered three costs  $c_1 =1$, $c_2 =2$ and $c_3 =3$. 
For each $c_i$ and $ {\bm x} =(x_1,x_2 )^\top \in \Omega$,  we defined the input distribution as 
$$
{\bm{S}} ({\bm x} , c_i ) =  {\bm x} +   (  G_{[0,1-x_1] } ( \zeta^{(i)},1) ,    G_{[0,2-x_2] } (\zeta^{(i)},1)  ) ^\top .
$$
Here, $G_{[0,a] } ( b,c )$ is a   gamma distribution with   parameters $b$ and $c$ which is restricted on    the interval  $[0,a]$. 
We assumed that  $G_{[0,1-x_1] } ( \zeta^{(i)},1) $ and $   G_{[0,2-x_2] } (\zeta^{(i)},1)$ are independent. 
Furthermore, we used  that 
$\zeta^{(1)} =4$ and $\zeta^{(2)} =1 $, $\zeta^{(3)} =0.01$.

\begin{figure*}[!t]
\begin{center}
\scalebox{1}{
 \begin{tabular}{c}
 \includegraphics[width=1\textwidth]{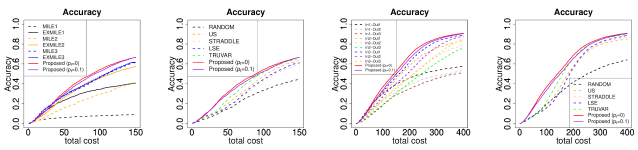} 
 \end{tabular}}
\end{center}
 \caption{
Average accuracy based on 20 Monte Carlo simulations in the 
Sinusoidal function (first and second columns) and Rosenbrock function (third and fourth columns).  
The  first figure shows the influence of integration against  the input distribution and that of cost in evaluating the input point. 
The second  figure shows the result of comparison with existing methods. 
The third figure  shows the influence of integration against  the input distribution and that of costs for evaluating the input point  and evaluating the function value. 
The fourth figure shows the result of comparison with existing methods.
}
 \label{fig:exp1}
\end{figure*}

Then,       we compared the following eight  methods $(i=1,2,3,\ a=0,0.1)$:
\begin{description}
\item [(MILE$i$)] Always take input points using cost $i$. In addition, the acquisition function is calculated without integrating against input distribution. 
\item [(EXMILE$i$)] Always take input points using cost $i$. In addition, the acquisition function is calculated with  integrating against input distribution. 
\item [(Proposed ($p_t =a$))] All costs are allowed, and $ a_t ({\bm x}, c_i) $ is used as the acquisition function. 
In addition, we set $p_t =a$.
\end{description}
In order to calculate integrals, we used the Monte Carlo approximation (details are given in  
  \eqref{eq:approx2}). 
Moreover,  
 to estimate the discrete distribution, $\tilde{ {\bm{S}} } ({\bm x},c_i)$ was estimated by generating independent samples from each ${\bm{S}} ({\bm x},c_i)$ thousand times. 
Note that the acquisition function used in MILE$i$ is the same as the original MILE acquisition function  proposed by \cite{zanette2018robust}.
Under this setting, one initial point was taken at random, and points were acquired until the total cost reached 150. The classification performance was evaluated using    accuracy. 
The average obtained by 20 Monte Carlo simulations is given in Figure \ref{fig:exp1}. 
From the leftmost figure of Figure  \ref{fig:exp1}, we can confirm that 
it is important to integrate against the input distribution  when calculating the acquisition function. 
We can also see  that the red and purple lines (proposed methods) that appropriately select the cost at each trial have higher  accuracy compared with other methods. 
Next, we compared with the following existing methods:
\begin{description}
\item [(RANDOM)] Perform random sampling. 
\item [(US)] Perform uncertainty sampling, i.e., we select the input point with the largest posterior variance. 
\item [(STRADDLE)] Perform straddle strategy \cite{bryan2006active}, where we used $\beta^{1/2}_t =1.96$. 
  \item [(LSE)] Perform LSE strategy \cite{Gotovos:2013:ALL:2540128.2540322}, where we used $\beta^{1/2}_t =1.96$. 
\item [(TRUVAR)] Perform TRUVAR strategy \cite{NIPS2016_6080}, where we used $0$, $0.1$, $1.96$ and $1$ as  parameters 
$\bar{\delta} $, $r$, $\beta_{ (i)} $ and $\eta_{(1)} $ in TRUVAR, respectively.
\end{description}
In this experiment, US, STRADDLE,  LSE and TRUVAR were tested in advance in the same way as the proposed method with a total of seven types including the presence or absence of integration against the input distribution and the presence or absence of cost sensitive. 
Among them, the one with the highest accuracy was used for comparison. 
Similarly, for RANDOM, we tried a total of four types with or without cost sensitive and used the best results for comparison.
From the  second  from the left in  
Figure \ref{fig:exp1}, we can confirm that the proposed methods have higher accuracy than other existing methods. 
Note that MILE focuses on maximizing the  expected classification improvement when the new point is added.
On the other hand, acquisition functions except  for MILE are not derived based on the improvement in the number of classifications.
This is the reason why proposed methods have higher accuracy compared with other methods. 
Moreover, since the input distribution is correctly estimated in this experiment, the accuracy of Proposed ($p_t=0$) is higher than that of Proposed ($p_t=0.1$).

%%%%%%%%%%%%%%%%%%%%%%%%%%%%%%%%%%%%%%%%%%%%%%%%%%%%%%%%%%%%%%%%%%%%%%%%%%%%%%%%%%%%%%%%
%%%%%%%%%%%%%%%%%%%%%%%%%%%%%%%%%%%%%%%%%%%%%%%%%%%%%%%%%%%%%%%%%%%%%%%%%%%%%%%%%%%%%%%%
%%%%%%%%%%%%%%%%%%%%%%%%%%%%%%%%          Figure 7           %%%%%%%%%%%%%%%%%%%%%%%%%%%%%%%%%%%%
%%%%%%%%%%%%%%%%%%%%%%%%%%%%%%%%%%%%%%%%%%%%%%%%%%%%%%%%%%%%%%%%%%%%%%%%%%%%%%%%%%%%%%%%
%%%%%%%%%%%%%%%%%%%%%%%%%%%%%%%%%%%%%%%%%%%%%%%%%%%%%%%%%%%%%%%%%%%%%%%%%%%%%%%%%%%%%%%%
%%%%%%%%%%%%%%%%%%%%%%%%%%%%%%%%%%%%%%%%%%%%%%%%%%%%%%%%%%%%%%%%%%%%%%%%%%%%%%%%%%%%%%%%

\paragraph{Two-dimensional Rosenbrock function with cost dependent noise variance}
Here, we  considered the  2-dimensional Rosenbrock function (reduced to 1/100 and moved)  
$
f(x_1,x_2)=(x_2-x^2_1)^2+(1-x_1)^2/100 -5
$
as the true function, and defined the grid point obtained by uniformly cutting the region  $[-2,2] \times [-1,3]$ into $40 \times 40$ as $\Omega$. 
Furthermore, we used the Gaussian kernel 
with  $\sigma^2_f = 64 $ and  $L = 0.5$. In addition, we set $\sigma^2 = 0.25$, 
$h=0$, $\epsilon = 10^{-12}$ and 
  $\beta^{1/2} =1.96$.  
Similarly in this experiment, we considered three costs   $c_1 =1$, $c_2 =2$ and $c_3 =3$. 
Moreover, for each $c_i$ and $ {\bm x} =(x_1,x_2 )^\top \in \Omega$,  we assumed that 
$$
{\bm{S}} ({\bm x} , c_i ) =  {\bm x} +   (  G_{[0,2-x_1] } ( \zeta^{(i)},1) ,    G_{[0,3-x_2] } (\zeta^{(i)},1)  ) ^\top , 
$$
where $G_{[0,2-x_1] } ( \zeta^{(i)},1) $ and $   G_{[0,3-x_2] } (\zeta^{(i)},1)$ are independent. 
Furthermore, we used  
$\zeta^{(1)} =4$, $\zeta^{(2)} =1 $, $\zeta^{(3)} =0.01$. 
Moreover, 
 we considered the situation where the noise in the output also changes according to the cost. 
In this experiment, we considered three output costs $c^{(\text{out})}_1 =1$, $c^{(\text{out})}_2 =2$ and $c^{(\text{out})}_3 =3$, and 
then we defined $\varrho ^{(j)} \sim \mathcal{N} (0, \sigma^{(j) 2}  )$ as the error distribution, where 
  $ j \in\{1,2,3\}$. 
Furthermore, we set that 
$\sigma^{(1)2}  =0.5$, $\sigma^{(2)2}  =0.3$ and $\sigma^{(3)2}  =0.1$. 
Then, we compared the following eleven methods ($i,j \in \{1,2,3 \}, a=0,0.1$):     
\begin{description}
\item [(In$i$-Out$j$)] Take the input point using the cost $ i $ and observe the function value using the cost $ j $. In addition, in the calculation of the acquisition function, integration is performed against the input distribution.
\item [(Proposed ($p_t =a$))] All costs are allowed, and the acquisition function is calculated by   \eqref {eq:AFteigiver2}.
 In addition, we set $p_t =a$. 
\end{description}
Under this setting,  we performed the similar experiment as in sinusoidal function until the total cost reached 400. 
From the two figures on the right in Figure \ref{fig:exp1}, 
even when the output   has   the cost-dependent error variance, 
we can see that the proposed methods have higher accuracy than the other methods.

\subsection{Real data experiment}
We conducted a real data experiment using the Rhodopsin-family protein data set \footnote{The same dataset was recently used in \cite{inatsu2020b}.} provided by 
 \cite {karasuyama2018understanding}.
Rhodopsin-family proteins have a function to absorb a light with certain wavelength, and this function is effectively used in 
optogenetics \cite{deisseroth2015optogenetics}.
The goal of this experiment is to estimate the level set in the protein feature space in which the absorption wavelength is sufficiently large for optogenetics usage. 
This dataset contains 677 proteins, where each protein $ i $ has a 210-dimensional amino acids sequence vector   and a scalar absorption wavelength output. 
We first constructed a Bayesian linear model using amino acid sequences, modeled the relationship between amino acid sequences and absorption wavelengths, and conducted experiments using this model as the oracle model.
In the experiment, 400 pseudo-proteins were constructed by changing the amino acids of the 150th and 200th residues of the 338th rhodopsin, which has an intermediate absorption wavelength, to 20 different amino acids.
The absorption wavelength of this protein was determined based on the constructed prediction model, and this was set as $ y_ {i, j} $, $ i, j \in [20] $, where the average of $ y_ {i, j} $ was standardized to be 0. 
Here, $(i,j)$ in $ y_ {i, j} $ means that the 150th residue is changed to the $ i $ th amino acid and the 200th residue is changed to the $ j $ th amino acid. 
In addition, the $i$th amino acid means the $i$th amino acid when the one-letter code of the amino acid is arranged in alphabetical order.
As an input corresponding to the response variable $ y_ {i, j} $, we used a 42-dimensional feature vector $ {\bm {x}} _ {i, j} = ({\bm {x}} ^ \top_i, {\bm {x}} ^ \top_j) ^ \top $ consisting of amino acid features (e.g., volume, molecular weight), where   ${\bm{x}} _i , {\bm{x}}_j \in \mathbb{R}^{21}$.

\begin{figure*}[!t]
\begin{center}
\scalebox{1}{
 \begin{tabular}{c}
 \includegraphics[width=1\textwidth]{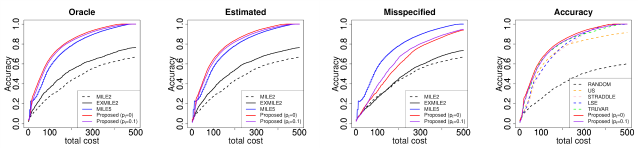} 
 \end{tabular}}
\end{center}
 \caption{
Average accuracy based on 50 Monte Carlo simulations in the   
Rhodopsin Data Set. 
 The first, second and third  figures  show the influences of cost-sensitive and input distribution estimation. 
The fourth figure shows the result of comparison with existing methods.}
 \label{fig:exp3}
\end{figure*}

We assumed that the true output value $y_{i,j} $  can be observed without any noise.   
However, for convenience of calculation, we used $\sigma^2= 10^{-6}$. 
Furthermore, we defined the input domain as $\Omega = \{ {\bm{x}}_{i,j} \ | \ i,j \in [20] \}$. 
We used the Gaussian kernel with  $\sigma^2_f = 10$ and $L = 200$.
In addition, we set $h= 0$, $\epsilon = 10^{-12}$ and   $\beta^{1/2} =3$. 
In this experiment, we considered two costs  $c_1 =2$ and $c_2 =5$. 
Then, for input distributions, we assumed the following synthetic discrete distribution ${\bm{S}} ({\bm{x}}_{u} , c_k )$:
\begin{align*}
&{\rm P} ( {\bm{S}} ({{\bm x}}_u , c_1 )  = {\bm{x}}_v )    =    0.8/3 \quad  (v=7,9,15), \\
&{\rm P} ( {\bm{S}} ({{\bm x}}_u , c_1 )  = {\bm{x}}_v )    =    0.1/2 \quad  (v=3,4), \\ 
&{\rm P} ( {\bm{S}} ({{\bm x}}_u , c_1 )  = {\bm{x}}_v )    =    0.1/15 \quad  (v \in [20] \setminus \{3,4,7,9,15 \}),\\
&{\rm P} ( {\bm{S}} ({{\bm x}}_u , c_2 )  = {\bm{x}} _u )    =    1 .
\end{align*}
In other words, $ {\bm {S}} ({{\bm x}} _u, c_2) $ takes  $ {\bm {x}} _ u $ with probability 1. 
Moreover, $ {\bm {S}} ({{\bm x}} _u, c_1) $ is a random mutation where the probability which takes each acidic, basic and neutral  amino acid are  $0.8/3$, $0.1/2$  and  $0.1/15$, respectively.
Therefore, it is a mutation that easily becomes acidic amino acids.
Based on these, we defined the  input distributions as
\begin{align*}
& {\bm{S}} ({{\bm x}}_{u,v} , c_1 )  = ( {\bm{S}} ({{\bm x}}_u , c_1 ) ,   {\bm{S}} ({{\bm x}}_v , c_1 )    ) ^\top , \\
& {\bm{S}} ({{\bm x}}_{u,v} , c_1 )  = ( {\bm{S}} ({{\bm x}}_u , c_1 ) ,   {\bm{x}}_v    ) ^\top \quad (\text{if \ there \ exists} \ y_{\cdot,v} )  , \\ 
& {\bm{S}} ({{\bm x}}_{u,v} , c_1 )  = ({{\bm x}}_u ,   {\bm{S}} ({{\bm x}}_v , c_1 )     ) ^\top \quad (\text{if \ there \ exists} \ y_{u,\cdot} ) .
\end{align*}
Similarly, we defined 
\begin{align*}
& {\bm{S}} ({{\bm x}}_{u,v} , 2c_2 )  = ({\bm{x}}_u ,  {\bm{x}}_v    ) ^\top \quad (\text{if \ there \ are \ no} \ y_{u,\cdot},y_{\cdot,v} )  \\
& {\bm{S}} ({{\bm x}}_{u,v} , c_2 )  = ({\bm{x}}_u ,   {\bm{x}}_v    ) ^\top \quad (\text{if \ there \ exists} \ y_{\cdot,v} \ \text{or} \ y_{u,\cdot}).
\end{align*}
Under this setting, we considered the following three cases for $ {\bm {S}} ({{\bm x}} _u, c_1) $:
1)  the true distribution is known. 
2) when estimating using categorical distribution and using Dirichlet distribution whose parameter is ${\bm\alpha } = (0.75,0.5,3.75)^\top$ as prior distribution.
3) when the discrete uniform distribution is used without estimation, i.e., the distribution is misspecified. 
Then,  we performed the similar experiment as in sinusoidal function until the total cost reached 500. 
The average obtained by 50 Mote Carlo simulations is given in 
Figure \ref{fig:exp3}. 
From Figure \ref {fig:exp3}, we can confirm   that the proposed methods have higher accuracy  except for the misspecified case. 
Moreover, from the first and second figures in Figure  \ref {fig:exp3}, even if the distribution is unknown, it can be confirmed that its    performance is almost the same as oracle by estimating distribution parameters under the assumption that the true distribution form is known. 
In addition, 
since the input distribution is misspecified, 
   the accuracy of Proposed ($p_t =0.1$) is higher than that of  Proposed ($p_t =0$) 
in the misspecified setting. 
Furthermore, from the rightmost figure in Figure \ref {fig:exp3}, it can be seen that the accuracy of  Proposed ($p_t=0$) is the highest.

%%%%%%%%%%%%%%%%%%%%%%%%%%%%%%%%%%%%%%%%%%%%%%%%%%%%%%%%%%%%%%%%%%%%%%%%%%%%%%%%%%%%%%%%
%%%%%%%%%%%%%%%%%%%%%%%%%%%%%%%%%%%%%%%%%%%%%%%%%%%%%%%%%%%%%%%%%%%%%%%%%%%%%%%%%%%%%%%%
%%%%%%%%%%%%%%%%%%%%%%%%%%%%%%%%          Figure 8           %%%%%%%%%%%%%%%%%%%%%%%%%%%%%%%%%%%%
%%%%%%%%%%%%%%%%%%%%%%%%%%%%%%%%%%%%%%%%%%%%%%%%%%%%%%%%%%%%%%%%%%%%%%%%%%%%%%%%%%%%%%%%
%%%%%%%%%%%%%%%%%%%%%%%%%%%%%%%%%%%%%%%%%%%%%%%%%%%%%%%%%%%%%%%%%%%%%%%%%%%%%%%%%%%%%%%%
%%%%%%%%%%%%%%%%%%%%%%%%%%%%%%%%%%%%%%%%%%%%%%%%%%%%%%%%%%%%%%%%%%%%%%%%%%%%%%%%%%%%%%%%

%%%%%%%%%%%%%%%%%%%%%%%%%%%%%%%%%%%%%%%%%%%%%%%%%%%%%%%%%%%%%%%%%%%%%%%
%%%%%%%%%%%%%%%%%%%%%%%%%%%%%%%%%%%%%%%%%%%%%%%%%%%%%%%%%%%%%%%%%%%%%%%
%%%%%%%%%%%%%%%%%%%%%%%%%%%%%%%%%%%%%%%%%%%%%%%%%%%%%%%%%%%%%%%%%%%%%%%
%%%%%%%%%%%%%%%%%%%%%     			Section 7            %%%%%%%%%%%%%%%%%%%%%%%%
%%%%%%%%%%%%%%%%%%%%%%%%%%%%%%%%%%%%%%%%%%%%%%%%%%%%%%%%%%%%%%%%%%%%%%%
%%%%%%%%%%%%%%%%%%%%%%%%%%%%%%%%%%%%%%%%%%%%%%%%%%%%%%%%%%%%%%%%%%%%%%%
%%%%%%%%%%%%%%%%%%%%%%%%%%%%%%%%%%%%%%%%%%%%%%%%%%%%%%%%%%%%%%%%%%%%%%%
\section{Conclusion}
In this paper, we proposed a new AL method for LSE under 
input uncertainty. 
We also extended the proposed method to two cases:  the cost-dependent input/output  uncertainty case, and  the unknown input distribution case.
The acquisition function in the proposed method is based on  
the integral of the expected increase in classification per unit cost and 
the randomized strategy.
The usefulness of the proposed method was confirmed through both numerical experiments and theoretical analysis. 
%

%%%%%%%%%%%%%%%%%%%%%%%%%%%%%%%%%%%%%%%%%%%%%%%%%%%%%%%%%%%%%%%%%%%%%%%
%%%%%%%%%%%%%%%%%%%%%%%%%%%%%%%%%%%%%%%%%%%%%%%%%%%%%%%%%%%%%%%%%%%%%%%
%%%%%%%%%%%%%%%%%%%%%%%%%%%%%%%%%%%%%%%%%%%%%%%%%%%%%%%%%%%%%%%%%%%%%%%
%%%%%%%%%%%%%%%%%%%%%     			Acknowledgments            %%%%%%%%%%%%%%%%%
%%%%%%%%%%%%%%%%%%%%%%%%%%%%%%%%%%%%%%%%%%%%%%%%%%%%%%%%%%%%%%%%%%%%%%%
%%%%%%%%%%%%%%%%%%%%%%%%%%%%%%%%%%%%%%%%%%%%%%%%%%%%%%%%%%%%%%%%%%%%%%%
%%%%%%%%%%%%%%%%%%%%%%%%%%%%%%%%%%%%%%%%%%%%%%%%%%%%%%%%%%%%%%%%%%%%%%%

\subsection*{Acknowledgments}
This work was supported by MEXT KAKENHI to I.T. (16H06538, 17H00758) and M.K. (16H06538, 17H04694); from JST CREST awarded to I.T. (JPMJCR1302, JPMJCR1502) and PRESTO
awarded to M.K. (JPMJPR15N2); from the MI2I project
of the Support Program for Starting Up Innovation Hub from JST awarded to I.T., and M.K.; and from
RIKEN Center for AIP awarded to  I.T.

%%%%%%%%%%%%%%%%%%%%%%%%%%%%%%%%%%%%%%%%%%%%%%%%%%%%%%%%%%%%%%%%%%%%%%%
%%%%%%%%%%%%%%%%%%%%%%%%%%%%%%%%%%%%%%%%%%%%%%%%%%%%%%%%%%%%%%%%%%%%%%%
%%%%%%%%%%%%%%%%%%%%%%%%%%%%%%%%%%%%%%%%%%%%%%%%%%%%%%%%%%%%%%%%%%%%%%%
%%%%%%%%%%%%%%%%%%%%%     			Appendix                        %%%%%%%%%%%%%%%%%
%%%%%%%%%%%%%%%%%%%%%%%%%%%%%%%%%%%%%%%%%%%%%%%%%%%%%%%%%%%%%%%%%%%%%%%
%%%%%%%%%%%%%%%%%%%%%%%%%%%%%%%%%%%%%%%%%%%%%%%%%%%%%%%%%%%%%%%%%%%%%%%
%%%%%%%%%%%%%%%%%%%%%%%%%%%%%%%%%%%%%%%%%%%%%%%%%%%%%%%%%%%%%%%%%%%%%%%

\section*{Appendix}
\setcounter{section}{0}
\renewcommand{\thesection}{\Alph{section}}
\numberwithin{equation}{section}

%%%%%%%%%%%%%%%%%%%%%%%%%%%%%%%%%%%%%%%%%%%%%%%%%%%%%%%%%%%%%%%%%%%%%%%
%%%%%%%%%%%%%%%%%%%%%%%%%%%%%%%%%%%%%%%%%%%%%%%%%%%%%%%%%%%%%%%%%%%%%%%
%%%%%%%%%%%%%%%%%%%%%%%%%%%%%%%%%%%%%%%%%%%%%%%%%%%%%%%%%%%%%%%%%%%%%%%
%%%%%%%%%%%%%%%%%%%%%     			Appendix        A                %%%%%%%%%%%%%%%%%
%%%%%%%%%%%%%%%%%%%%%%%%%%%%%%%%%%%%%%%%%%%%%%%%%%%%%%%%%%%%%%%%%%%%%%%
%%%%%%%%%%%%%%%%%%%%%%%%%%%%%%%%%%%%%%%%%%%%%%%%%%%%%%%%%%%%%%%%%%%%%%%
%%%%%%%%%%%%%%%%%%%%%%%%%%%%%%%%%%%%%%%%%%%%%%%%%%%%%%%%%%%%%%%%%%%%%%%
\section{Proof of Lemma \ref{lem:AF}}\label{app1}
\begin{proof}
The proof is given in the same way as the proof of Lemma 2 in \cite{zanette2018robust}.
 Let $h_t (y^\ast)$ be a probability density function of  $y^\ast$ at $t$th trial. 
Then, ${\rm E}_{y^\ast} [|  H_{t+1} ({\bm{s}}^\ast,y^\ast) \cup L_{t+1} ({\bm{s}}^\ast,y^\ast) |
] $ can be expressed as follows:
\begin{align*}
&{\rm E}_{y^\ast} [|  H_{t+1} ({\bm{s}}^\ast,y^\ast) \cup L_{t+1} ({\bm{s}}^\ast,y^\ast) | ] \\
&={\rm E}_{y^\ast}   \left [ \sum_{ {\bm{a}} \in \Omega} \1 \{ \mu _t ( {\bm{a}} |({\bm{s}}^\ast,y^\ast) ) - \beta^{1/2} \sigma_t ({\bm{a}}|{\bm{s}}^\ast) 
> h - \epsilon \}  
 \right .\\
& \left . \quad +
 \sum_{ {\bm{a}} \in \Omega} \1 \{ \mu _t ( {\bm{a}} |({\bm{s}}^\ast,y^\ast) ) + \beta^{1/2} \sigma_t ({\bm{a}}|{\bm{s}}^\ast) 
< h + \epsilon \}  \right ] \\
&=    \sum_{ {\bm{a}} \in \Omega} {\rm E}_{y^\ast}[ \1 \{ \mu _t ( {\bm{a}} |({\bm{s}}^\ast,y^\ast) ) - \beta^{1/2} \sigma_t ({\bm{a}}|{\bm{s}}^\ast) 
> h - \epsilon \} ]   
  \\
& \quad +
 \sum_{ {\bm{a}} \in \Omega} {\rm E}_{y^\ast}[ \1 \{ \mu _t ( {\bm{a}} |({\bm{s}}^\ast,y^\ast) ) + \beta^{1/2} \sigma_t ({\bm{a}}|{\bm{s}}^\ast) 
< h + \epsilon \}  ] \\
&=    \sum_{ {\bm{a}} \in \Omega} {\rm P} ( \{ \mu _t ( {\bm{a}} |({\bm{s}}^\ast,y^\ast) ) - \beta^{1/2} \sigma_t ({\bm{a}}|{\bm{s}}^\ast) 
> h - \epsilon \} )  
   \\
&\quad +
 \sum_{ {\bm{a}} \in \Omega} {\rm P} (  \{ \mu _t ( {\bm{a}} |({\bm{s}}^\ast,y^\ast) ) + \beta^{1/2} \sigma_t ({\bm{a}}|{\bm{s}}^\ast) 
< h + \epsilon \}  ) \\
&=    \sum_{ {\bm{a}} \in \Omega} \int _{-\infty}^\infty [ \1 \{ \{ \mu _t ( {\bm{a}} |({\bm{s}}^\ast,y^\ast) ) - \beta^{1/2} \sigma_t ({\bm{a}}|{\bm{s}}^\ast) 
> h - \epsilon \} \}   \\
 & \quad +
  \1 \{   \{ \mu _t ( {\bm{a}} |({\bm{s}}^\ast,y^\ast) ) + \beta^{1/2} \sigma_t ({\bm{a}}|{\bm{s}}^\ast) 
< h + \epsilon \}  \} ] h_t (y^\ast) d y^\ast, 
\end{align*}
where  $\1 \{ \cdot \}$ is  an indicator function, and $\mu _t ( {\bm{a}} |({\bm{s}}^\ast,y^\ast) ) $ is 
the posterior mean of $f({\bm{a}})$ after adding  $({\bm{s}}^\ast,y^\ast)$ to $\{ (  {\bm{s}}_j ( {\bm{x}}_j,c_{i_j}  ) , y_j \}^t_{j=1} $. 
Moreover, from basic properties of GP  (see, e.g., \cite{Rasmussen:2005:GPM:1162254}), 
$\mu _t ( {\bm{a}} |({\bm{s}}^\ast,y^\ast) ) $ and $\sigma^2_t ({\bm{a}} |{\bm{s}}^\ast )$ are given by 
\begin{align*}
\sigma^2_t ({\bm{a}} |{\bm{s}}^\ast )
&= \sigma^2_t ( {\bm{a}}) - \frac{ k^2_t ({\bm{a}},{\bm{s}}^\ast)  }{\sigma^2_t ({\bm{s}}^\ast )+\sigma^2} , \\
\mu _t ( {\bm{a}} |({\bm{s}}^\ast,y^\ast) ) &= 
\mu_t ({\bm{a}}) - \frac{ k_t ({\bm{a}},{\bm{s}}^\ast)  }{\sigma^2_t ({\bm{s}}^\ast )+\sigma^2} ( y^\ast - \mu_t ({\bm{s}}^\ast) ).
\end{align*}
Hence, by using these we have 
\begin{align*}
& \ \mu _t ( {\bm{a}} |({\bm{s}}^\ast,y^\ast) ) - \beta^{1/2} \sigma_t ({\bm{a}}|{\bm{s}}^\ast) 
> h - \epsilon \\
\Leftrightarrow & \ 
\mu_t ({\bm{a}}) - \frac{ k_t ({\bm{a}},{\bm{s}}^\ast)  }{\sigma^2_t ({\bm{s}}^\ast )+\sigma^2} ( y^\ast - \mu_t ({\bm{s}}^\ast) ) -
 \beta^{1/2} \sigma_t ({\bm{a}}|{\bm{s}}^\ast) 
> h - \epsilon  \\
\Leftrightarrow & \ 
 \frac{ k_t ({\bm{a}},{\bm{s}}^\ast)  }{\sigma^2_t ({\bm{s}}^\ast )+\sigma^2} ( y^\ast - \mu_t ({\bm{s}}^\ast) ) <c^+_t ({\bm{a}}|{\bm{s}}^\ast).
\end{align*}
Therefore, noting that $ ( y^\ast - \mu_t ({\bm{s}}^\ast) ) / (\sigma^2_t ({\bm{s}}^\ast )+\sigma^2)^{1/2}      \sim \mathcal{N} (0,1)$, if 
$k_t ({\bm{a}},{\bm{s}}^\ast) >0$,  the following holds:
\begin{align*}
&\int _{-\infty} ^\infty \1 \{ \{ \mu _t ( {\bm{a}} |({\bm{s}}^\ast,y^\ast) ) - \beta^{1/2} \sigma_t ({\bm{a}}|{\bm{s}}^\ast) 
> h - \epsilon \} \}  h_t (y^\ast) d y^\ast \\
&=
\int _{-\infty} ^\infty \1 \left \{   \frac{ y^\ast - \mu_t ({\bm{s}}^\ast) } { \sqrt{\sigma^2_t ({\bm{s}}^\ast )+\sigma^2}    } 
 <
\frac{ \sqrt{\sigma^2_t ({\bm{s}}^\ast  )    +\sigma^2} }{|  k_t ({\bm{a}},{\bm{s}}^\ast)  |  }   c^+_t ({\bm{a}}|{\bm{s}}^\ast) \right    \} h_t (y^\ast) d y^\ast \\
&=
\int _{-\infty} ^\infty \1 \left \{  z <
\frac{ \sqrt{\sigma^2_t ({\bm{s}}^\ast  )    +\sigma^2} }{|  k_t ({\bm{a}},{\bm{s}}^\ast)  |  }    c^+_t ({\bm{a}}|{\bm{s}}^\ast) \right    \} \phi(z) dz \\
&= \Phi  \left ( 
\frac{\sqrt{\sigma^2_{t} ({\bm{s}}^\ast)+\sigma^2}}{|k_t ({\bm{a}},{\bm{s}}^\ast)|} \times ( \mu_{t} ({\bm{a}}) - \beta^{1/2} \sigma_{t} ({\bm{a}}|{\bm{s}}^\ast ) -h+\epsilon )
\right ) .
\end{align*}
Similarly, if $k_t ({\bm{a}},{\bm{s}}^\ast) <0$, it holds that 
\begin{align*}
&\int _{-\infty} ^\infty \1 \{ \{ \mu _t ( {\bm{a}} |({\bm{s}}^\ast,y^\ast) ) - \beta^{1/2} \sigma_t ({\bm{a}}|{\bm{s}}^\ast) 
> h - \epsilon \} \}  h_t (y^\ast) d y^\ast \\
&=
\int _{-\infty} ^\infty \1 \left \{ -  \frac{ y^\ast - \mu_t ({\bm{s}}^\ast) } { \sqrt{\sigma^2_t ({\bm{s}}^\ast )+\sigma^2}    } 
 <
\frac{ \sqrt{\sigma^2_t ({\bm{s}}^\ast  )    +\sigma^2} }{|  k_t ({\bm{a}},{\bm{s}}^\ast)  |  }   c^+_t ({\bm{a}}|{\bm{s}}^\ast) \right    \} h_t (y^\ast) d y^\ast \\
&=
\int _{-\infty} ^\infty \1 \left \{  -z <
\frac{ \sqrt{\sigma^2_t ({\bm{s}}^\ast  )    +\sigma^2} }{|  k_t ({\bm{a}},{\bm{s}}^\ast)  |  }   c^+_t ({\bm{a}}|{\bm{s}}^\ast) \right    \} \phi(z) dz \\
&=
\int _{\infty} ^{-\infty} \1 \left \{  z' <
\frac{ \sqrt{\sigma^2_t ({\bm{s}}^\ast  )    +\sigma^2} }{|  k_t ({\bm{a}},{\bm{s}}^\ast)  |  }   c^+_t ({\bm{a}}|{\bm{s}}^\ast) \right    \} \phi(-z') (-dz') \\
&=
\int _{-\infty} ^{\infty} \1 \left \{  z' <
\frac{ \sqrt{\sigma^2_t ({\bm{s}}^\ast  )    +\sigma^2} }{|  k_t ({\bm{a}},{\bm{s}}^\ast)  |  }   c^+_t ({\bm{a}}|{\bm{s}}^\ast) \right    \} \phi(z') dz' \\
&= \Phi  \left ( 
\frac{\sqrt{\sigma^2_{t} ({\bm{s}}^\ast)+\sigma^2}}{|k_t ({\bm{a}},{\bm{s}}^\ast)|} c^+_t ({\bm{a}}|{\bm{s}}^\ast)
\right ) . 
\end{align*}
Finally, if  $k_t ({\bm{a}},{\bm{s}}^\ast) =0$, we obtain 
\begin{align*}
&\int _{-\infty} ^\infty \1 \{ \{ \mu _t ( {\bm{a}} |({\bm{s}}^\ast,y^\ast) ) - \beta^{1/2} \sigma_t ({\bm{a}}|{\bm{s}}^\ast) 
> h - \epsilon \} \}  h_t (y^\ast) d y^\ast \\
&=\int _{-\infty} ^\infty \1 \{ \mu_t ({\bm{a}})  - \beta^{1/2} \sigma_t ({\bm{a}}|{\bm{s}}^\ast) -  h + \epsilon >0 \} h_t (y^\ast) d y^\ast \\
&= \left \{
\begin{array}{ll}
1 & \text{if} \ \mu_{t} ({\bm{a}}) - \beta^{1/2} \sigma_{t} ({\bm{a}}|{\bm{s}}^\ast ) -h+\epsilon >0 \\
0 & \text{if} \ \mu_{t} ({\bm{a}}) - \beta^{1/2} \sigma_{t} ({\bm{a}}|{\bm{s}}^\ast ) -h+\epsilon \leq 0 \\
\end{array}
\right . .
\end{align*}
By using the same argument, the following integral 
$$
\int _{-\infty}^\infty  \1 \{   \{ \mu _t ( {\bm{a}} |({\bm{s}}^\ast,y^\ast) ) + \beta^{1/2} \sigma_t ({\bm{a}}|{\bm{s}}^\ast) 
< h + \epsilon \}  \} h_t (y^\ast) d y^\ast
$$
can be also calculated. 
\end{proof}

%%%%%%%%%%%%%%%%%%%%%%%%%%%%%%%%%%%%%%%%%%%%%%%%%%%%%%%%%%%%%%%%%%%%%%%
%%%%%%%%%%%%%%%%%%%%%%%%%%%%%%%%%%%%%%%%%%%%%%%%%%%%%%%%%%%%%%%%%%%%%%%
%%%%%%%%%%%%%%%%%%%%%%%%%%%%%%%%%%%%%%%%%%%%%%%%%%%%%%%%%%%%%%%%%%%%%%%
%%%%%%%%%%%%%%%%%%%%%     			Appendix        B                %%%%%%%%%%%%%%%%%
%%%%%%%%%%%%%%%%%%%%%%%%%%%%%%%%%%%%%%%%%%%%%%%%%%%%%%%%%%%%%%%%%%%%%%%
%%%%%%%%%%%%%%%%%%%%%%%%%%%%%%%%%%%%%%%%%%%%%%%%%%%%%%%%%%%%%%%%%%%%%%%
%%%%%%%%%%%%%%%%%%%%%%%%%%%%%%%%%%%%%%%%%%%%%%%%%%%%%%%%%%%%%%%%%%%%%%%
\section{Proof of Theorem \ref{thm:seido}}\label{app2}
\begin{proof}
For any $t \geq 1$, 
 the following inequality holds with probability at least 
$1- |\Omega| e^{-\beta/2} $ (see, e.g., Lemma 5.1 in \cite{Srinivas:2010:GPO:3104322.3104451}): 
\begin{equation}
| f({\bm x}) -\mu_t ({\bm x}) | \leq \beta^{1/2} \sigma_t ({\bm x})  , \quad ^\forall {\bm x} \in \Omega. \label{eq:wpalAAA}
\end{equation}
Thus, by letting $\beta= 2 \log ( |\Omega| \delta ^{-1} )$, \eqref{eq:wpalAAA} holds with probability at least $1-\delta$. 
In addition, let $T$ be  $t$ at the end of the algorithm. 
Then, with probability at least $1-\delta$, it holds that 
\begin{align}
f ({\bm x}) \in Q_T ({\bm x}) , \quad ^\forall {\bm x} \in \Omega. \label{eq:fqt}
\end{align}
Hence, from   \eqref{eq:fqt}  and \eqref{eq:HtLt} we get Theorem \ref{thm:seido}
\end{proof}

%%%%%%%%%%%%%%%%%%%%%%%%%%%%%%%%%%%%%%%%%%%%%%%%%%%%%%%%%%%%%%%%%%%%%%%
%%%%%%%%%%%%%%%%%%%%%%%%%%%%%%%%%%%%%%%%%%%%%%%%%%%%%%%%%%%%%%%%%%%%%%%
%%%%%%%%%%%%%%%%%%%%%%%%%%%%%%%%%%%%%%%%%%%%%%%%%%%%%%%%%%%%%%%%%%%%%%%
%%%%%%%%%%%%%%%%%%%%%     			Appendix        C                %%%%%%%%%%%%%%%%%
%%%%%%%%%%%%%%%%%%%%%%%%%%%%%%%%%%%%%%%%%%%%%%%%%%%%%%%%%%%%%%%%%%%%%%%
%%%%%%%%%%%%%%%%%%%%%%%%%%%%%%%%%%%%%%%%%%%%%%%%%%%%%%%%%%%%%%%%%%%%%%%
%%%%%%%%%%%%%%%%%%%%%%%%%%%%%%%%%%%%%%%%%%%%%%%%%%%%%%%%%%%%%%%%%%%%%%%
\section{Proof of Theorem \ref{thm:owaru}}\label{app3}
First,  let $E$ be an event, and let  $\1 _E$ be an indicator function which takes one if  $E$ holds and zero otherwise. 
Furthermore, for each $t \geq 1$, ${\bm x} \in \Omega$, and cost $c_i $, define  
\begin{align}
E_t ( {\bm x}, c_i ) = \{    ( {\bm{x}} _t , c_{i_t} )  =    ( {\bm x}, c_i ) \} . \label{eq:jishou}
\end{align}
Note that  $E_t ( {\bm x}, c_i )$ is an event where ${\bm x}$ is chosen using the cost $c_i$ at $t$th trial. 
Next, for each $ ({\bm x} , c_i)$, suppose that  
${\bm{W}} _1  ({\bm x} , c_i) , {\bm{W}} _2  ({\bm x} , c_i) , \ldots  $ are random variables where 
${\bm{W}} _1  ({\bm x} , c_i) , {\bm{W}} _2  ({\bm x} , c_i) , \ldots    \sim \text{i.i.d.} \ {\bm{S}} ({\bm x}, c_i )$. 
Moreover, for any  $t \geq 1$,  let ${\bm{A}}_t$ be an input random variable at $t$th trial.
Thus, ${\bm{A}}_t$ can be expressed as 
\begin{align}
{\bm{A}}_t =     \sum_{i=1}^k   \sum _{ {\bm x} \in \Omega}    \1_{E_t ( {\bm x}, c_i )} {\bm{W}} _t  ({\bm x} , c_i) . \label{eq:hensuA}
\end{align}
Finally, for each $ t \geq 1$ and ${\bm x} \in \Omega$, define 
\begin{align}
\hat{\sigma}^2_ t ({\bm x} ) = k({\bm x}, {\bm x} ) - \hat{{\bm{k}} }_t ({\bm x})    ^\top   ( \hat{ {\bm{K}} }_t + \sigma^2 {\bm{I}}_t )^{-1} \hat{{\bm{k}} }_t ({\bm x}) , \label{eq:sigmahat}
\end{align}
where  $\hat{{\bm{k}} } _t ({\bm x})$ 
is a $t$-dimensional vector whose $j$th element is  $k({\bm{A}}_j,{\bm x} )$, and $ \hat{ {\bm{K}} }_t $ is a
$t \times t $ matrix whose 
$(u,v)$ element is $k( {\bm{A}}_u , {\bm{A}}_v ) $. 
Note that  
$ \sigma^2_t ({\bm x})$ is an observed value of the random variable  
$\hat{\sigma}^2_ t ({\bm x} )$. 
 Hence, in order to prove the first half of 
 Theorem \ref{thm:owaru}, it is necessary to show that 
\begin{align}
\hat{\sigma}^2_ t ({\bm x} ) \xrightarrow{\text{a.s.}}  0 , \label{eq:as0}
\end{align}
 where \eqref{eq:as0} means that $\hat{\sigma}^2_ t ({\bm x} )$ converges to zero almost surely. 
The equation \eqref{eq:as0} can be proven by showing the following three facts:
\begin{description}
\item [(Fact1)] For any ${\bm x} \in \Omega $, it holds that 
\begin{align}
\lim _{t \to \infty }   \varsigma^2_t ({\bm x} ) =0. \label{eq:varsigma0}
\end{align}
\item [(Fact2)] For any ${\bm x} \in \Omega $, $\hat{\sigma}^2_ t ({\bm x} )$ converges in probability to zero  (i.e., $\hat{\sigma}^2_ t ({\bm x} ) \xrightarrow{\text{p}}  0 $).
\item [(Fact3)] For any ${\bm x} \in \Omega $, $\hat{\sigma}^2_ t ({\bm x} )$ converges to zero almost surely (i.e., $\hat{\sigma}^2_ t ({\bm x} ) \xrightarrow{\text{a.s.}}  0 $).
\end{description}
First, we prove {\sf (Fact1)}. 
\begin{proof}
Let ${\bm{H}} $ be a 
$ t \times t $ non-singular matrix. 
Then, for any $t$-dimensional vector ${\bm{a}}$ and  ${\bm{b}}$ where 
   ${\bm{H}} + {\bm{a}} {\bm{b}}^\top $ is a non-singular matrix, 
the following holds (see, e.g., \cite{schott2016matrix}):
\begin{align}
( {\bm{H}} + {\bm{a}} {\bm{b}}^\top )^{-1} = {\bm{H}}^{-1} - \frac{ {\bm{H}}^{-1} {\bm{a}} {\bm{b}}^\top {\bm{H}}^{-1}   }{1+ {\bm{b}}^\top {\bm{H}}^{-1} {\bm{a}}}. \label{eq:gyakuA}
\end{align}
Thus, by letting ${\bm{H}} = \sigma^2 {\bm{I}}_t$ and  ${\bm{a}} = {\bm{b}} = k( {\bm x},{\bm x})^{1/2}   {\bm{1}}_t $, from 
\eqref{eq:gyakuA} we have 
\begin{align*}
(  k ({\bm x},{\bm x}) {\bm{1}}_t {\bm{1}}^\top_t +\sigma^2 {\bm{I}}_t ) ^{-1} =  
\sigma^{-2} {\bm{I}}_t -  \frac{\sigma^{-4}  k({\bm x},{\bm x})  {\bm{1}}_t {\bm{1}}^\top_t}{1+t \sigma^{-2} k({\bm x},{\bm x})}.
\end{align*}
Therefore, we get 
\begin{align}
&(k({\bm x},{\bm x}) {\bm{1}}_t )^\top    (  k ({\bm x},{\bm x}) {\bm{1}}_t {\bm{1}}^\top_t +\sigma^2 {\bm{I}}_t ) ^{-1} (k({\bm x},{\bm x}) {\bm{1}}_t ) \nonumber \\
&=
\sigma^{-2} t k({\bm x},{\bm x})^2 -   \frac{\sigma^{-4}  t^2 k({\bm x},{\bm x})^3 }{1+t \sigma^{-2} k({\bm x},{\bm x})} \nonumber  \\
&=    \frac{\sigma^{-2} t k({\bm x},{\bm x})^2+\sigma^{-4}  t^2 k({\bm x},{\bm x})^3 }{1+t \sigma^{-2} k({\bm x},{\bm x})}    - \frac{\sigma^{-4}  t^2 k({\bm x},{\bm x})^3 }{1+t \sigma^{-2} k({\bm x},{\bm x})} \nonumber \\
&=  \frac{\sigma^{-2} t k({\bm x},{\bm x})^2 }{1+t \sigma^{-2} k({\bm x},{\bm x})} . \label{eq:nikoume}
\end{align}
Hence, by substituting \eqref{eq:nikoume} into \eqref{eq:varsigma}, we obtain 
\begin{align}
\varsigma^2_t ({\bm x}) &= k( {\bm x},{\bm x}) 
 -   (  k({\bm x},{\bm x})    {\bm{1}}_t)^\top   (  k({\bm x},{\bm x})  {\bm{1}}_t {\bm{1}}^\top_t + \sigma^2 {\bm{I}}_t )^{-1} 
  (  k({\bm x},{\bm x})    {\bm{1}}_t)  \nonumber \\
&= k( {\bm x},{\bm x})- \frac{\sigma^{-2} t k({\bm x},{\bm x})^2 }{1+t \sigma^{-2} k({\bm x},{\bm x})}   \nonumber \\
&= \frac{k( {\bm x},{\bm x}) +\sigma^{-2} t k({\bm x},{\bm x})^2 }{1+t \sigma^{-2} k({\bm x},{\bm x})}  -\frac{\sigma^{-2} t k({\bm x},{\bm x})^2 }{1+t \sigma^{-2} k({\bm x},{\bm x})}  \nonumber \\ 
&=  \frac{k( {\bm x},{\bm x})  }{1+t \sigma^{-2} k({\bm x},{\bm x})} \label{eq:varsigmabound}.
\end{align}
Thus, for any ${\bm x} \in \Omega$, it holds that $\lim _{t \to \infty } \varsigma^2_t ({\bm x}) =0$.
\end{proof}

Next, we prove {\sf (Fact2)}. 
\begin{proof}
From the definition of convergence in probability, it is sufficient to show that 
\begin{align}
^\forall {\bm x} \in \Omega,   ^\forall a> 0 , ^\forall \varepsilon \in (0,1) , \ ^\exists N ( {\bm x} ) \in \mathbb{N} \ \text{s.t.} \nonumber \\ \ ^\forall n \geq N({\bm x}) , \ {\rm P} (|\hat{\sigma}^2_n ({\bm x}) | <a ) > 1- \varepsilon. \label{eq:pteigi}
\end{align}
Let ${\bm x} $ be an element of $\Omega$, and let $a$ be a positive number. 
In addition, let $\varepsilon$ be a positive number with $ \varepsilon \in (0,1)$. 
Then, from \eqref{eq:varsigma0}, there exists a natural number $N_0 ({\bm x}) \in \mathbb{N}$ such that the following inequality holds for any $N \geq N _0 ({\bm x}) $: 
\begin{align}
\varsigma^2_{ N   }   ({\bm x})   < a/2 . \label{eq:hanbun}
\end{align}
Next, for a natural number $K$ with $K \geq N_0 ({\bm x})$,  we evaluate $\hat{\sigma}^2_{ K  } ({\bm x})$. 
 For a set of random variables 
$ \mathcal{B} = \{ {\bm{B}}_1, \ldots , {\bm{B}}_l \}$, 
let 
\begin{align}
\hat{\sigma}^2_{\mathcal{B} } ({\bm x}) = k({\bm x},{\bm x}) - \hat{{\bm{k}}} _{\mathcal{B} } ({\bm x}) ^\top ( \hat{ {\bm{K}}}_{\mathcal{B}} +\sigma^2 {\bm{I}} _{|\mathcal{B}|} ) ^{-1} \hat{{\bm{k}}} _{\mathcal{B} } ({\bm x}) . \label{eq:setdef}
\end{align}
Here, the $j$th element of $\hat{{\bm{k}}} _{\mathcal{B} } ({\bm x}) $ is $k( {\bm{B}}_j , {\bm x})$, and 
the $(u,v)$th element is  
$k({\bm{B}}_u , {\bm{B}}_v )$. 
Moreover, let 
$ \mathcal{A} = \{  {\bm{A}}_1 , \ldots , {\bm{A}} _{   K            }   \}$. 
Then, we make a random variable 
$\tilde{\sigma}^2_{ K  } ({\bm x}) $ to bound 
$\hat{\sigma}^2_{ K  } ({\bm x})$ as follows:
If $ | \hat{\sigma}^2 _{  \mathcal{A}'   } ({\bm x}) - \varsigma^2_{N_0 ({\bm x})} ({\bm x}) | \geq a/2 $ 
for any $ \mathcal{A}' \subset \mathcal{A} $ with 
 $ | \mathcal{A}'| = N_0 ({\bm x}) $, then we define 
 $\tilde{\sigma}^2_{ K  } ({\bm x}) = k({\bm x},{\bm x})$. 
On the other hand, if $ | \hat{\sigma}^2 _{  \mathcal{A}'   } ({\bm x}) - \varsigma^2_{N_0 ({\bm x})} ({\bm x}) | < a/2$ 
for some $ \mathcal{A}' \subset \mathcal{A} $ with 
$ | \mathcal{A}'| = N_0 ({\bm x}) $, then we define 
$\tilde{\sigma}^2_{ K  } ({\bm x}) = \hat{\sigma}^2 _{  \mathcal{A}'   } ({\bm x}) $. 
Therefore, 
from the definition of $\tilde{\sigma}^2_{ K  } ({\bm x}) $, noting that 
  the posterior variance in GP is monotonically non-increasing, we have 
\begin{align}
|\hat{\sigma}^2_{ K  } ({\bm x}) | \leq |\tilde{\sigma}^2_{ K } ({\bm x}) | . \label{eq:ineq}
\end{align}

Next, 
 we prove that the following inequality holds for some large $K$:
$$   {\rm P}(| \varsigma^2_{N_0 ({\bm x}) } ({\bm x})-  \tilde{\sigma}^2_{ K  } ({\bm x}) | <a/2 ) > 1- \varepsilon . $$
Let $ {\bm{A}}_{j_1} , \ldots ,{\bm{A}}_{ j_{N_0 ({\bm x})} } $ be a sub-sequence of $\{{\bm{A}}_j\}_{j=1}^\infty$, and 
let $\mathcal{A}' \equiv  \{ {\bm{A}}_{j_1} , \ldots ,{\bm{A}}_{ j_{N_0 ({\bm x})} } \}$. Then, 
from  {\sf (A3)}, there exists a positive number $\eta$ such that  
$|\varsigma^2_{N_0 ({\bm x}) } ({\bm x})- \hat{\sigma}^2 _{  \mathcal{A}'   } ({\bm x}) | <a/2$ 
when ${\bm{A}}'_j (\in \mathcal{A}')$ satisfies $ {\bm{A}}'_j  \in \mathscr{N} ({\bm x};\eta ) $.
In order to construct $\mathcal{A}'$, we consider a probability that at least one ${\bm{A}}_ j $ 
from ${\bm{A}}_1$ to $ {\bm{A}}_{K_1 } $ satisfies ${\bm{A}} _j \in  \mathscr{N} ({\bm x};\eta ) $. 
This probability is given by 
\begin{align}
1-   {\rm P} ({\bm{A}} _1 \notin  \mathscr{N} ({\bm x};\eta ) \land \cdots \land {\bm{A}} _{K_1} \notin  \mathscr{N} ({\bm x};\eta ) ) 
. \label{eq:sukunakutomo}
\end{align}
Furthermore, ${\rm P} ({\bm{A}} _2 \notin  \mathscr{N} ({\bm x};\eta ) \land \cdots \land {\bm{A}} _{K_1} \notin  \mathscr{N} ({\bm x};\eta ) )$
 can be expressed as
\begin{align}
&{\rm P} ({\bm{A}} _2 \notin  \mathscr{N} ({\bm x};\eta ) \land \cdots \land {\bm{A}} _{K_1} \notin  \mathscr{N} ({\bm x};\eta ) ) \nonumber \\
&=  {\rm P} ({\bm{A}} _2 \notin  \mathscr{N} ({\bm x};\eta ) ) \times 
{\rm P} ({\bm{A}} _3 \notin  \mathscr{N} ({\bm x};\eta ) |  {\bm{A}} _2 \notin  \mathscr{N} ({\bm x};\eta )       )  \nonumber \\
&\quad \times \nonumber \\
&\quad  \vdots \nonumber \\
&\quad \times \nonumber \\
&\quad {\rm P} ({\bm{A}} _{K_1} \notin  \mathscr{N} ({\bm x};\eta ) |  {\bm{A}} _l \notin  \mathscr{N} ({\bm x};\eta ), \ l \in \{2,\ldots,K_1-1\}      ) . \label{eq:condpr}
\end{align}
Moreover, ${\rm P} ({\bm{A}} _{j} \notin  \mathscr{N} ({\bm x};\eta ) |  {\bm{A}} _2 \notin  \mathscr{N} ({\bm x};\eta )  \land \cdots \land  {\bm{A}} _{j-1} \notin  \mathscr{N} ({\bm x};\eta )    )$ can be written as follows: 
\begin{align}
&{\rm P} ({\bm{A}} _{j} \notin  \mathscr{N} ({\bm x};\eta ) |  {\bm{A}} _2 \notin  \mathscr{N} ({\bm x};\eta )  \land \cdots \land  {\bm{A}} _{j-1} \notin  \mathscr{N} ({\bm x};\eta )    ) \nonumber \\
&= 1- {\rm P} ({\bm{A}} _{j} \in  \mathscr{N} ({\bm x};\eta ) |  {\bm{A}} _l \notin  \mathscr{N} ({\bm x};\eta ),\ l \in \{2,\ldots, j-1\}    )  . \label{eq:akakikae}
\end{align}
In addition, from  {\sf (A2)}, 
 there exists 
${\bm x}^\ast \in \Omega$ and $c_i$ such that ${\rm P}({\bm{S}} ({\bm x}^\ast ,c_i) \in \mathscr{N} ({\bm x}; \eta ) ) \equiv q >0$. 
Hence, by noting that Line \ref{line11}--\ref{line12} in Algorithm \ref{ALG1}, we have 
\begin{align}
&{\rm P} ({\bm{A}} _{j} \in  \mathscr{N} ({\bm x};\eta ) |  {\bm{A}} _2 \notin  \mathscr{N} ({\bm x};\eta )  \land \cdots \land  {\bm{A}} _{j-1} \notin  \mathscr{N} ({\bm x};\eta )    ) \nonumber \\
& \geq {\rm P} ( r_j =1 \land C_j = ({\bm x}^\ast , c_i) \land {\bm{S}} ({\bm x}^\ast ,c_i) \in \mathscr{N} ({\bm x}; \eta )   |  {\bm{A}} _l \notin  \mathscr{N} ({\bm x};\eta ), l \in \{2,\ldots, j-1\}    ) \nonumber \\
& = {\rm P} ( r_j =1 )  {\rm P}( {\bm{S}} ({\bm x}^\ast ,c_i) \in \mathscr{N} ({\bm x}; \eta ) )    {\rm P}( C_j = ({\bm x}^\ast , c_i) 
  ) \nonumber \\
& \geq p_j q \kappa_{min} ,  \label{eq:sitakara}
\end{align}
where  $\kappa_{min } = \min \{ \kappa_1,\ldots ,\kappa_k \} >0$. 
Therefore, from \eqref{eq:condpr}, \eqref{eq:akakikae} and  \eqref{eq:sitakara}, we get 
\begin{align}
{\rm P} ({\bm{A}} _2 \notin  \mathscr{N} ({\bm x};\eta ) \land \cdots \land {\bm{A}} _{K_1} \notin  \mathscr{N} ({\bm x};\eta ) )  \leq 
\prod _{j=2}^{K_1}     (1- p_j q \kappa_{min }  ). \label{eq:mugenseki}
\end{align}
Moreover, by noting that 
$ e^{x} $ can be expanded as $ e^{x} = 1+ x +x^2 e^{x^\star} /2 $, we obtain  the following inequality:
$$e^ { -p_j q \kappa_{min } } = 1  -p_j q \kappa_{min }  + c_{>0}   \geq 
 1  -p_j q \kappa_{min } ,$$
 where $c_{>0}$ is a positive constant. 
Thus, by substituting this inequality to \eqref{eq:mugenseki}, we have 
\begin{align}
{\rm P} ({\bm{A}} _2 \notin  \mathscr{N} ({\bm x};\eta ) \land \cdots \land {\bm{A}} _{K_1} \notin  \mathscr{N} ({\bm x};\eta ) )  \leq  \prod_{j=2}^{K_1}  e^ { -p_j q \kappa_{min } } = e^{- q \kappa_{min} \sum_{j=2}^{K_1} p_j } . \label{eq:sumde}
\end{align}
Hence, by combining \eqref{eq:sukunakutomo} and \eqref{eq:sumde}, the following  holds:
\begin{align*}
&1-   {\rm P} ({\bm{A}} _1 \notin  \mathscr{N} ({\bm x};\eta ) \land \cdots \land {\bm{A}} _{K_1} \notin  \mathscr{N} ({\bm x};\eta ) )  \\
& \geq 1-   {\rm P} ({\bm{A}} _2 \notin  \mathscr{N} ({\bm x};\eta ) \land \cdots \land {\bm{A}} _{K_1} \notin  \mathscr{N} ({\bm x};\eta ) )  \\
& \geq 1- e^{- q \kappa_{min} \sum_{j=2}^{K_1} p_j }  .
\end{align*}
Thus, from {\sf (A1)}, there exists a natural number $K_1$ such that  $e^{- q \kappa_{min } \sum_{j=2}^{K_1} p_j }   < \varepsilon / {N_0 ({\bm x})} $. 
This implies that 
the probability which at least one ${\bm{A}}_ j $ from ${\bm{A}}_1 $ to $ {\bm{A}}_{K_1 } $ satisfies 
${\bm{A}} _j \in  \mathscr{N} ({\bm x};\eta ) $ is greater than $1- \varepsilon / {N_0 ({\bm x})} $. 
Similarly, 
there exists a natural number $K_2$ such that 
the probability which at least one ${\bm{A}}_ {j'} $ from ${\bm{A}}_{K_1+1} $ to $ {\bm{A}}_{K_2 } $ satisfies 
${\bm{A}} _{j'} \in  \mathscr{N} ({\bm x};\eta ) $ is greater than $1- \varepsilon / {N_0 ({\bm x})} $. 
By repeating the same argument, we have 
$K_1,K_2,K_3,\ldots , K_{N_0 ({\bm x})} $. 
Let $K= K_{N_0 ({\bm x})} $, and let $\mathcal{A}_1 = \{ {\bm{A}}_1, \ldots , {\bm{A}} _{K_1} \} 
 ,\mathcal{A}_2 = \{ {\bm{A}}_{K_1+1}, \ldots , {\bm{A}} _{K_2} \}, \ldots , \mathcal{A} _{N_0 ({\bm x}) } =
\{ {\bm{A}}_{K_{N_0 ({\bm x})-1}+1}, \ldots , {\bm{A}} _{K} \}$. 
Then, it holds that 
\begin{align*}
& \{ ^\exists {\bm{A}}_{j_1} \in \mathcal{A}_1 , {\bm{A}} _{j_1} \in  \mathscr{N} ({\bm x};\eta ) \}  
\cap  
\cdots \cap 
\{ ^\exists {\bm{A}}_{j_{N_0 ({\bm x})}} \in \mathcal{A}_{N_0 ({\bm x})} , {\bm{A}} _{j_{N_0 ({\bm x})}} \in  \mathscr{N} ({\bm x};\eta )  \} \\
& \Rightarrow 
|\varsigma^2_{N_0 ({\bm x}) } ({\bm x})- \hat{\sigma}^2 _{  \{ {\bm{A}}_{j_1},  {\bm{A}}_{j_2} , \ldots , {\bm{A}}_{j_{N_0 ({\bm x})}} \}  } ({\bm x}) | <a/2 \\
& \Rightarrow 
|\varsigma^2_{N_0 ({\bm x}) } ({\bm x})- \tilde{\sigma}^2 _K ({\bm x}) | <a/2 .
\end{align*}
Thus, ${\rm P} ( |\varsigma^2_{N_0 ({\bm x}) } ({\bm x})- \tilde{\sigma}^2 _K ({\bm x}) | <a/2     )$ can be bounded as 
\begin{align}
&{\rm P} ( |\varsigma^2_{N_0 ({\bm x}) } ({\bm x})- \tilde{\sigma}^2 _K ({\bm x}) | <a/2     ) \nonumber \\
&\geq 
{\rm P}( \{ ^\exists {\bm{A}}_{j_1} \in \mathcal{A}_1 , {\bm{A}} _{j_1} \in  \mathscr{N} ({\bm x};\eta ) \} 
 \cap   \{ ^\exists {\bm{A}}_{j_2} \in \mathcal{A}_2 , {\bm{A}} _{j_2} \in  \mathscr{N} ({\bm x};\eta ) \} \cap 
\cdots  \nonumber \\
& \quad \quad  
\cap 
\{ ^\exists {\bm{A}}_{j_{N_0 ({\bm x})}} \in \mathcal{A}_{N_0 ({\bm x})} , {\bm{A}} _{j_{N_0 ({\bm x})}} \in  \mathscr{N} ({\bm x};\eta )  \}) \nonumber \\
& \geq \left \{ \sum _{v=1}^{N_0 ({\bm x}) }   {\rm P} \left (   ^\exists {\bm{A}}_{j_v} \in \mathcal{A}_v , {\bm{A}} _{j_v} \in  \mathscr{N} ({\bm x};\eta ) \right ) \right \} -( N_0 ({\bm x})  -1) \nonumber \\
& >  \left \{ \sum _{v=1}^{N_0 ({\bm x}) }   (1-   \varepsilon /{N_0 ({\bm x}) }   ) \right \}-
( N_0 ({\bm x})  -1)  = 1- \varepsilon. \label{eq:epminus}
\end{align}

Finally, we consider \eqref{eq:ineq}. From the triangle inequality, we have 
\begin{align*}
|\hat{\sigma}^2_{ K  } ({\bm x}) | \leq |\tilde{\sigma}^2_{ K } ({\bm x}) | &=  |\tilde{\sigma}^2_{ K } ({\bm x})  -\varsigma^2_{N_0 ({\bm x}) } ({\bm x}) +\varsigma^2_{N_0 ({\bm x}) } ({\bm x})| \\ 
&\leq 
 |\tilde{\sigma}^2_{ K } ({\bm x})  -\varsigma^2_{N_0 ({\bm x}) } ({\bm x}) | + |\varsigma^2_{N_0 ({\bm x}) } ({\bm x})|.
\end{align*}
This implies that  
\begin{align*}
\{  |\tilde{\sigma}^2_{ K } ({\bm x})  -\varsigma^2_{N_0 ({\bm x}) } ({\bm x}) | < a/2   \} \cap  
 \{ |\varsigma^2_{N_0 ({\bm x}) } ({\bm x})| < a/2  \} 
\Rightarrow 
\{ |\hat{\sigma}^2_{ K  } ({\bm x}) | <a \}.
\end{align*}
Therefore, by using \eqref{eq:hanbun} and \eqref{eq:epminus}, it holds that 
\begin{align*}
{\rm P}(|\hat{\sigma}^2_{ K  } ({\bm x}) | <a ) 
& \geq {\rm P}(  \{  |\tilde{\sigma}^2_{ K } ({\bm x})  -\varsigma^2_{N_0 ({\bm x}) } ({\bm x}) | < a/2   \} \cap  
 \{ |\varsigma^2_{N_0 ({\bm x}) } ({\bm x})| < a/2  \}  ) \\
&\geq {\rm P}( |\tilde{\sigma}^2_{ K } ({\bm x})  -\varsigma^2_{N_0 ({\bm x}) } ({\bm x}) | < a/2)  +
{\rm P}(|\varsigma^2_{N_0 ({\bm x}) } ({\bm x})| < a/2) -1 \\
&> 1- \varepsilon +1 -1 = 1- \varepsilon.
\end{align*}
Furthermore, for any $K'$ with $K' \geq K$, it holds that 
$|\hat{\sigma}^2_{ K  } ({\bm x}) |  \geq |\hat{\sigma}^2_{ K'  } ({\bm x}) |$ because posterior variances of GP are non-increasing. 
Hence, noting that 
$$
|\hat{\sigma}^2_{ K  } ({\bm x}) |  < a \Rightarrow |\hat{\sigma}^2_{ K'  } ({\bm x}) |  < a,
$$
we have 
$$
{\rm P} ( |\hat{\sigma}^2_{ K'  } ({\bm x}) |  < a)   \geq {\rm P}(|\hat{\sigma}^2_{ K  } ({\bm x}) | <a )  > 1-\varepsilon.
$$
Consequently,  
$\hat{\sigma}^2_t ({\bm x}) $ converges in probability to zero.
\end{proof}

Next, we prove {\sf (Fact3)}. 
\begin{proof}
From  {\sf (Fact2)},  $\hat{\sigma}^2_t ({\bm x}) $ converges in probability to zero. 
Furthermore, it is known that if a random variable sequence 
 $F_1, F_2 , \ldots $ converges in probability to 
$\alpha$, then there exists a sub-sequence $F_{n_1} , F_{n_2} , \ldots $ such that 
$F_{n_1} , F_{n_2} , \ldots $ converges to $\alpha$ almost surely  (see, e.g., \cite{Shao_2003_book}). 
Hence, there exists a sub-sequence $\hat{\sigma}^2_{n_1} ({\bm x}), \hat{\sigma}^2_{n_2} ({\bm x})  , \ldots $
 such that 
\begin{align}
\hat{\sigma}^2_{n_t} ({\bm x}) \xrightarrow{ \text{a.s.} } 0 , \  (\text{as} \ t \to \infty). \label{eq:bubunas}
\end{align}
In addition, noting that  posterior variances of GP are non-increasing, $\hat{\sigma}^2_t ({\bm x})$ satisfies that 
$\hat{\sigma}^2_1 ({\bm x}) \geq \hat{\sigma}^2_2 ({\bm x}) \geq \cdots \geq 0$. 
Thus, by using this inequality and \eqref{eq:bubunas}, we have 
\begin{align}
\hat{\sigma}^2_{t} ({\bm x}) \xrightarrow{ \text{a.s.} } 0 , \  (\text{as} \ t \to \infty). \label{eq:zenbuas}
\end{align}
\end{proof}

Finally, we prove the second half of Theorem \ref{thm:owaru}.
\begin{proof}
For any ${\bm x} \in \Omega$, 
\eqref{eq:zenbuas} can be expressed as follows: 
\begin{align*}
&\hat{\sigma}^2_{t} ({\bm x}) \xrightarrow{ \text{a.s.} } 0 , \  (\text{as} \ t \to \infty) \\
& \Leftrightarrow 
^\exists \text{event} \ E_{\bm x} \in \mathscr{B} \ \text{s.t.} \ {\rm P}(E_{\bm x}) =1, \ ^\forall \omega \in E_{\bm x} , \ \lim_{t \to \infty} \left ( \hat{\sigma}^2_{t} ({\bm x}) \right ) (\omega)   =0,
\end{align*}
where $\mathscr{B}$ is a $\sigma$-field of a probability space $ (\mathcal{S}, \mathscr{B} , {\rm P} )$, and 
$\left ( \hat{\sigma}^2_{t} ({\bm x}) \right ) (\omega) $
 is the observed value of the random variable $\hat{\sigma}^2_{t} ({\bm x}) $ at the point $\omega \in \mathcal{S}$. 
By using  $E_ {\bm x}$, we define an event  $E$ as 
$$
E \equiv  \bigcap _{ {\bm x} \in \Omega }  E_ {\bm x} . 
$$
From the definition of $E$, the following holds:
\begin{align}
 E \in  \mathscr{B} , \   {\rm P} (E) =1 , \ ^\forall \omega \in E, \ ^\forall {\bm x} \in \Omega , \ \lim_{t \to \infty} \left ( \hat{\sigma}^2_{t} ({\bm x}) \right ) (\omega)   =0. \label{eq:zenbujisyoas}
\end{align}
Hence, from the classification rule \eqref{eq:HtLt}, 
if $ \beta^{1/2} \sigma _t ({\bm x} ) < \epsilon $ for any 
 ${\bm x} \in \Omega$, then all the points are classified. 
Thus, noting that $\beta$ is positive, it is sufficient to show that 
$ \sigma^2_t ({\bm x}) < \epsilon ^2 \beta^{-1}$. 
Since $\sigma^2 _t ({\bm x}) $ is the observed value of $\hat{\sigma}^2_{t} ({\bm x})$, 
 from \eqref{eq:zenbujisyoas}, 
there exists a natural number $N_{\omega,{\bm x}} \in \mathbb{N}$ such that 
$\left ( \hat{\sigma}^2_{N_{\omega,{\bm x}}} ({\bm x}) \right ) (\omega) < \epsilon ^2 \beta^{-1}$ 
for any $\omega \in E$ and  ${\bm x} \in \Omega$. 
Therefore, by letting $ N _\omega = \max _{{\bm x} \in \Omega}  N_{\omega,{\bm x}}$, it holds that 
$\left ( \hat{\sigma}^2_{N_\omega} ({\bm x}) \right ) (\omega) < \epsilon ^2 \beta^{-1}$
 for any $ {\bm x} \in \Omega$.  
This implies that 
$$
 ^\exists \text{event} \ E \in \mathscr{B} , \ {\rm P} (E) =1 
$$
and 
\begin{align*}
^\forall \omega \in E , \ ^\exists N_\omega \in \mathbb{N} \ \text{s.t.} \ 
^\forall {\bm x} \in \Omega , \ \left ( \hat{\sigma}^2_{N_\omega} ({\bm x}) \right ) (\omega) \equiv \sigma^2_{N_\omega} ({\bm x}) < \epsilon ^2 \beta^{-1}.
\end{align*}
Consequently, we have the second half of Theorem \ref{thm:owaru}.
\end{proof}

%%%%%%%%%%%%%%%%%%%%%%%%%%%%%%%%%%%%%%%%%%%%%%%%%%%%%%%%%%%%%%%%%%%%%%%
%%%%%%%%%%%%%%%%%%%%%%%%%%%%%%%%%%%%%%%%%%%%%%%%%%%%%%%%%%%%%%%%%%%%%%%
%%%%%%%%%%%%%%%%%%%%%%%%%%%%%%%%%%%%%%%%%%%%%%%%%%%%%%%%%%%%%%%%%%%%%%%
%%%%%%%%%%%%%%%%%%%%%     			Appendix        D                %%%%%%%%%%%%%%%%%
%%%%%%%%%%%%%%%%%%%%%%%%%%%%%%%%%%%%%%%%%%%%%%%%%%%%%%%%%%%%%%%%%%%%%%%
%%%%%%%%%%%%%%%%%%%%%%%%%%%%%%%%%%%%%%%%%%%%%%%%%%%%%%%%%%%%%%%%%%%%%%%
%%%%%%%%%%%%%%%%%%%%%%%%%%%%%%%%%%%%%%%%%%%%%%%%%%%%%%%%%%%%%%%%%%%%%%%

\section{Proofs of Theorem \ref{thm:pro_owaru}--\ref{thm:owaru5}}\label{app:lemmaproof}
\begin{proof}
First, we prove Theorem \ref{thm:pro_owaru}. 
From \eqref{eq:varsigmabound}, for any ${\bm x} \in \Omega$, the following inequality holds:
$$
\varsigma^2 _t ({\bm x}) =\frac{ k({\bm x},{\bm x})}{1+ t \sigma^{-2} k({\bm x},{\bm x}) }  \leq \frac{ k({\bm x},{\bm x})}{ t \sigma^{-2} k({\bm x},{\bm x}) }=t^{-1}  \sigma^2.
$$
Thus, for  the positive integer $T^\ast$ satisfying the inequality $T^\ast > 2 \beta \sigma^2 \epsilon ^{-2}$, it holds that  $\varsigma^2_{T^\ast} ({\bm x}) <2^{-1} \beta^{-1} \epsilon ^2$. 
In addition, noting that the kernel function $k({\bm x},{\bm x}')$ is continuous at $({\bm x},{\bm x})$, we get 
$$\varsigma^2_{   {\bm{a}} ^{(T^\ast)}  } ({\bm x})  \to \varsigma^2_{T^\ast} ({\bm x}) , \quad ( {\rm as} \  {\bm{a}} ^{(T^\ast)} \to {\bm x} {\bm{1}}_{T^\ast} ).
$$
This implies that there exists a positive number  $\nu $ such that 
$$
^\forall {\bm{a}}^{(T^\ast)} =({\bm{a}}_1,\ldots, {\bm{a}}_{T^\ast} ) \in \bigotimes_{i=1}^{T^\ast}  \mathscr{N} ({\bm x} ;\nu) ,\quad 
| \varsigma^2_{T^\ast} ({\bm x}) - \varsigma^2_{{\bm{a}}^{(T^\ast)} } ({\bm x}) | < \frac{\epsilon^2}{2 \beta}.
$$
On the other hand,  when $r_t$ in 8th line of Algorithm \ref{ALG1} is one, 
the probability where ${\bm{s}}_t \in \mathscr{N} ({\bm x}, \nu)$ is 
$\tilde{p} _{\bm x} $. 
Therefore, noting that $\tilde{p} _{\bm x} \geq \tilde{p}^\ast$, for any 
$ t \geq 1$ and ${\bm x} \in \Omega$, the probability where   ${\bm{s}}_t \in \mathscr{N} ({\bm x}, \nu)$  is larger than 
$p_t \tilde{p}^\ast$. Here,      let   $N_{j-1} $ and $N_j $  be non-negative integers satisfying  $N_{j-1} <N_j $  and 
\begin{align}
\prod_{t=N_{j-1}+1}^{N_j} (1-p_t \tilde{p}^\ast) < \frac{\delta}{|\Omega| T^\ast }. \label{eq:njpro}
\end{align}
Then, with probability at least $1- \delta (|\Omega| T^\ast ) ^{-1}$, 
there exists at least one ${\bm{s}}_ t$ satisfying  ${\bm{s}}_t \in \mathscr{N} ({\bm x}; \nu )$ during the trials from $N_{j-1}+1$ to $N_j$.
Recall that since $1- x \leq  e^{-x} $, the following inequality holds:
$$
\prod_{t=N_{j-1}+1}^{N_j} (1-p_t \tilde{p}^\ast) \leq e^{-\tilde{p}^\ast   \sum_{t=N_{j-1}+1}^{N_j} p_t}.
$$
Moreover, from the assumption $\sum_{t=1}^\infty p_t $ tends to infinity. 
Hence, there exists   $N_{j-1} $ and $N_j$ satisfying \eqref{eq:njpro} and $N_{j-1} <N_j $. 
Thus, after at most $N_{|\Omega| T^\ast }$ trials, with probability at least $1-\delta$, 
for any ${\bm x} \in \Omega$ there exists a subsequence ${\bm{s}}_{{\bm x},1},\ldots , {\bm{s}}_{{\bm x},T^\ast }$ of ${\bm{s}}_1,\ldots, {\bm{s}}_{ N_{|\Omega| T^\ast }} $ such that 
$${\bm{s}}^{(T^\ast)}_{\bm x} = ( {\bm{s}}_{{\bm x},1} , \ldots , {\bm{s}}_{{\bm x}, T^\ast } )  \in \bigotimes _{i=1} ^{T^\ast} \mathscr{N} ({\bm x}; \nu).$$
Therefore, we get 
\begin{align*}
\sigma^2_{ N_{|\Omega| T^\ast }} ({\bm x})  &\leq \varsigma^2_{ {\bm{s}}^{(T^\ast)}_{\bm x} } ({\bm x})   \\
&\leq |\varsigma^2_{ {\bm{s}}^{(T^\ast)}_{\bm x} } ({\bm x})  - \varsigma ^2_{T^\ast } ({\bm x} ) | + \varsigma ^2_{T^\ast } ({\bm x} ) \\
& <  \frac{\epsilon^2}{2 \beta } +  \frac{\epsilon^2}{2 \beta } = \frac{\epsilon^2}{\beta}.
\end{align*}
This implies that $\beta^{1/2} \sigma _{ N_{|\Omega| T^\ast }} ({\bm x}) < \epsilon$. 
Hence, from the classification rule, all candidate points are classified.

Second, we prove Theorem \ref{thm:pro_owaru2}. 
From {\sf (A1')}, it holds that $p \equiv p^\ast \tilde{p} ^\ast  \leq p_t \tilde{p}^\ast$. 
Thus, we obtain 
$$
\prod_{t=N_{j-1}+1}^{N_j} (1-p_t \tilde{p}^\ast) \leq  (1- p )^{N_j -N_{j-1} }.
$$
Here, let $r$ be the smallest positive integer satisfying 
$$
r > \frac{\log ( |\Omega| T^\ast \delta ^{-1} )}{ -\log (1-p)}.
$$
Then, it holds that 
$$
 (1-p)^r < \frac{\delta}{| \Omega| T^\ast }.
$$
Hence, for the non-negative integers $0=N_0,N_1, \ldots , N_{  |\Omega| T^\ast } $ with 
$r= N_j -N_{j-1}$, the following inequality holds:
$$
\prod_{t=N_{j-1}+1}^{N_j} (1-p_t \tilde{p}^\ast) \leq  (1- p )^{N_j -N_{j-1} } = (1-p)^r < \frac{\delta}{| \Omega| T^\ast }.
$$
Then, $N_ { |\Omega| T^\ast }$ can be expressed as
$$
N_ { |\Omega| T^\ast } = \sum _{j=1} ^{|\Omega| T^\ast}  ( N_j - N_{j-1} ) + N_0 = r |\Omega| T^\ast +0=r |\Omega| T^\ast.
$$

Third, we prove Theorem \ref{thm:pro_owaru3} and \ref{thm:pro_owaru4}. 
For any 
 ${\bm x} ^{(t)} = ({\bm x}_1,\ldots,{\bm x}_t ) $ and ${\bm x} \in \Omega$, it holds that $\sigma^2 _t ({\bm x}) \leq \tilde{\varsigma}^2_ { {\bm x}^{(t)}  } ({\bm x}) $ and 
$$
\tilde{\varsigma}^2_ { {\bm x}^{(t)}  } ({\bm x}) \leq | \tilde{\varsigma}^2_ { {\bm x}^{(t)}  } ({\bm x}) - \tilde{\varsigma}^2_t ({\bm x}) | +  \tilde{\varsigma}^2_t ({\bm x}).
$$
Then, by using the same arguments as the proofs of  Theorem \ref{thm:pro_owaru} and \ref{thm:pro_owaru2}, 
 Theorem \ref{thm:pro_owaru3} and \ref{thm:pro_owaru4} can be proved.

Finally, we prove Theorem \ref{thm:owaru5}. 
Note that the proofs of Theorem \ref{thm:owaru}--\ref{thm:pro_owaru4} %do not depend on whether input distributions are correctly specified, and
 are given by using only the randomized strategy.
Furthermore, %from Definition \ref{def:AF2} and \ref{def:AF22},    
the randomized strategy does not depend on whether input distributions are correctly specified. 
Therefore, Theorem \ref{thm:owaru}--\ref{thm:pro_owaru4} hold even if input distributions are unknown.  
\end{proof}

%%%%%%%%%%%%%%%%%%%%%%%%%%%%%%%%%%%%%%%%%%%%%%%%%%%%%%%%%%%%%%%%%%%%%%%
%%%%%%%%%%%%%%%%%%%%%%%%%%%%%%%%%%%%%%%%%%%%%%%%%%%%%%%%%%%%%%%%%%%%%%%
%%%%%%%%%%%%%%%%%%%%%%%%%%%%%%%%%%%%%%%%%%%%%%%%%%%%%%%%%%%%%%%%%%%%%%%
%%%%%%%%%%%%%%%%%%%%%     			Appendix        E                %%%%%%%%%%%%%%%%%
%%%%%%%%%%%%%%%%%%%%%%%%%%%%%%%%%%%%%%%%%%%%%%%%%%%%%%%%%%%%%%%%%%%%%%%
%%%%%%%%%%%%%%%%%%%%%%%%%%%%%%%%%%%%%%%%%%%%%%%%%%%%%%%%%%%%%%%%%%%%%%%
%%%%%%%%%%%%%%%%%%%%%%%%%%%%%%%%%%%%%%%%%%%%%%%%%%%%%%%%%%%%%%%%%%%%%%%
\section{Additional numerical experiments}\label{app-additional}

\subsection{Effect of probability $ p_t $}
\begin{figure*}[t]
\begin{center}
\scalebox{1}{
 \begin{tabular}{c}
 \includegraphics[width=1\textwidth]{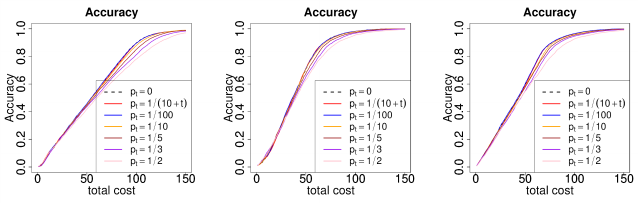} 
 \end{tabular}}
\end{center}
 \caption{Average accuracy based on 20 Monte Carlo simulations in cases 1 to 3.
 The left, center and right figure show the 
Case1, Case2 and Case3, respectively.}
 \label{fig:exp000}
\end{figure*}

In this subsection, we confirm the difference in behavior due to the difference in probability $ p_t $ through a two-dimensional synthetic function. 
We first set $ \Omega $ as a grid point that is obtained by uniformly cutting the region $ [-5,5] \times [-5,5] $ into $ 30 \times 30 $. 
As the kernel function, we used Gaussian kernel $ k ({\bm x}, {\bm x} ') = \sigma ^ 2_f \exp (-\| {\bm x}-{\bm x}' \| ^ 2_2 / L) $.
 Furthermore, we used the error variance $\sigma^2 = 10^{-4} $, 
the accuracy parameter $\epsilon = 10^{-12}$ and  
  $\beta^{1/2} =1.96$. 
In this experiment, we considered two costs  $c_1 =1$ and $c_2 =2$.  
For each $c_i$ and $ {\bm x} =(x_1,x_2 )^\top \in \Omega$, the following was used as ${\bm{S}} ({\bm x} , c_i )$: 
$$
{\bm{S}} ({\bm x} , c_i ) =  (\mathcal{U} _{[L_1 ({\bm x},c_i),U_1 ({\bm x},c_i)]}   , \mathcal{U} _{[L_2 ({\bm x},c_i),U_2 ({\bm x},c_i)]} )          ^\top , 
$$
where $\mathcal{U}_{[a,b]}$ is a uniform distribution on $[a,b]$, and 
\begin{align*}
L_1 ({\bm x},c_i) &=    \max \{       (x_1 - \zeta^{(i)} ), -5 \} , \\
   U_1 ({\bm x},c_i) &=    \min \{       (x_1 + \zeta^{(i)} ), 5 \} ,\\
L_2 ({\bm x},c_i) &=    \max \{       (x_2 - \zeta^{(i)} ), -5 \} , \\
   U_2 ({\bm x},c_i) &=    \min \{       (x_2 + \zeta^{(i)} ), 5 \}  .
\end{align*}
Here, we set $\zeta^{(1)} = 1/2.9$  and  $\zeta^{(2)}  =0.05$. 
 Moreover, we assumed  that $\mathcal{U} _{[L_1 ({\bm x},c_i),U_1 ({\bm x},c_i)]} $ and $ \mathcal{U} _{[L_2 ({\bm x},c_i),U_2 ({\bm x},c_i)]}$
 are mutually independent. 
Then, by letting 
$p_t =0$, $p_t = 1/(10+t)$ and 
  $p_t = 1/a $, $a \in \{2,3,5,10,100 \}$,  we confirmed the behavior when each probability was used.
In addition, we also set  $\kappa_1= (1- |\Omega| 10^{-8})/|\Omega| $ and  $\kappa_2 =10^{-8} $. 
For true functions, kernel parameters and thresholds, we considered the following three cases:
\begin{description}
\item [(Case1)] True function: $f(x_1,x_2) = x^2_1 +x^2_2$, kernel parameters: $\sigma^2_f =225$, $L=2$, threshold: $h=20$. 
\item [(Case2)] True function: $f(x_1,x_2) = -(x^2_1+x_2-11)^2-(x_1+x^2_2-7)^2$, kernel parameters: $\sigma^2_f =3000$, $L=2$, threshold: $h=-50$.
\item [(Case3)] True function: $f(x_1,x_2) = \sum_{j=1}^2 (x^4_j -16x^2_j +5x_j )/2$, kernel parameters: $\sigma^2_f =900$, $L=2$, threshold: $h=-10$.
\end{description}
In order to compute integrals, we used the approximation method based on  \eqref{eq:approx2}.
Note that the discrete distribution $ \tilde {{\bm {S}}} ({\bm x}, c_i) $ can be derived analytically in the settings of this subsection.
Under this setting, one initial point was taken at random, and points were acquired until the total cost reached 150. 
The average obtained by 20 Monte Carlo simulations is given in Figure \ref{fig:exp000}.
From Figure \ref{fig:exp000}, 
we can confirm that $ p_t $ for establishing the theoretical guarantee does not have a dramatic effect on the result if a sufficiently small value is set.
Moreover, we can also confirm that 
the proposed method can achieve high accuracy at low cost.

%%%%%%%%%%%%%%%%%%%%%%%%%%%%%%%%%%%%%%%%%%%%%%%%%%%%%%%%%%%%%%%%%%%%%%%%%%%%%%%%%%%%%%%%
%%%%%%%%%%%%%%%%%%%%%%%%%%%%%%%%%%%%%%%%%%%%%%%%%%%%%%%%%%%%%%%%%%%%%%%%%%%%%%%%%%%%%%%%
%%%%%%%%%%%%%%%%%%%%%%%%%%%%%%%%          Figure 9           %%%%%%%%%%%%%%%%%%%%%%%%%%%%%%%%%%%%
%%%%%%%%%%%%%%%%%%%%%%%%%%%%%%%%%%%%%%%%%%%%%%%%%%%%%%%%%%%%%%%%%%%%%%%%%%%%%%%%%%%%%%%%
%%%%%%%%%%%%%%%%%%%%%%%%%%%%%%%%%%%%%%%%%%%%%%%%%%%%%%%%%%%%%%%%%%%%%%%%%%%%%%%%%%%%%%%%
%%%%%%%%%%%%%%%%%%%%%%%%%%%%%%%%%%%%%%%%%%%%%%%%%%%%%%%%%%%%%%%%%%%%%%%%%%%%%%%%%%%%%%%%

\subsection{Synthetic experiments}
In this subsection, we compare the proposed method with some existing methods using synthetic functions.
Hereafter, for simplicity, we used 
 $p_t =0 $.

\begin{figure*}[t]
\begin{center}
\scalebox{1}{
 \begin{tabular}{c}
\includegraphics[width=1\textwidth]{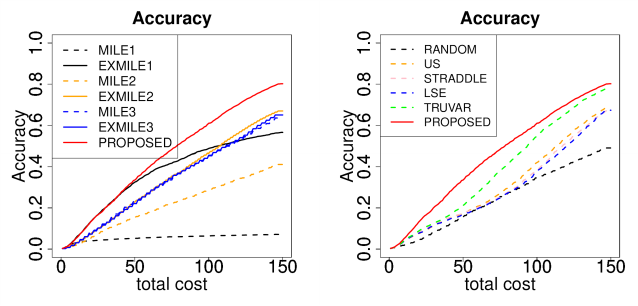} 
 \end{tabular}}
\end{center}
 \caption{
Average accuracy based on 20 Monte Carlo simulations in the 
Rosenbrock function. 
The left figure  shows the influence of integration against  the input distribution and that of cost in evaluating the input point. 
The right figure shows the result of comparison with existing methods.}
 \label{fig:exp2}
\end{figure*}

\paragraph{Two-dimensional Rosenbrock function} \label{experiment2}
We also considered the  2-dimensional Rosenbrock function (reduced to 1/100 and moved)  
$$
f(x_1,x_2)=(x_2-x^2_1)^2+(1-x_1)^2/100 -5
$$
as the true function, and defined the grid point obtained by uniformly cutting the region  $[-2,2] \times [-1,3]$ into $40 \times 40$ as $\Omega$. 
Furthermore, we used the Gaussian kernel 
with  $\sigma^2_f = 64 $ and  $L = 0.5$. In addition, we set $\sigma^2 = 0.25$, 
$h=0$, $\epsilon = 10^{-12}$ and 
  $\beta^{1/2} =1.96$.  
Similarly in this experiment, we considered three costs   $c_1 =1$, $c_2 =2$ and $c_3 =3$. 
Moreover, for each $c_i$ and $ {\bm x} =(x_1,x_2 )^\top \in \Omega$,  we assumed that 
$$
{\bm{S}} ({\bm x} , c_i ) =  {\bm x} +   (  G_{[0,2-x_1] } ( \zeta^{(i)},1) ,    G_{[0,3-x_2] } (\zeta^{(i)},1)  ) ^\top , 
$$
where $G_{[0,2-x_1] } ( \zeta^{(i)},1) $ and $   G_{[0,3-x_2] } (\zeta^{(i)},1)$ are independent. 
Furthermore, we used  
$\zeta^{(1)} =4$, $\zeta^{(2)} =1 $, $\zeta^{(3)} =0.01$. 
Under this setting,   we performed   similar experiments  to sinusoidal function in Subsection \ref {experiment1}.
From Figure \ref{fig:exp2}, even in the case of the Rosenbrock function, we can see that the proposed method has higher accuracy than the other methods.

%%%%%%%%%%%%%%%%%%%%%%%%%%%%%%%%%%%%%%%%%%%%%%%%%%%%%%%%%%%%%%%%%%%%%%%%%%%%%%%%%%%%%%%%
%%%%%%%%%%%%%%%%%%%%%%%%%%%%%%%%%%%%%%%%%%%%%%%%%%%%%%%%%%%%%%%%%%%%%%%%%%%%%%%%%%%%%%%%
%%%%%%%%%%%%%%%%%%%%%%%%%%%%%%%%          Figure 10           %%%%%%%%%%%%%%%%%%%%%%%%%%%%%%%%%%%
%%%%%%%%%%%%%%%%%%%%%%%%%%%%%%%%%%%%%%%%%%%%%%%%%%%%%%%%%%%%%%%%%%%%%%%%%%%%%%%%%%%%%%%%
%%%%%%%%%%%%%%%%%%%%%%%%%%%%%%%%%%%%%%%%%%%%%%%%%%%%%%%%%%%%%%%%%%%%%%%%%%%%%%%%%%%%%%%%
%%%%%%%%%%%%%%%%%%%%%%%%%%%%%%%%%%%%%%%%%%%%%%%%%%%%%%%%%%%%%%%%%%%%%%%%%%%%%%%%%%%%%%%%

\begin{figure*}[t]
\begin{center}
\scalebox{1}{
 \begin{tabular}{c}
 \includegraphics[width=1\textwidth]{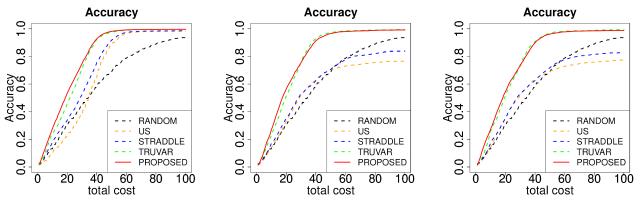} 
 \end{tabular}}
\end{center}
 \caption{Average accuracy based on 50 Monte Carlo simulations in case1--3.
 The left, center and right figure show the 
Case1, Case2 and Case3, respectively.}
 \label{fig:result_unknown1}
\end{figure*}

\begin{figure*}[t]
\begin{center}
\scalebox{1}{
 \begin{tabular}{c}
 \includegraphics[width=1\textwidth]{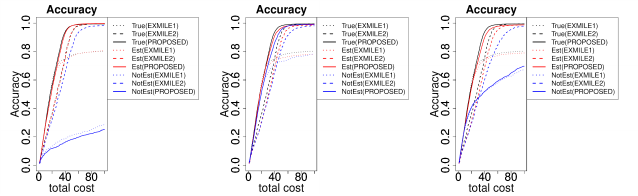} 
 \end{tabular}}
\end{center}
 \caption{
Average accuracy based on 50 Monte Carlo simulations in case1--3.
 The left, center and right figure show the 
Case1, Case2 and Case3, respectively. 
Moreover, True, Est, and NotEst indicate the results when the true density function is known, when parameter estimation is performed, and when parameter estimation is not performed.}
 \label{fig:result_unknown2}
\end{figure*}

\paragraph{One-dimensional function and unknown input distributions} \label{subsub:unknown}
Here, we considered the one dimensional function 
$$
f(x)= \cos (10x) + \sin (12x) +x^2/10
$$
as the true function, and defined the grid point observed by uniformly cutting the interval 
  $[0,5]$ into 100 as $\Omega$. In addition, we used the Gaussian kernel with
$\sigma^2_f = 2 $ and $L = 0.1$. 
Furthermore, we set   $\sigma^2 = 10^{-4}$, 
  $h=0.4$,  $\epsilon = 10^{-12}$ and 
  $\beta^{1/2} =3$. 
In this experiment, we considered two costs 
  $c_1 =1$ and $c_2 =2$. 
Moreover, for each 
 $c_i$ and $ x \in \Omega$, we defined that
$$
{{S}} (x , c_i ) = x + \mathcal{N} ( \mu_{c_i} , \sigma^2_{ c_i} ), 
$$
where $(\mu_{c_1},\sigma^2_{c_1} )^\top =(2,0.16)^\top$ and 
 $(\mu_{c_2},\sigma^2_{c_2} )^\top =(0,10^{-4})^\top$. 
Then, we considered the following three cases:
\begin{description}
\item [Case1] Assume that $\mu_{c_i} $ and  $\sigma^2_{c_i}$ are unknown and known, respectively. 
 Moreover, we used  $\mu_{c_i} \sim \mathcal{N} (\mu_{c_i,0},\sigma^2_{c_i,0} )$ as a prior distribution of 
 $\mu_{c_i}$, where  $(\mu_{c_1,0},\sigma^2_{c_1,0} )^\top =(0.5,0.1)^\top$ and 
$(\mu_{c_2,0},\sigma^2_{c_2,0} )^\top =(0.15,0.03)^\top$.
%%%%%%%%%%%%%%%%%%%%%%%%%%%%%%%%%%%%%%%%%%%%%%%%%%%%
\item [Case2] Assume that $\mu_{c_i} $ and  $\sigma^2_{c_i}$ are known and unknown, respectively. In addition, 
we used $\sigma^{-2} _{c_i} \sim \mathcal{G} (\alpha_{c_i,0},\beta_{c_i,0} )$ as a prior of 
 $\sigma^{-2} _{c_i}$, where   $(\alpha_{c_1,0},\beta_{c_1,0} )^\top =(2,2)^\top$ and 
$(\alpha_{c_2,0},\beta_{c_2,0} )^\top =(5,1)^\top$.
%%%%%%%%%%%%%%%%%%%%%%%%%%%%%%%%%%%%%%%%%%%%%%%%%%%%%%%
\item [Case3] Assume that both $\mu_{c_i} $ and $\sigma^2_{c_i}$ are unknown. 
Moreover,  as priors of $\mu_{c_i } $ and $\sigma^{-2}_{c_i} $, we used  
 $\mu_{c_i } \sim \mathcal{N} (\mu_{c_i,0}, \sigma^2_{c_i}/\kappa_{c_i,0} ) $ and  $\sigma^{-2}_{c_i} \sim  \mathcal{G} (\alpha_{c_i,0},\beta_{c_i,0} )$, where 
  \begin{align*}
( \mu_{c_1,0},\kappa_{c_1,0},\alpha_{c_1,0},\beta_{c_1,0} )^\top &=(  0.5,1,2,2)^\top, \\
( \mu_{c_2,0},\kappa_{c_2,0},\alpha_{c_2,0},\beta_{c_2,0} )^\top &=(  0.15,1,5,1)^\top .
\end{align*}
\end{description}
Note that in Case1, $ g_t (x | {\bm \theta} ^ {(c_i)} _ {\bm x}) $ is a density function with  normal distribution, and also note that in Case2-3,  $g_t (x|{\bm\theta}^{(c_i)}_{\bm x} )$ is a density function with $t$-distribution (see, e.g., \cite{bishop2006pattern}). 
Under this setting,  
 we performed similar experiments until the total cost reached 100, where we used 
 the approximation 
 \eqref{eq:approx1} with $M=1000$. 
The average obtained by 50 Mote Carlo simulations is given in 
Figure \ref{fig:result_unknown1}. 
 From  Figure \ref{fig:result_unknown1}, 
 we can confirm that the proposed method and TRUVAR  have higher accuracy than other existing methods. 
Similarly, Figure  \ref{fig:result_unknown2} shows the accuracy 
when the true density function, estimated function $g_t(x|{\bm\theta}^{(c_i)}_x )$ and 
not-estimated function (i.e.,  $g_0(x|{\bm\theta}^{(c_i)}_x )$) are used as an approximation of $g(x|{\bm\theta}^{(c_i)}_x )$ in 
\eqref{eq:approx1}.
From Figure \ref{fig:result_unknown2}, we can see that when parameter estimation is not performed, efficient classification cannot be performed. 
On the other hand, it can be confirmed that accuracy improvement has been achieved by parameter estimation.
In particular, in the case of this experimental, it can be confirmed that performance equivalent to that obtained when the true distribution was known was achieved by parameter estimation.

%%%%%%%%%%%%%%%%%%%%%%%%%%%%%%%%%%%%%%%%%%%%%%%%%%%%%%%%%%%%%%%%%%%%%%%%%%%%%%%%%%%%%%%%
%%%%%%%%%%%%%%%%%%%%%%%%%%%%%%%%%%%%%%%%%%%%%%%%%%%%%%%%%%%%%%%%%%%%%%%%%%%%%%%%%%%%%%%%
%%%%%%%%%%%%%%%%%%%%%%%%%%%%%%%%          Figure 11           %%%%%%%%%%%%%%%%%%%%%%%%%%%%%%%%%%%
%%%%%%%%%%%%%%%%%%%%%%%%%%%%%%%%%%%%%%%%%%%%%%%%%%%%%%%%%%%%%%%%%%%%%%%%%%%%%%%%%%%%%%%%
%%%%%%%%%%%%%%%%%%%%%%%%%%%%%%%%%%%%%%%%%%%%%%%%%%%%%%%%%%%%%%%%%%%%%%%%%%%%%%%%%%%%%%%%
%%%%%%%%%%%%%%%%%%%%%%%%%%%%%%%%%%%%%%%%%%%%%%%%%%%%%%%%%%%%%%%%%%%%%%%%%%%%%%%%%%%%%%%%

%%%%%%%%%%%%%%%%%%%%%%%%%%%%%%%%%%%%%%%%%%%%%%%%%%%%%%%%%%%%%%%%%%%%%%%%%%%%%%%%%%%%%%%%
%%%%%%%%%%%%%%%%%%%%%%%%%%%%%%%%%%%%%%%%%%%%%%%%%%%%%%%%%%%%%%%%%%%%%%%%%%%%%%%%%%%%%%%%
%%%%%%%%%%%%%%%%%%%%%%%%%%%%%%%%          Figure 12           %%%%%%%%%%%%%%%%%%%%%%%%%%%%%%%%%%%
%%%%%%%%%%%%%%%%%%%%%%%%%%%%%%%%%%%%%%%%%%%%%%%%%%%%%%%%%%%%%%%%%%%%%%%%%%%%%%%%%%%%%%%%
%%%%%%%%%%%%%%%%%%%%%%%%%%%%%%%%%%%%%%%%%%%%%%%%%%%%%%%%%%%%%%%%%%%%%%%%%%%%%%%%%%%%%%%%
%%%%%%%%%%%%%%%%%%%%%%%%%%%%%%%%%%%%%%%%%%%%%%%%%%%%%%%%%%%%%%%%%%%%%%%%%%%%%%%%%%%%%%%%

%%%%%%%%%%%%%%%%%%%%%%%%%%%%%%%%%%%%%%%%%%%%%%%%%%%%%%%%%%%%%%%%%%%%%%%
%%%%%%%%%%%%%%%%%%%%%%%%%%%%%%%%%%%%%%%%%%%%%%%%%%%%%%%%%%%%%%%%%%%%%%%
%%%%%%%%%%%%%%%%%%%%%%%%%%%%%%%%%%%%%%%%%%%%%%%%%%%%%%%%%%%%%%%%%%%%%%%
%%%%%%%%%%%%%%%%%%%%%     			References                %%%%%%%%%%%%%%%%%
%%%%%%%%%%%%%%%%%%%%%%%%%%%%%%%%%%%%%%%%%%%%%%%%%%%%%%%%%%%%%%%%%%%%%%%
%%%%%%%%%%%%%%%%%%%%%%%%%%%%%%%%%%%%%%%%%%%%%%%%%%%%%%%%%%%%%%%%%%%%%%%
%%%%%%%%%%%%%%%%%%%%%%%%%%%%%%%%%%%%%%%%%%%%%%%%%%%%%%%%%%%%%%%%%%%%%%%

\clearpage
%\bibliographystyle{apalike}
%\bibliography{myref}

\begin{thebibliography}{}

\bibitem[Beland and Nair, 2017]{beland2017bayesian}
Beland, J.~J. and Nair, P.~B. (2017).
\newblock Bayesian optimization under uncertainty.
\newblock In {\em NIPS BayesOpt 2017 workshop}.

\bibitem[Bishop, 2006]{bishop2006pattern}
Bishop, C.~M. (2006).
\newblock {\em Pattern recognition and machine learning}.
\newblock springer.

\bibitem[Bogunovic et~al., 2016]{NIPS2016_6080}
Bogunovic, I., Scarlett, J., Krause, A., and Cevher, V. (2016).
\newblock Truncated variance reduction: A unified approach to bayesian
  optimization and level-set estimation.
\newblock In Lee, D.~D., Sugiyama, M., Luxburg, U.~V., Guyon, I., and Garnett,
  R., editors, {\em Advances in Neural Information Processing Systems 29},
  pages 1507--1515. Curran Associates, Inc.

\bibitem[Bryan et~al., 2006]{bryan2006active}
Bryan, B., Nichol, R.~C., Genovese, C.~R., Schneider, J., Miller, C.~J., and
  Wasserman, L. (2006).
\newblock Active learning for identifying function threshold boundaries.
\newblock In {\em Advances in neural information processing systems}, pages
  163--170.

\bibitem[Deisseroth, 2015]{deisseroth2015optogenetics}
Deisseroth, K. (2015).
\newblock Optogenetics: 10 years of microbial opsins in neuroscience.
\newblock {\em Nature neuroscience}, 18(9):1213--1225.

\bibitem[Gessner et~al., 2019]{DBLP:conf/uai/GessnerGM19}
Gessner, A., Gonzalez, J., and Mahsereci, M. (2019).
\newblock Active multi-information source bayesian quadrature.
\newblock In {\em Proceedings of the Thirty-Fifth Conference on Uncertainty in
  Artificial Intelligence, {UAI} 2019, Tel Aviv, Israel, July 22-25, 2019},
  page 245.

\bibitem[Girard et~al., 2003]{NIPS2002_2313}
Girard, A., Rasmussen, C.~E., Candela, J. Q.~n., and Murray-Smith, R. (2003).
\newblock Gaussian process priors with uncertain inputs application to
  multiple-step ahead time series forecasting.
\newblock In Becker, S., Thrun, S., and Obermayer, K., editors, {\em Advances
  in Neural Information Processing Systems 15}, pages 545--552. MIT Press.

\bibitem[Gotovos et~al., 2013]{Gotovos:2013:ALL:2540128.2540322}
Gotovos, A., Casati, N., Hitz, G., and Krause, A. (2013).
\newblock Active learning for level set estimation.
\newblock In {\em Proceedings of the Twenty-Third International Joint
  Conference on Artificial Intelligence}, pages 1344--1350.

\bibitem[Inatsu et~al., a]{inatsu2020b}
Inatsu, Y., Karasuyama, M., Inoue, K., Kandori, H., and Takeuchi, I.
\newblock Active learning of bayesian linear models with high-dimensional
  binary features by parameter confidence-region estimation.
\newblock Neural Computation (to appear).

\bibitem[Inatsu et~al., b]{inatsu2020a}
Inatsu, Y., Sugita, D., Toyoura, K., and Takeuchi, I.
\newblock Active learning for enumerating local minima based on gaussian
  process derivatives.
\newblock Neural Computation (to appear).

\bibitem[Iwazaki et~al., 2019]{iwazaki2019bayesian}
Iwazaki, S., Inatsu, Y., and Takeuchi, I. (2019).
\newblock Bayesian experimental design for finding reliable level set under
  input uncertainty.
\newblock {\em arXiv preprint arXiv:1910.12043}.

\bibitem[Karasuyama et~al., 2018]{karasuyama2018understanding}
Karasuyama, M., Inoue, K., Nakamura, R., Kandori, H., and Takeuchi, I. (2018).
\newblock Understanding colour tuning rules and predicting absorption
  wavelengths of microbial rhodopsins by data-driven machine-learning approach.
\newblock {\em Scientific reports}, 8(1):1--11.

\bibitem[O'Hagan, 1991]{o1991bayes}
O'Hagan, A. (1991).
\newblock Bayes--hermite quadrature.
\newblock {\em Journal of statistical planning and inference}, 29(3):245--260.

\bibitem[Oliveira et~al., 2019]{pmlr-v89-oliveira19a}
Oliveira, R., Ott, L., and Ramos, F. (2019).
\newblock Bayesian optimisation under uncertain inputs.
\newblock In Chaudhuri, K. and Sugiyama, M., editors, {\em Proceedings of
  Machine Learning Research}, volume~89 of {\em Proceedings of Machine Learning
  Research}, pages 1177--1184. PMLR.

\bibitem[Poloczek et~al., 2017]{NIPS2017_7016}
Poloczek, M., Wang, J., and Frazier, P. (2017).
\newblock Multi-information source optimization.
\newblock In Guyon, I., Luxburg, U.~V., Bengio, S., Wallach, H., Fergus, R.,
  Vishwanathan, S., and Garnett, R., editors, {\em Advances in Neural
  Information Processing Systems 30}, pages 4288--4298. Curran Associates, Inc.

\bibitem[Rasmussen and Williams, 2005]{Rasmussen:2005:GPM:1162254}
Rasmussen, C.~E. and Williams, C. K.~I. (2005).
\newblock {\em Gaussian Processes for Machine Learning (Adaptive Computation
  and Machine Learning)}.
\newblock The MIT Press.

\bibitem[Schott, 2016]{schott2016matrix}
Schott, J.~R. (2016).
\newblock {\em Matrix analysis for statistics}.
\newblock John Wiley \& Sons.

\bibitem[Scott et~al., 2011]{scott2011correlated}
Scott, W., Frazier, P., and Powell, W. (2011).
\newblock The correlated knowledge gradient for simulation optimization of
  continuous parameters using gaussian process regression.
\newblock {\em SIAM Journal on Optimization}, 21(3):996--1026.

\bibitem[Settles, 2009]{settles2009active}
Settles, B. (2009).
\newblock Active learning literature survey.
\newblock Computer Sciences Technical Report 1648, University of
  Wisconsin--Madison.

\bibitem[Shahriari et~al., 2016]{shahriari2016taking}
Shahriari, B., Swersky, K., Wang, Z., Adams, R.~P., and De~Freitas, N. (2016).
\newblock Taking the human out of the loop: A review of bayesian optimization.
\newblock {\em Proceedings of the IEEE}, 104(1):148--175.

\bibitem[Shao, 2003]{Shao_2003_book}
Shao, J. (2003).
\newblock {\em Mathematical Statistics}.
\newblock Springer-Verlag New York Inc, 2nd edition.

\bibitem[Shekhar and Javidi, 2019]{pmlr-v89-shekhar19a}
Shekhar, S. and Javidi, T. (2019).
\newblock Multiscale gaussian process level set estimation.
\newblock In Chaudhuri, K. and Sugiyama, M., editors, {\em Proceedings of
  Machine Learning Research}, volume~89 of {\em Proceedings of Machine Learning
  Research}, pages 3283--3291. PMLR.

\bibitem[Song et~al., 2019]{pmlr-v89-song19b}
Song, J., Chen, Y., and Yue, Y. (2019).
\newblock A general framework for multi-fidelity bayesian optimization with
  gaussian processes.
\newblock In Chaudhuri, K. and Sugiyama, M., editors, {\em Proceedings of
  Machine Learning Research}, volume~89 of {\em Proceedings of Machine Learning
  Research}, pages 3158--3167. PMLR.

\bibitem[Srinivas et~al., 2010]{Srinivas:2010:GPO:3104322.3104451}
Srinivas, N., Krause, A., Kakade, S., and Seeger, M. (2010).
\newblock Gaussian process optimization in the bandit setting: No regret and
  experimental design.
\newblock In {\em Proceedings of the 27th International Conference on Machine
  Learning}, pages 1015--1022.

\bibitem[Sui et~al., 2015]{sui2015safe}
Sui, Y., Gotovos, A., Burdick, J., and Krause, A. (2015).
\newblock Safe exploration for optimization with gaussian processes.
\newblock In {\em International Conference on Machine Learning}, pages
  997--1005.

\bibitem[Sui et~al., 2018]{DBLP:conf/icml/SuiZBY18}
Sui, Y., Zhuang, V., Burdick, J.~W., and Yue, Y. (2018).
\newblock Stagewise safe bayesian optimization with gaussian processes.
\newblock In {\em {ICML}}, volume~80 of {\em Proceedings of Machine Learning
  Research}, pages 4788--4796. {PMLR}.

\bibitem[Sutton and Barto, 2018]{sutton2018reinforcement}
Sutton, R.~S. and Barto, A.~G. (2018).
\newblock {\em Reinforcement learning: An introduction}.
\newblock MIT press.

\bibitem[Swersky et~al., 2013]{NIPS2013_5086}
Swersky, K., Snoek, J., and Adams, R.~P. (2013).
\newblock Multi-task bayesian optimization.
\newblock In Burges, C. J.~C., Bottou, L., Welling, M., Ghahramani, Z., and
  Weinberger, K.~Q., editors, {\em Advances in Neural Information Processing
  Systems 26}, pages 2004--2012. Curran Associates, Inc.

\bibitem[Turchetta et~al., 2016]{turchetta2016safe}
Turchetta, M., Berkenkamp, F., and Krause, A. (2016).
\newblock Safe exploration in finite markov decision processes with gaussian
  processes.
\newblock In {\em Advances in Neural Information Processing Systems}, pages
  4312--4320.

\bibitem[Wachi et~al., 2018]{DBLP:conf/aaai/WachiSYO18}
Wachi, A., Sui, Y., Yue, Y., and Ono, M. (2018).
\newblock Safe exploration and optimization of constrained mdps using gaussian
  processes.
\newblock In {\em {AAAI}}, pages 6548--6556. {AAAI} Press.

\bibitem[Xi et~al., 2018]{pmlr-v80-xi18a}
Xi, X., Briol, F.-X., and Girolami, M. (2018).
\newblock Bayesian quadrature for multiple related integrals.
\newblock In {\em International Conference on Machine Learning}, pages
  5369--5378.

\bibitem[Zanette et~al., 2018]{zanette2018robust}
Zanette, A., Zhang, J., and Kochenderfer, M.~J. (2018).
\newblock Robust super-level set estimation using gaussian processes.
\newblock In {\em Joint European Conference on Machine Learning and Knowledge
  Discovery in Databases}, pages 276--291. Springer.

\end{thebibliography}

\end{document}